\documentclass[11pt]{article}

\usepackage[preprint]{acl}

\usepackage{times}
\usepackage{latexsym}

\usepackage[T1]{fontenc}
\usepackage[utf8]{inputenc}

\usepackage{microtype}

\usepackage{inconsolata}

\usepackage{graphicx}
\usepackage{amsmath}
\usepackage{booktabs}
\usepackage{subcaption}
\usepackage[capitalize,noabbrev]{cleveref}

\title{Towards Understanding What State Space Models Learn About Code}

\author{
 \textbf{Jiali Wu\textsuperscript{1, *}} ,
 \textbf{Shweta Verma\textsuperscript{1, *}} ,
 \textbf{Abhinav Anand\textsuperscript{1, *}} ,
 \textbf{Mira Mezini\textsuperscript{1,2,3}}
 \\
 \textsuperscript{1}TU Darmstadt,
 \textsuperscript{2}Hessian Center for Artificial Intelligence, Darmstadt, Germany,
 \\
 \textsuperscript{3}National Research Center for Applied Cybersecurity ATHENE
 \\
  \textsuperscript{*}Equal Contribution,
\\
 \small{
   \textbf{Correspondence:} \href{mailto:email@domain}{shweta.verma@tu-darmstadt.de}
 }
}

\begin{document}
\maketitle
\begin{abstract}

State Space Models (SSMs) have emerged as an efficient alternative to the Transformer architecture. Prior work shows that, when trained under comparable conditions, SSMs can match or surpass Transformers on code understanding tasks. However, their internal mechanisms remain a black box.
We present the first systematic analysis of what SSM-based code models learn along with the direct comparison between SSM and Transformer models in this domain. Our analysis shows that SSMs capture syntactic and semantic structure more effectively than Transformers during pretraining but forgets certain relations during fine-tuning on some tasks. To investigate this behavior, we introduce \textit{SSM-Interpret}, a frequency-domain framework that exposes a “spectral shift" toward short-range dependencies during fine-tuning. Guided by these findings, we propose architectural modifications that significantly improve the performance of SSM-based code model by upto +6 MRR on NLCodeSearch. This demonstrates that our analysis not only explains model behavior but also leads directly to better designs. 
%by upto +6 MRR on NLCodeSearch, +3.8 MRR on Long Context Retrieval and +2 F1 on type inference.% validating that our analysis directly enables better models.

\end{abstract}

\section{Introduction}
Transformers are the dominant architecture for sequence modeling across domains, including the code understanding domain, which is in the focus of this work. However, they suffer from inherent limitations, including quadratic complexity, large data requirement, and positional biases.

State Space Models (SSMs) \cite{linear-ssm} have emerged as a computationally efficient alternative, particularly promising for long-context tasks. While SSM-based architectures have historically struggled to match the empirical performance of transformers \cite{spade, ren2025samba}, \citet{verma-etal-2025-codessm} showed that SSMs can outperform Transformers on code understanding tasks such as retrieval and classification while achieving better compute and sample efficiency. However, the internal mechanisms enabling the performance gains and the extent to which state space models capture syntactic or semantic code structure remain unexplored. Preliminary interpretability work focuses exclusively on selective SSMs\footnote{We make a distinction between SSMs which rely on RNN mode and hardware optimized parallel scan algorithms, like Mamba, and SSMs that rely on convolution during training, like S4D. For simplicity, we refer to the former as selective-SSMs and the latter as SSMs.} and is limited to synthetic tasks. For SSMs no interpretability studies exist. This gap limits our ability to diagnose failures, guide architectural improvements, and predict when SSMs will succeed or struggle on code understanding and generation tasks.

%However, the internal mechanisms enabling these performance gains and the extent to which state-space models capture syntactic or semantic code structure remain unexplored. While there exist preliminary interpretability studies on selective-SSMs \footnote{We make a distinction between SSMs that rely on RNN mode and hardware-optimized parallel-scan algorithms, like Mamba, and SSMs that rely on convolution during training, like S4D. We refer to the former as selective-SSMs and the latter simply as SSMs.}, is missing for SSMs. Moreover, existing studies on selective-SSMs are limited to synthetic data.

The interpretability gap stands in stark contrast to the Transformer ecosystem, where attention maps and learned representations have been rigorously analyzed to understand how code properties are captured \cite{anand-etal-2024-critical, capture, astprobe, probe_karmakar}. In this work, we turn our attention to SSMs and real-world data. Specifically, we (a) conduct the first systematic comparative analysis of SSM and Transformer representations in the code domain and (b) propose the first framework for analyzing the convolution kernel of SSM blocks in a multi-layer model. 

%We distinguish between two classes of SSMs: \emph{selective SSMs} that operate in recurrent (RNN) mode with hardware-optimized parallel scan algorithms (e.g., Mamba), and \emph{convolutional SSMs} that rely on FFT-based convolution during training (e.g., S4D). While there exist preliminary interpretability studies on selective SSMs \cite{mamba2025achilles, transformerinterpretabilitytransfer, hiddenattention_mamba}, albeit limited to synthetic data, such studies are completely missing for convolutional SSMs. 

%To address this gap, in this work, we turn our attention to SSMs and real-world data. Specifically, (a) we conduct the first systematic comparative analysis of a multi-layer SSM and Transformer representations in the code domain and (b) propose the first framework for analyzing the convolution kernel of SSM blocks in a multi-layer model. 

%To address this gap, we present the first systematic interpretability study of native SSMs and real-world datasets. Specifically, we (1)  conduct the first comparative analysis of learned representations in multi-layer SSM and Transformer models for code, and (2) introduce a novel framework for analyzing convolution kernels across SSM layers, enabling inspection of how these models process and prioritize different dependency ranges.

Our focus on SSM, despite their current limited real-world application, is motivated by recent theoretical works, such as \citet{nishikawa2025state}, which showed that, when combined with non-nonlinearities, SSMs can indeed perform dynamic token selection comparable to self-attention. At the same time, multiple works have shown the failure of selective-SSMs on tasks that require dynamic token selection, such as input copying and state tracking \cite{mamba2025achilles, jafari2024mambalrp, repeatmetransformersbetter}. SSMs are also significantly more efficient compared to both transformers and selective-SSMs. %scaling linearly with respect to input context and hidden dimension. 
Also, SSMs can extrapolate to 8x the pretraining context \cite{verma-etal-2025-codessm} while Mamba cannot extrapolate \cite{azizi2025mambaextend}.

While SSMs have outperformed transformers at similar parameter and dataset scale, they have not been scaled to billions of parameters and trillions of tokens and have not been applied on generative tasks. One of the key reasons for this has been the failure of SSMs on certain tasks such as type inference, which are similar to generative tasks (large class size; long and short range dependency). While scaling is an important aspect to improve performance and generalizability of the models, naive scaling will lead to limited gains. Thus, in this work we focus on understanding why SSM fails on certain tasks, instead of scaling.

For our study, we use an existing SSM-based model, CodeSSM \cite{verma-etal-2025-codessm}, which has shown superior performance to Transformer on numerous tasks but lag behind on type inference. We compare the internal representation of CodeSSM with RoCoder \cite{verma-etal-2025-codessm} as both the models has been trained at the same parameter and data scale, removing these factors as confounders.

%on encoder-only architectures, as prior work indicates their superior capacity for capturing code syntax and semantics \cite{anand-etal-2024-critical}. Specifically, we analyze the hidden representation of an SSM-based code model (CodeSSM) and a transformer-based code model (RoCoder), both trained under similar conditions.

Our comparative investigation shows that while pretrained CodeSSM captures code properties more effectively than RoCoder, this advantage collapses during finetuning. Specifically, when finetuned on type inference task, CodeSSM forgets critical syntactic relations that RoCoder retains. 
To diagnose this failure, we develop a novel framework for analyzing the convolution kernels of multi-layer SSMs. The kernel analysis reveals a “spectral shift”: during fine-tuning, the kernels in early layers bias heavily towards short-range dependencies, effectively discarding the long-range context required for type inference. Guided by these insights, we introduce architectural changes that prevent this degeneration, improving the model’s ability to reason about complex code structures. These changes can form the basis for scaling and generative application of SSMs in future works.

\begin{figure*}[h!]
\centering
    \includegraphics[width=0.9\linewidth, page=1, viewport=7 7 477 205, clip]{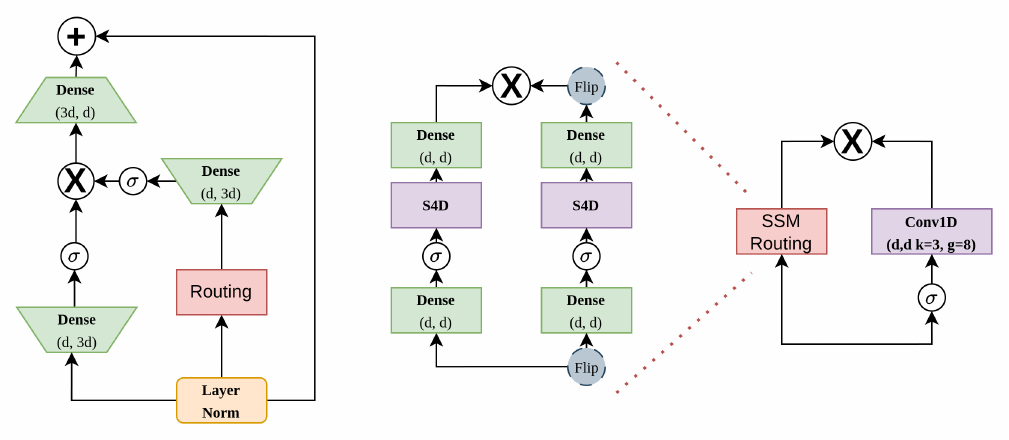}
    \caption{The CodeSSM layer architecture (left) showing the original routing mechanism (center) and the proposed routing (right).}
    \label{cssm}   
\end{figure*}

In summary, our main contributions are:

\begin{itemize}

\item \emph{Comparative Hidden Representation Analysis}: We provide the first direct comparison of hidden states in SSM and Transformer code models, where code structure (AST, DFG) enable rigorous evaluation of syntactic and semantic understanding. We empirically demonstrate that while CodeSSM captures better syntactic and semantic relations compared to RoCoder during pretraining, it forgets certain relations during fine-tuning.

\item \emph{SSM Kernel Analysis Framework:} We propose the first framework for analyzing convolution kernels in multi-layer SSMs. Using this method, we identify a strong correlation between the high-frequency spectral shift in CodeSSM's kernel and its failure on type inference.

\item \emph{Interpretability-Driven Improvements:} Leveraging our analytical findings, we propose two enhanced variants of CodeSSM. These architectures mitigate the spectral shift, yielding significant performance gains of upto +6 MRR on NLCodeSearch, +3.8 MRR on Long Context Retrieval and +2 F1 type inference, validating that insights from our analysis translate directly into performance improvements.
\end{itemize}

\section{Related Works}

\textbf{Hidden Representation Analysis.}
Various methodologies have been proposed to analyze the internal representations of LLMs. The most prevalent approach employs “probes” -- trained classifiers or learned transformations -- to evaluate hidden states \cite{probe_ahmed, probe_karmakar, astprobe, probe_yang}. However, classifier-based probes suffer from several significant limitations \cite{tale, control_task, survey_probe}. To address these shortcomings, \citet{anand-etal-2024-critical} proposed applying DirectProbe \cite{directprobe}, a classifier-free methodology, to analyze the hidden representations of pretrained code models directly.

\textbf{Interpretability of SSMs.}
Recent work has begun analyzing SSM architectures through various lenses. \citet{mamba2025achilles} identify limitations of Mamba on synthetic copying tasks but do not examine real-world domain adaptation or convolution-based SSMs. \citet{transformerinterpretabilitytransfer} analyzed if the interpretability methods developed for transformer can be used for architectures such as Mamba but do not evaluate if they apply to general convolution based SSMs. \citet{hiddenattention_mamba} reformulate Mamba as an implicit attention mechanism, offering theoretical insights and attention based analysis.

Our work fills critical gaps by comparing the capabilities of SSMs and transformers on real-world tasks, and we are the first to develop a framework for layer-wise frequency-domain analysis of the learned convolutional kernels, resulting in concrete architectural improvements. 
% Our work fills critical gaps by performing a comparative analysis of code understanding ability of SSMs and transformers on real world tasks, and we are the first to develop \textit{SSM-Interpret} which performs layer-wise frequency domain analysis of the learned kernels which translates into concrete architectural improvements. 

% we are the first to characterize how SSM representations degrade during finetuning; (3) Our \textit{SSM-Interpret} performs layer-wise frequency domain analysis of the learned kernels which translates into concrete architectural improvements. 
% operates directly on learned kernels in the frequency domain, enabling layer-wise analysis that translates into concrete architectural improvements.

We discuss additional related works on analysis using spectral methods and control theory in Appendix \ref{related mamba}.

\begin{figure*}[t]
\centering
    \begin{subfigure}[b]{0.99\textwidth}
     \includegraphics[width=0.33\linewidth]{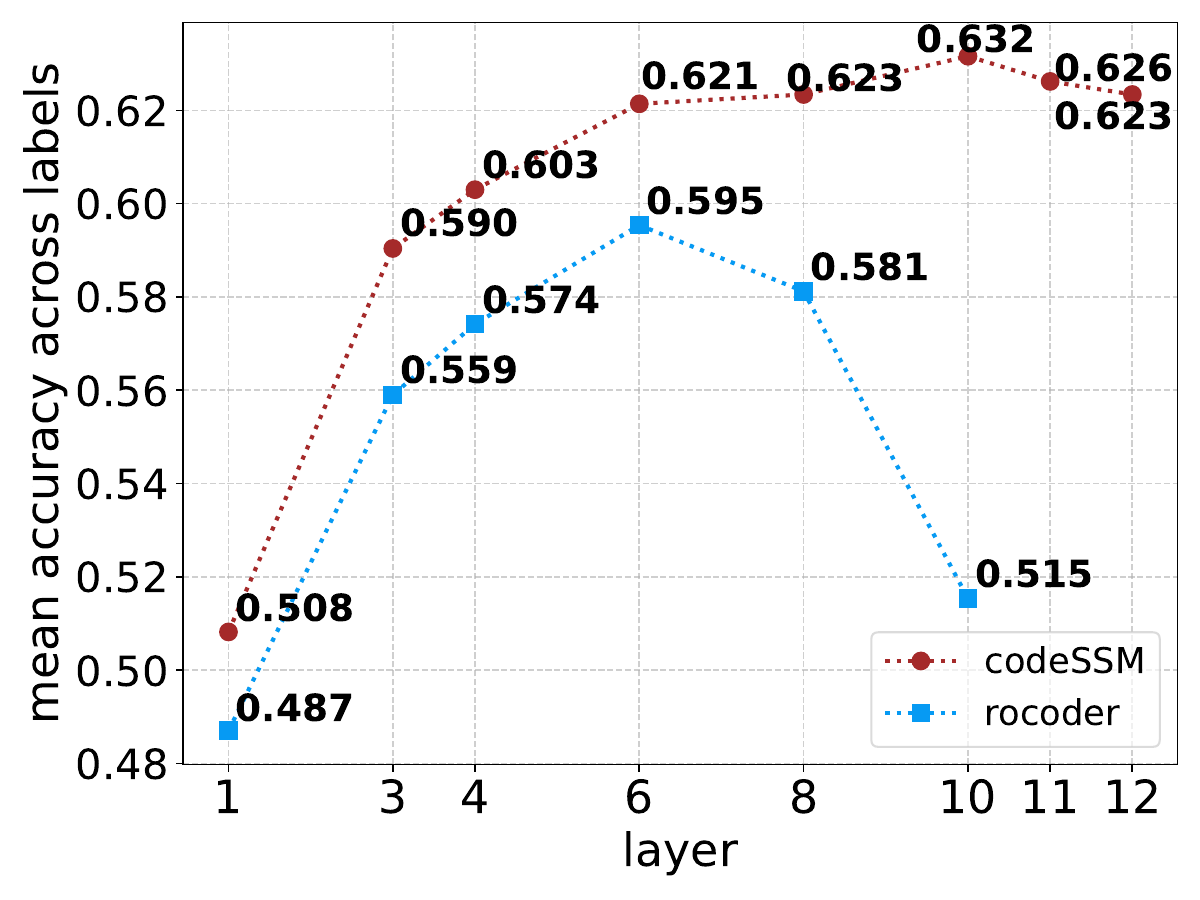} 
     \includegraphics[width=0.33\linewidth]{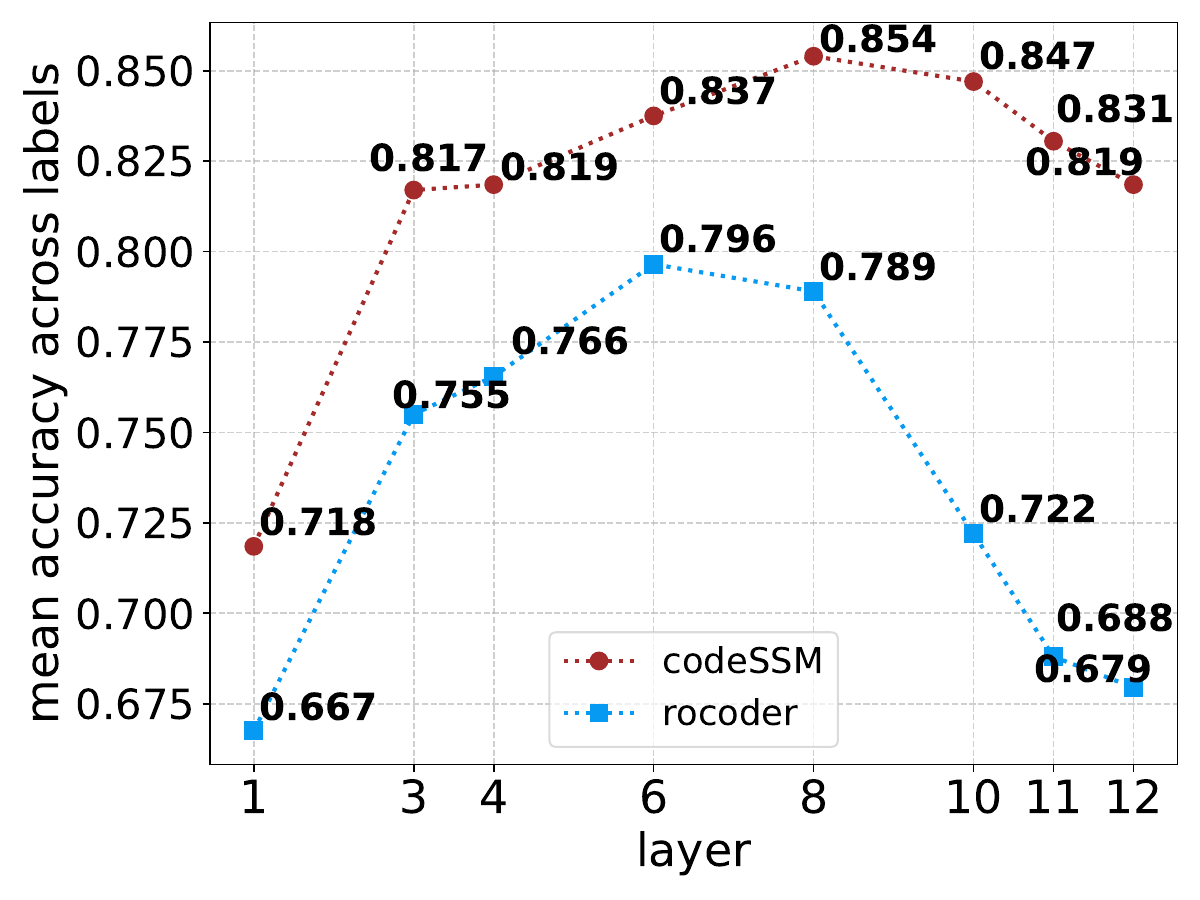}
     \includegraphics[width=0.33\linewidth]{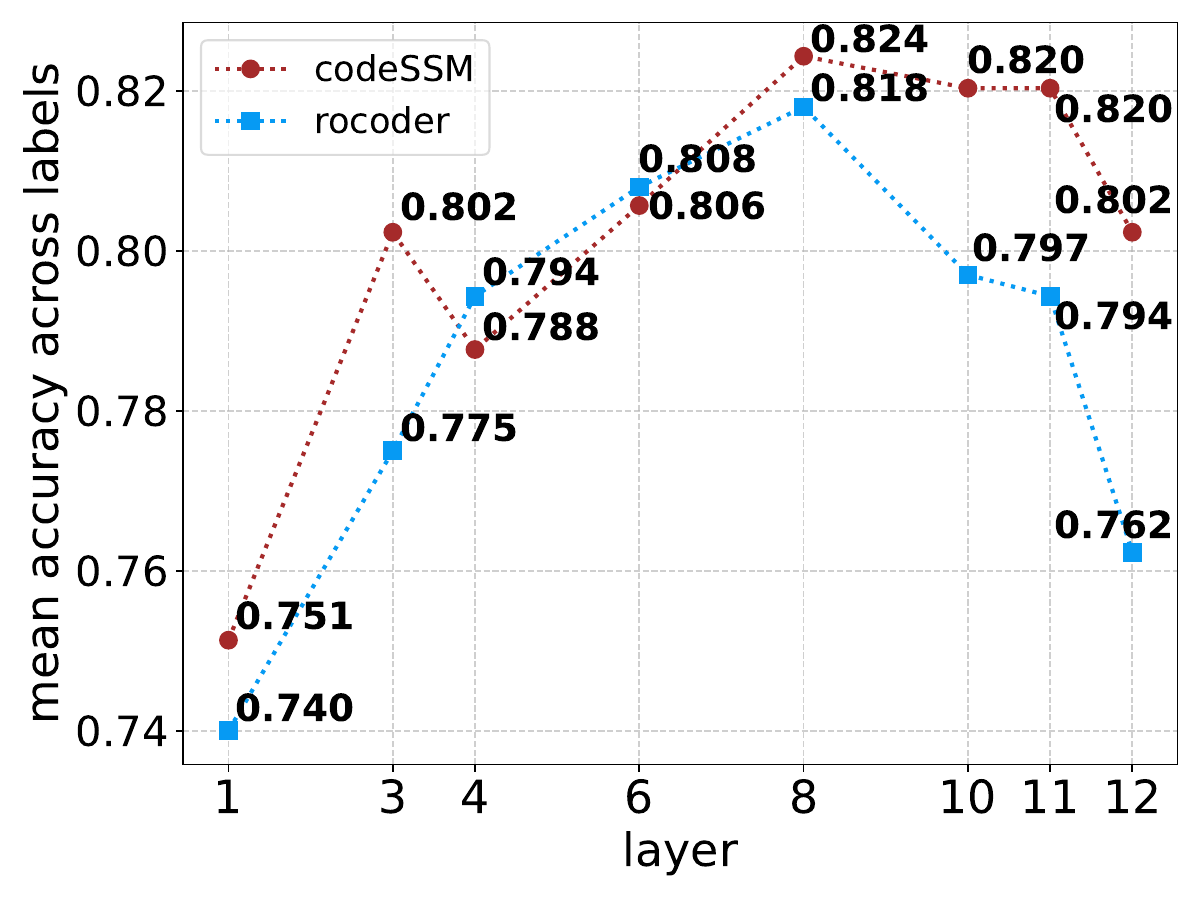}
     \caption{Comparison of hidden representation of CodeSSM and RoCoder.}% on distance (left), siblings (center) and edge (right) prediction tasks.}
       \label{fig: hiddenrepr_pretr}
    \end{subfigure}

%\end{figure*}

%\begin{figure*}[h!]
%\centering
    \begin{subfigure}[b]{0.99\textwidth}
     \includegraphics[width=0.33\linewidth]{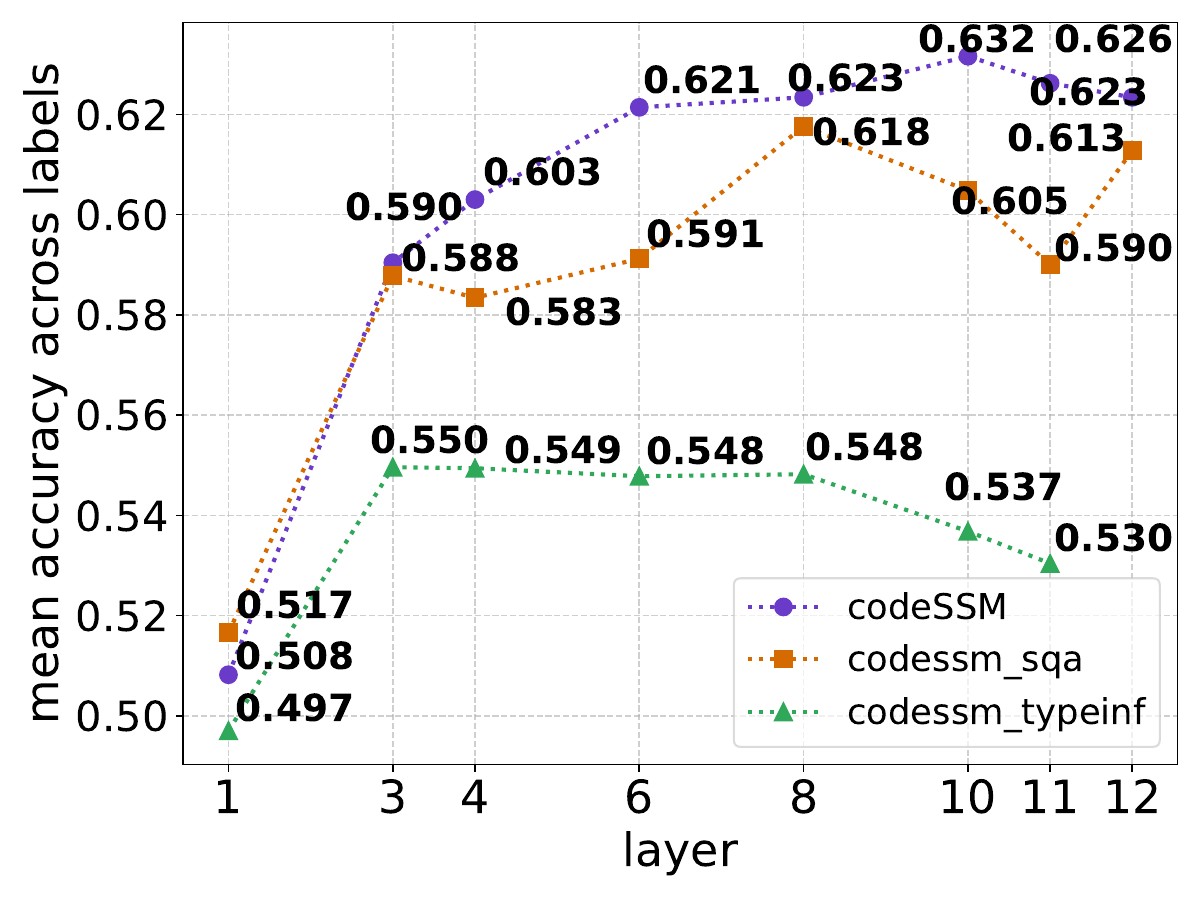} 
     \includegraphics[width=0.33\linewidth]{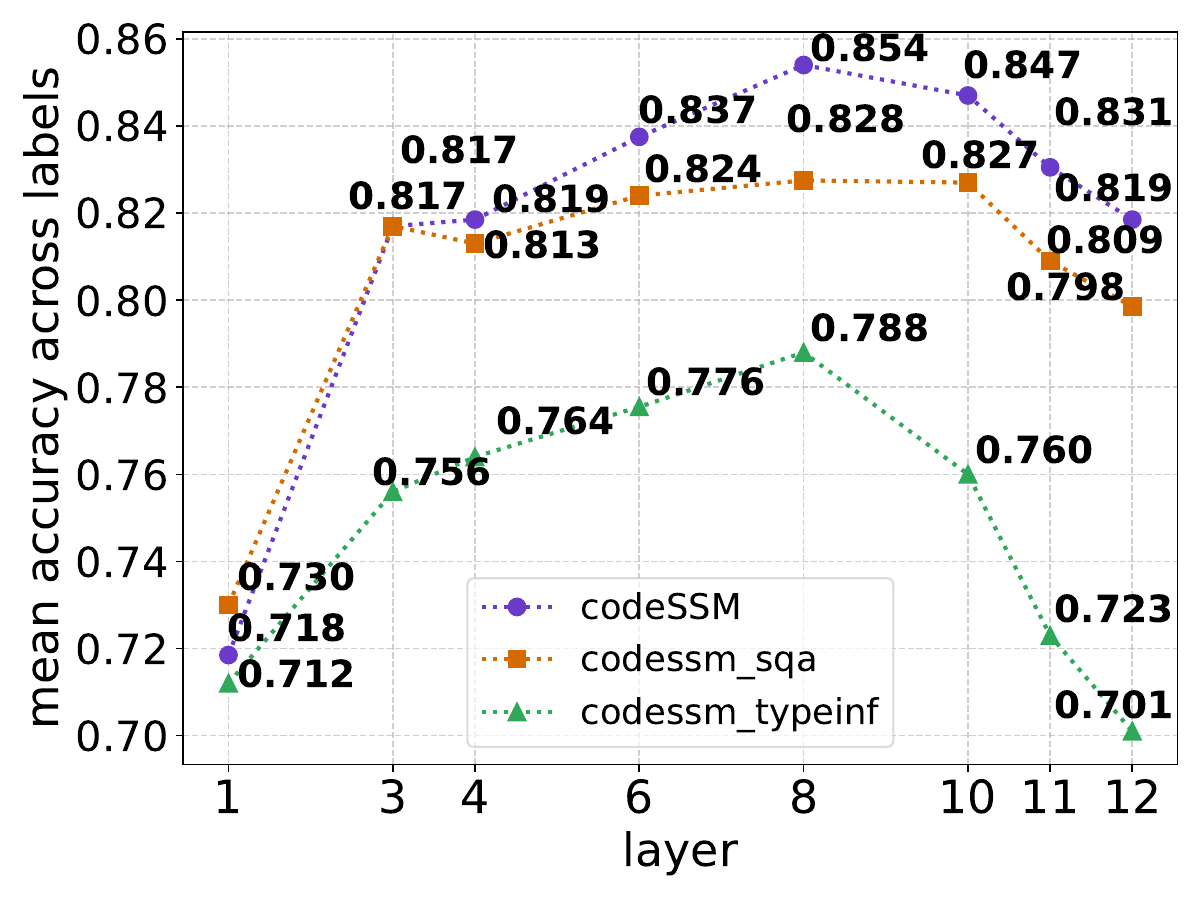}
     \includegraphics[width=0.33\linewidth]{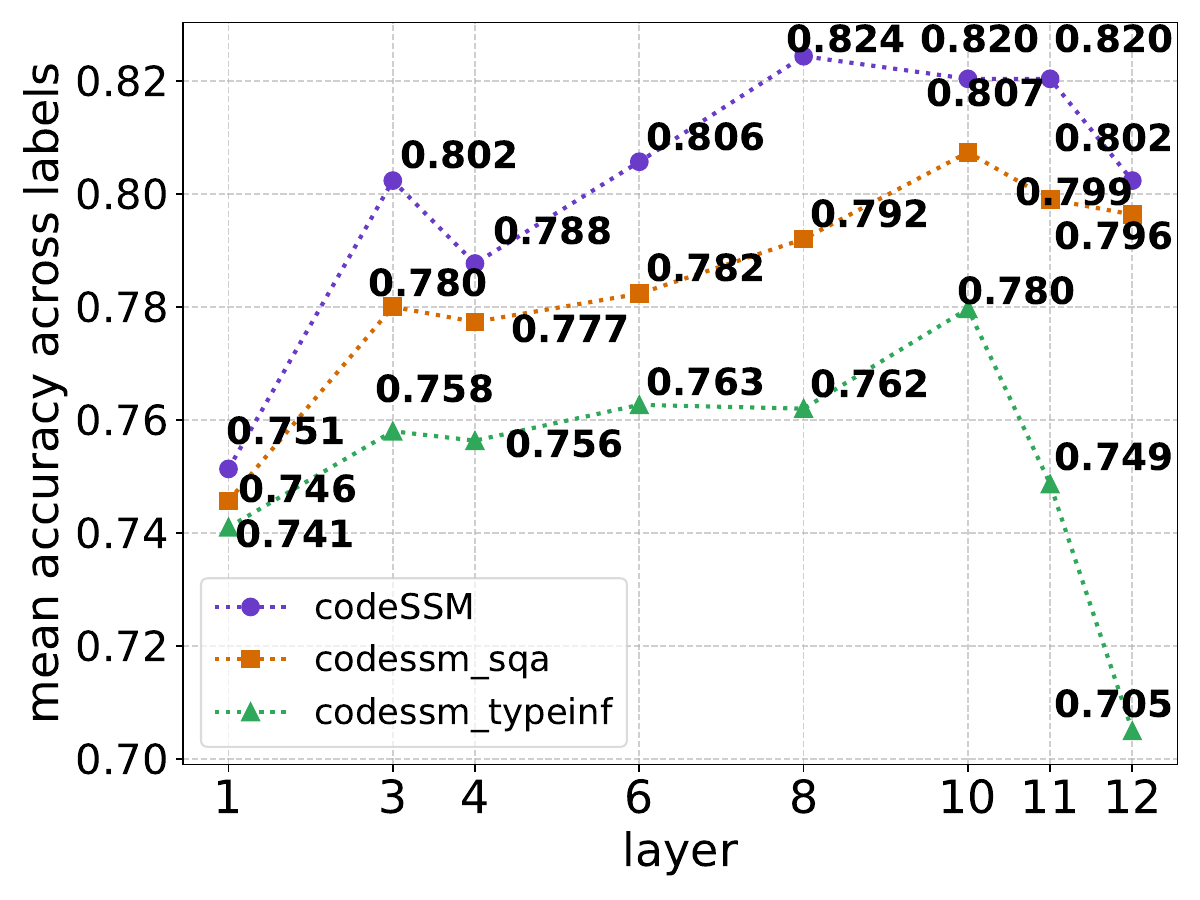} 
     \caption{Comparison of hidden representation of CodeSSM and its finetuned variants} %on distance (left), siblings (center) and edge (right) prediction tasks..}
     \label{fig: hidden repr finetuned_codessm}
    \end{subfigure}
    
%\end{figure*}    

%\begin{figure*}[h!]
%\centering
    \begin{subfigure}[b]{0.99\textwidth}
     \includegraphics[width=0.33\linewidth]{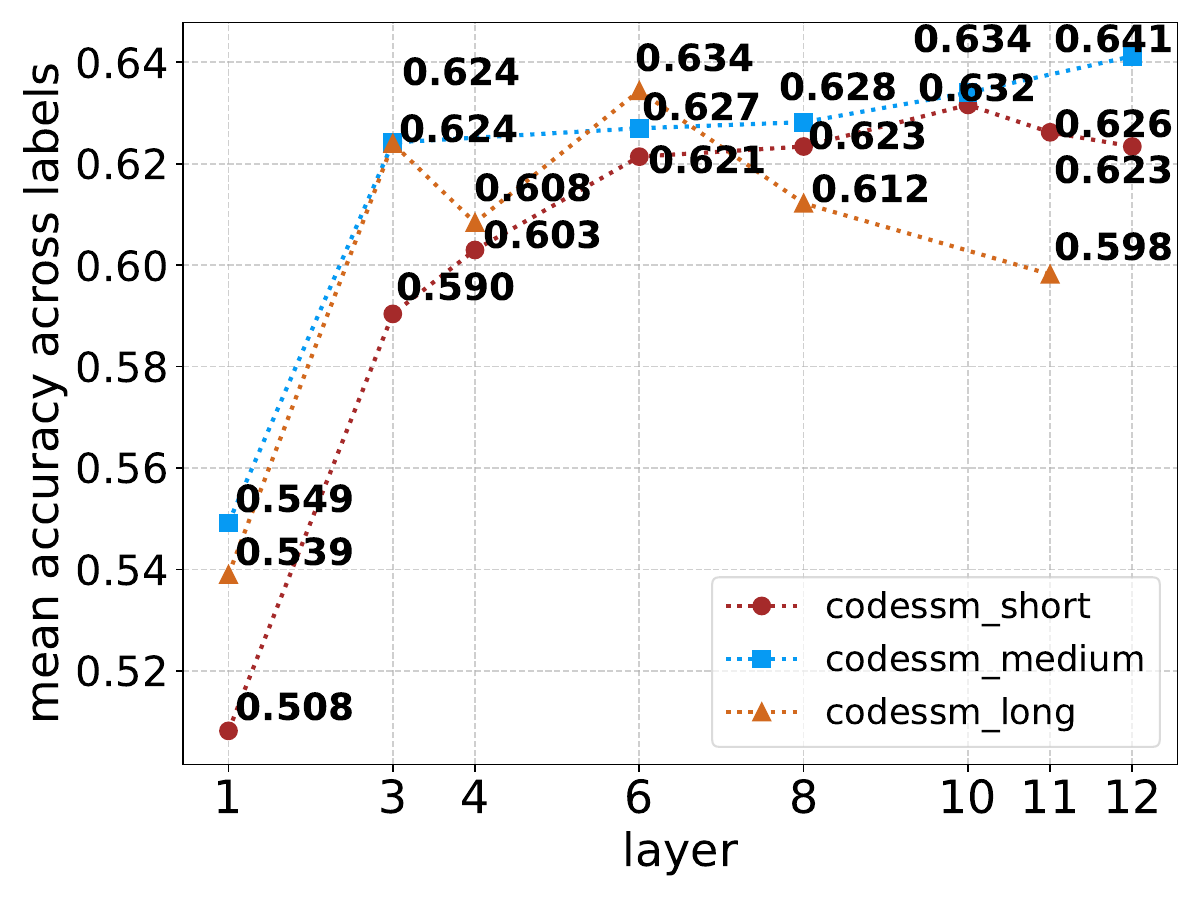} 
     \includegraphics[width=0.33\linewidth]{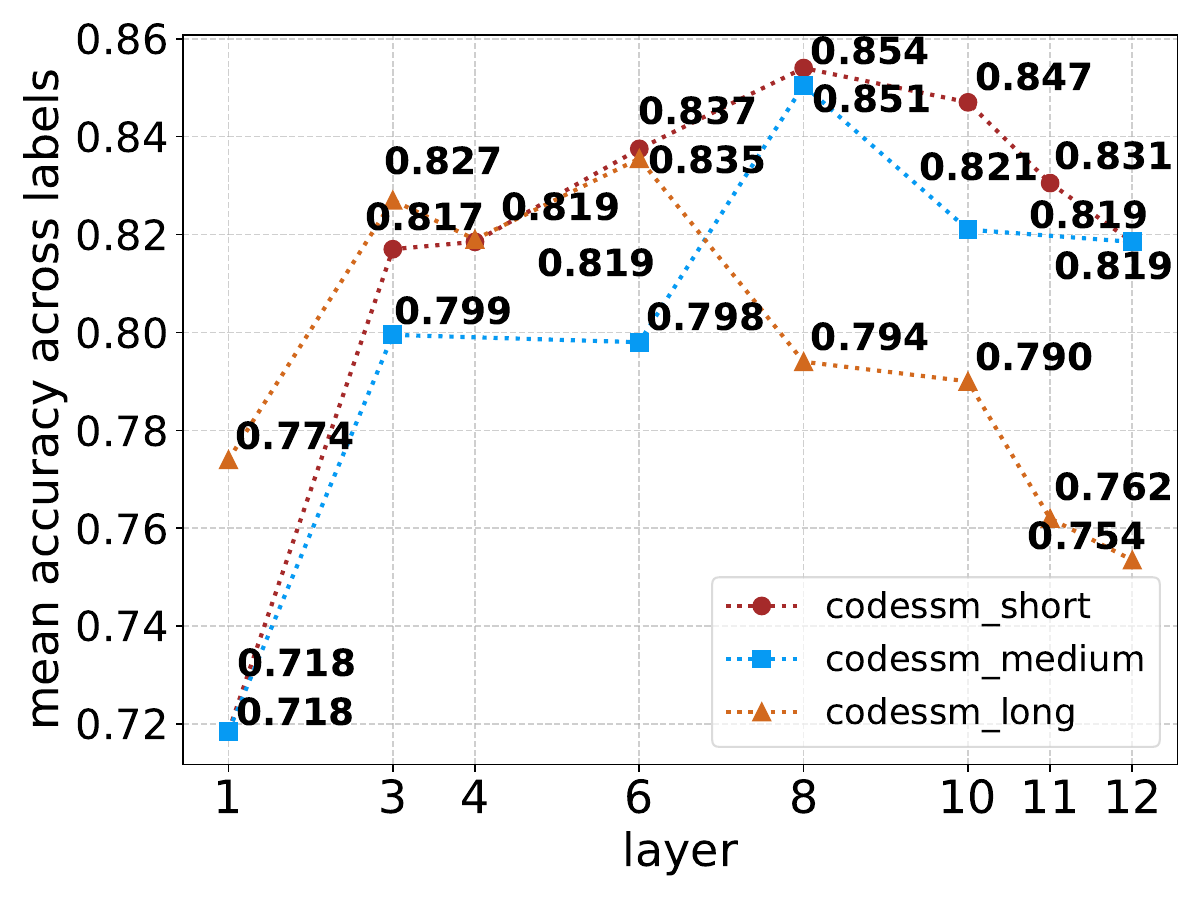}
     \includegraphics[width=0.33\linewidth]{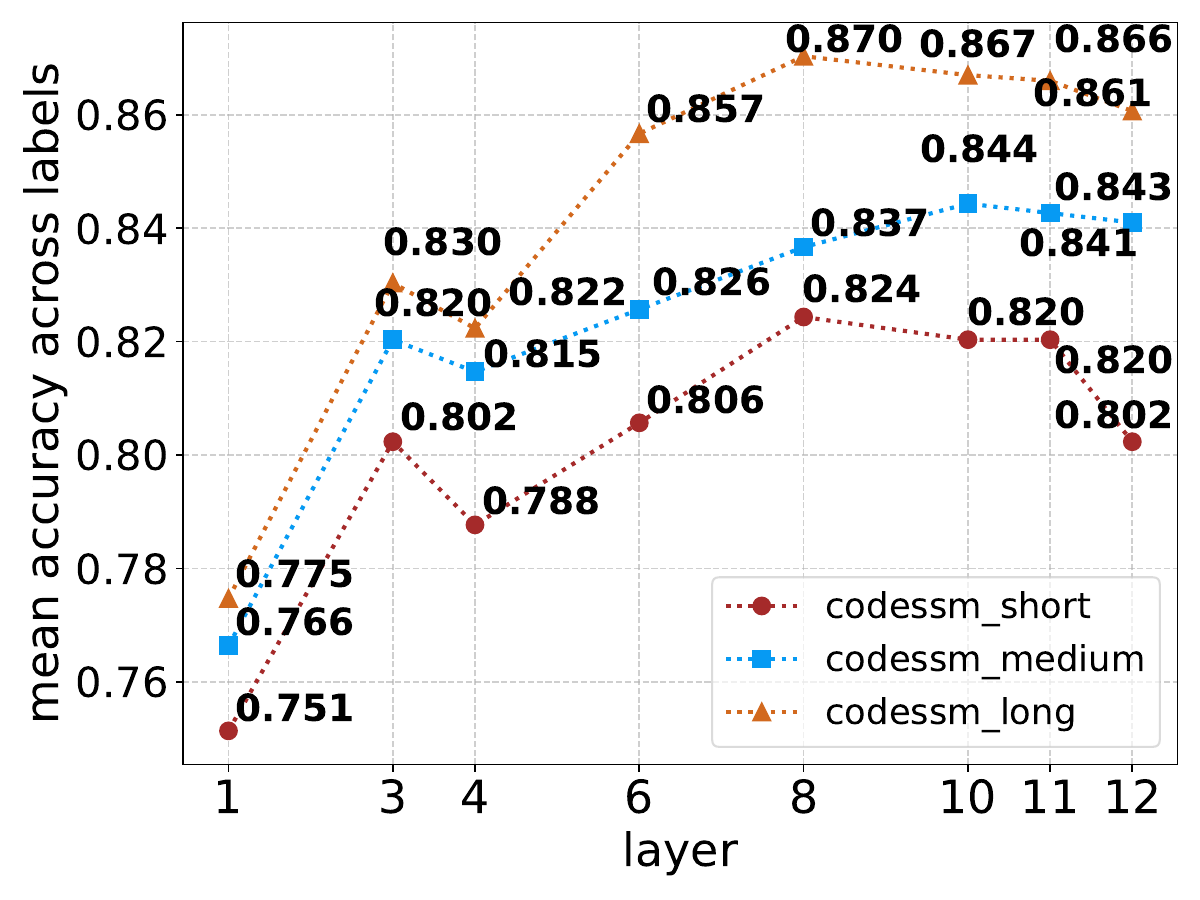} 
     \caption{Comparison of hidden representation of CodeSSM on varying context lengths}% on distance (left), siblings (center) and edge (right) prediction tasks.}
     \label{fig: hidden repr diff lengths}
    \end{subfigure}  
\caption{Result of hidden representation analysis in terms of mean accuracy across task labels on distance (left), siblings (center) and edge (right) prediction tasks.} %The left row shows the performance on distance prediction task, center shows performance on sibling prediction task and right shows it on edge prediction task.}
\label{fig: hidden_result}
\end{figure*}

\section{Comparative Hidden Representation Analysis}
In this section, we describe the comparative hidden representation analysis of SSM and transformers.

\subsection{Analysis Setup}
To evaluate the efficacy of SSM-based models in capturing code properties, we perform a comparative analysis between a SSM and a Transformer model.  We utilize the CodeSSM and RoCoder models \cite{verma-etal-2025-codessm} (architecture details in Appendix \ref{models}), both having similar size and trained on same amount of data, thus removing these factors as confounders. In addition to the pre-trained versions, we evaluate models fine-tuned on StackOverflow question-answer retrieval (SQA) and type inference tasks (denoted with suffixes \texttt{-sqa} and \texttt{-typeinf}). These specific tasks were selected to investigate a critical performance anomaly: CodeSSM significantly outperforms RoCoder on SQA but lags behind on type inference \cite{verma-etal-2025-codessm}. The performance anomaly suggests that CodeSSM might not be a good architecture for tasks with a large class size and long-and-short dependencies such as type inference.  Generative tasks also have similar modeling requirements as type inference and hence, improving the architecture, and performance, on type inference might also help improve SSM's performance on generative modeling. 

The evaluation is performed using the classifier-free framework established by \citet{anand-etal-2024-critical}, which leverages the \emph{DirectProbe} algorithm \cite{directprobe} to measure the quality of hidden representation. By using a classifier-free interpretability method we remove any impact of classifier training on conclusions drawn. Direct probe uses convex optimization to create disjoint clusters of hidden representation
where a single class label may be represented by multiple distinct clusters 
(see Appendix \ref{app: dp} for details). To quantify representational quality, these generated clusters are used for nearest-neighbor classification of held-out data. High classification accuracy indicates that the the model has effectively captured the code property.

 %of their hidden representations against a Transformer model  

%Shorten and move the rest to appendix or merge with evaluation task into one subsection: setup

%can be moved to Appendix
%\subsection{DirectProbe}
% 
%\subsection{Evaluation Tasks}
We evaluate the hidden representations using three probing tasks established by \citet{anand-etal-2024-critical}, designed to isolate specific syntactic and semantic capabilities. Syntactic structure is modeled via the Abstract Syntax Tree (AST), while semantic relationships are captured using the Data Flow Graph (DFG), where edges denote dependencies between variables. The tasks are: 1) \textbf{AST Sibling Prediction}: Determines whether two tokens share a parent node in the Abstract Syntax Tree (AST). This task measures the model’s grasp of local syntactic relationships; 2)\textbf{AST Distance Prediction}: Estimates the shortest path length between two tokens in the AST. This serves as a proxy for short and long-range syntactic understanding and the model’s understanding of program flow \cite{anand-etal-2024-critical}; and 3)\textbf{DFG Edge Prediction}: Identifies whether a data flow edge exists between two variable occurrences in the Data Flow Graph (DFG). This task evaluates the model’s encoding of semantic dependencies.

% \begin{enumerate}
% \item \emph{AST Sibling Prediction}: Determines whether two tokens share a parent node in the Abstract Syntax Tree (AST). This task measures the model’s grasp of local syntactic relationships.

% \item \emph{AST Distance Prediction}: Estimates the shortest path length between two tokens in the AST. This serves as a proxy for short and long-range syntactic understanding and the model’s understanding of program flow \cite{anand-etal-2024-critical}.

% \item \emph{DFG Edge Prediction}: Identifies whether a data flow edge exists between two variable occurrences in the Data Flow Graph (DFG). This task evaluates the model’s encoding of semantic dependencies.
% \end{enumerate}
\begin{figure}[!]
\centering
    \begin{tabular}{@{}c@{\hspace{0.1mm}}c@{}}
        \includegraphics[width=0.53\linewidth]{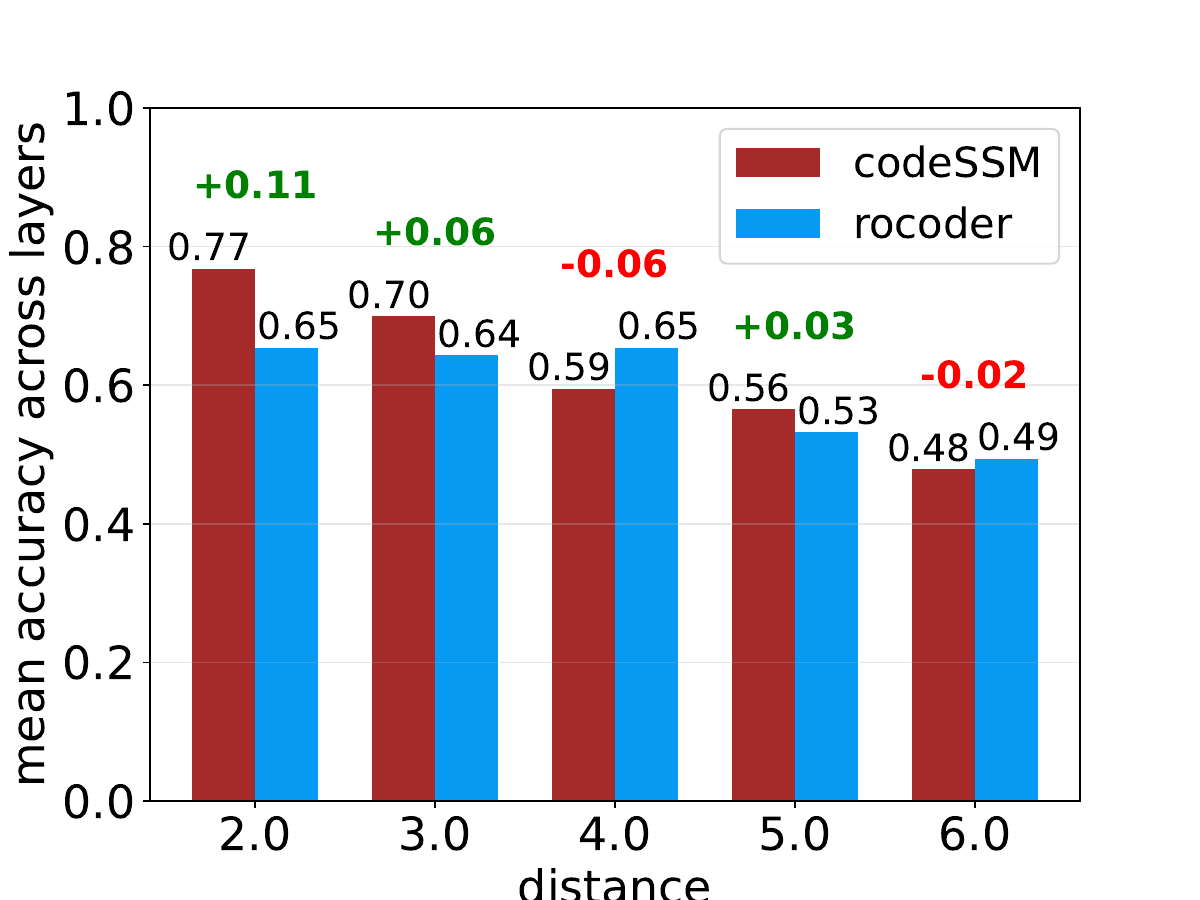} &
        \includegraphics[width=0.53\linewidth]{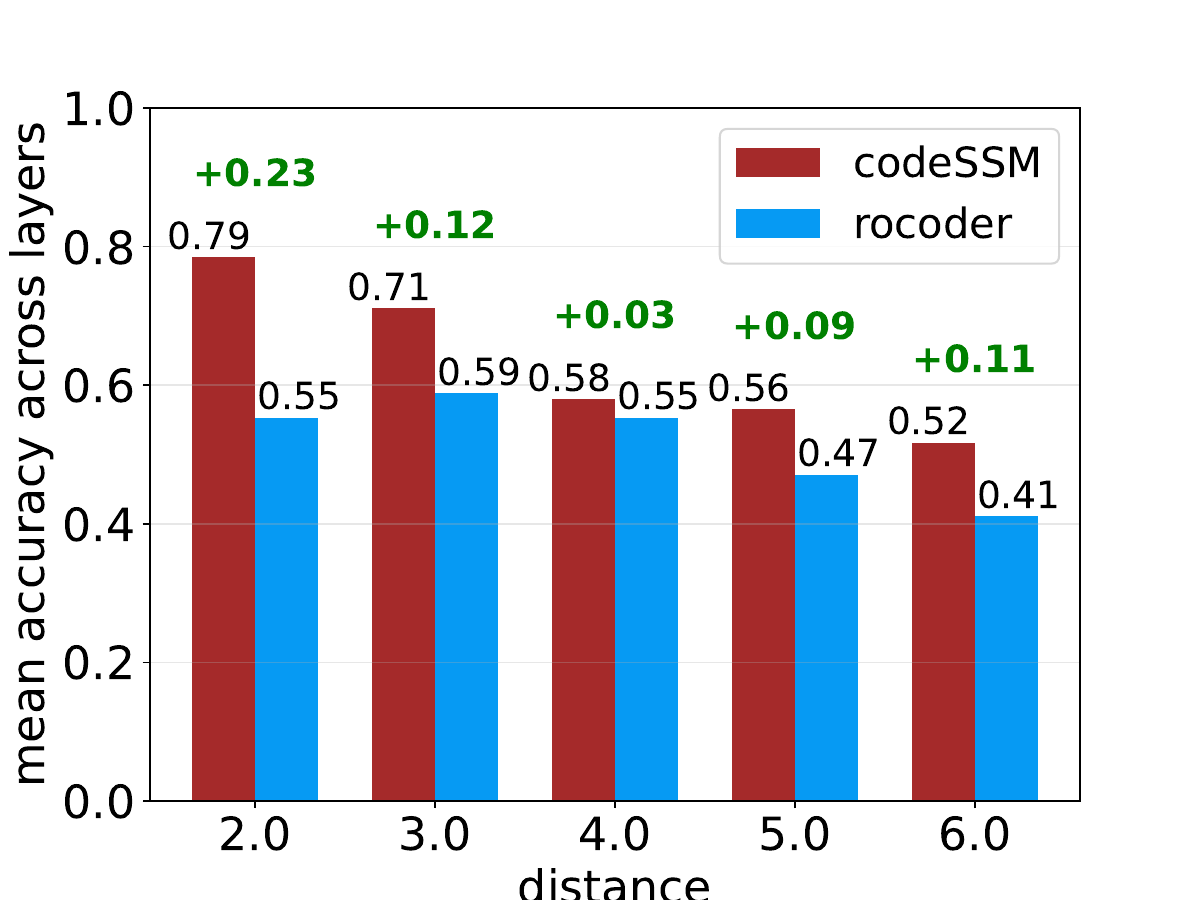}
        \\[0.5ex]
        \includegraphics[width=0.53\linewidth]{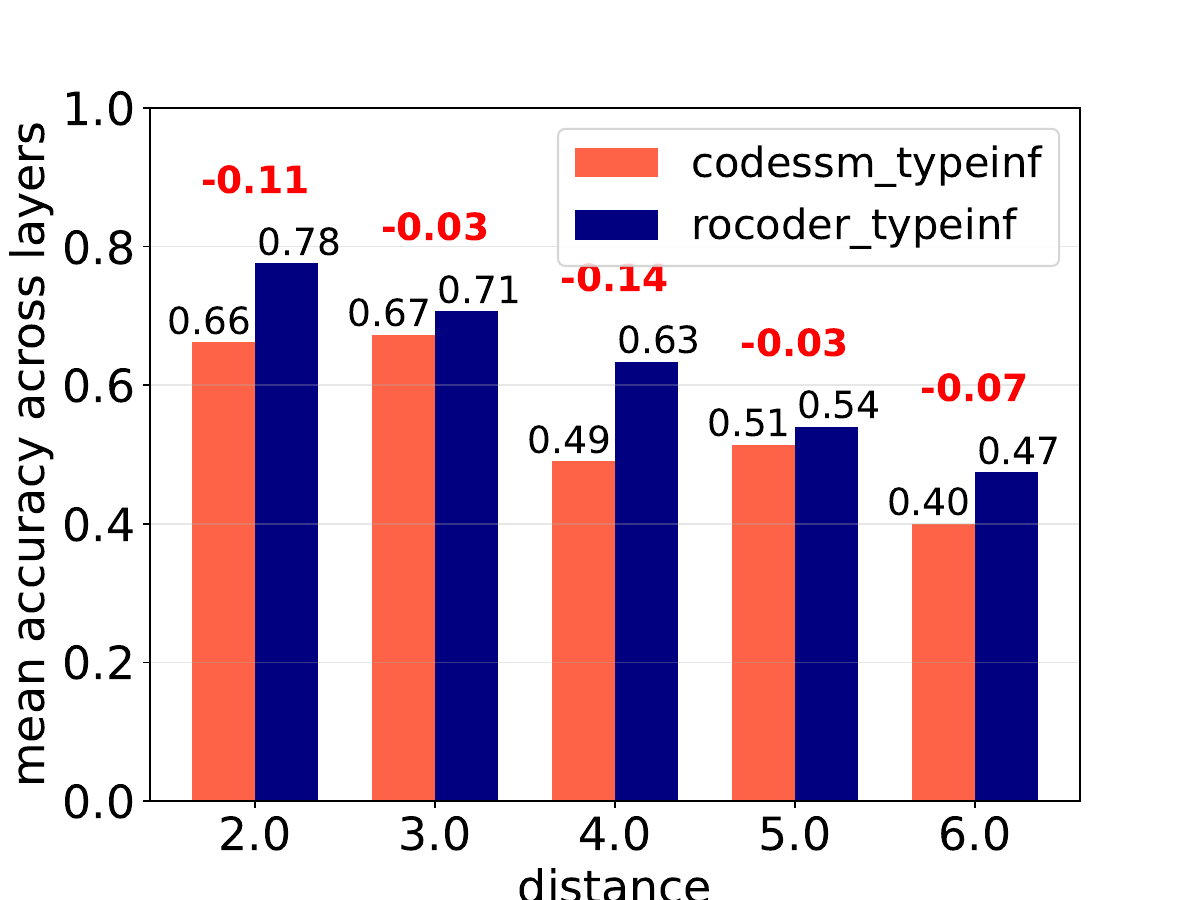} &
        \includegraphics[width=0.53\linewidth]{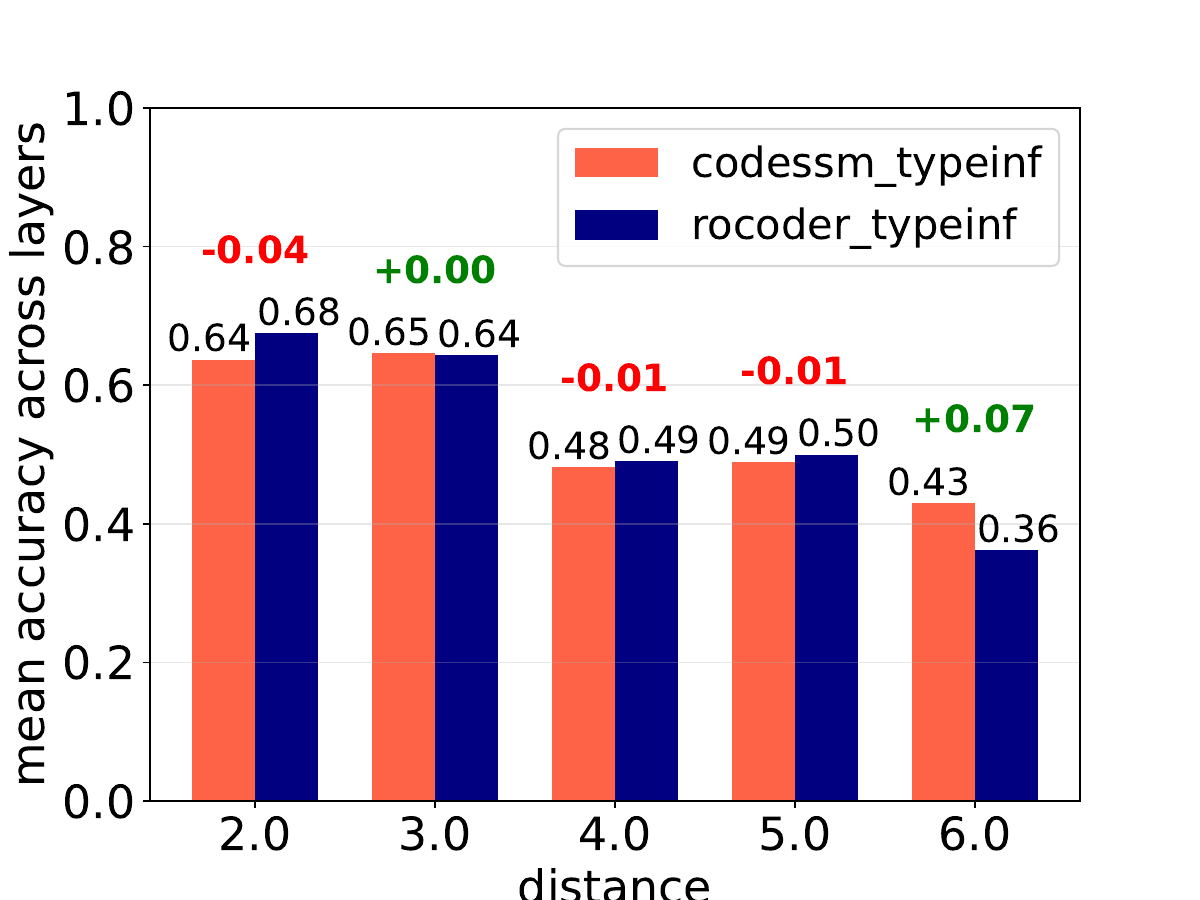}
    \end{tabular}
    
     \caption{Accuracy of CodeSSM, CodeSSM-typeinf, Rocoder and Rocoder-typeinf on distance prediction tasks for layers 6 and 10.}
     \label{fig: distancelayer8910-codessmtypeinf}
\end{figure}

\subsection{Analysis Results} \label{hidden_analysis}
We performed direct probe analysis of CodeSSM and RoCoder on varying lengths: Short (up to 500), Medium (1k-2k), Long (2k-8k). \cref{fig: hidden_result} presents the probing accuracy across tasks.

\textbf{Pre-training.} CodeSSM exhibits superior syntactic capture compared to RoCoder (\cref{fig: hiddenrepr_pretr}). While semantic performance is comparable in early layers, CodeSSM significantly outperforms RoCoder in deeper layers. Notably, RoCoder suffers from representational degradation in the final layers, a phenomenon consistent with other Transformer architectures \cite{anand-etal-2024-critical}; the inclusion of Rotary Positional Embeddings (RoPE) appears insufficient to mitigate this limitation.

\textbf{Length Extrapolation.} In \cref{fig: hidden repr diff lengths}, we observe that CodeSSM shows the best accuracy in capturing semantic properties (DFG task) for long lengths, showing remarkable length extrapolation up to 8k context length (32x the pretraining context of 256). The superior performance on the DFG task implies that CodeSSM captures long-range dependencies very well especially for long lengths. However, CodeSSM shows representational degradation in deeper layers for local dependencies (i.e., siblings tasks) with long inputs. Nevertheless, CodeSSM retains both syntactic and semantic information upto 8x pretraining context, while RoCoder does not (see Appendix \ref{length-extra}).%; along with explanation of length extrapolation in SSMs).
%The results for RoCoder on longer lengths is presented in \cref{length-extra} and it shows that the performance gap between CodeSSM and RoCoder increases with increasing context length.
%CodeSSM captures better syntactic and semantic properties at longer lengths than RoCoder. 

\textbf{Fine-tuning} reveals distinct adaptation dynamics between the architectures. CodeSSM forgets certain code relations learned during pretraining (\cref{fig: hidden repr finetuned_codessm}). While this degradation is negligible for SQA, with CodeSSM\_sqa performing better than RoCoder\_sqa, it is significant in case of type inference. To diagnose this, we analyze performance by token distance in layers 6 and 10 (\cref{fig: distancelayer8910-codessmtypeinf}). We observe that fine-tuning on type inference causes CodeSSM to lose syntactic information, particularly at short ranges. Conversely, RoCoder significantly improves its short-range syntactic modeling after fine-tuning on type inference.

This divergence is consistent with the downstream performance. Type inference demands both local (short-range) and global (long-range) context understanding. While RoCoder successfully adapts to these requirements, improving its short-range syntactic understanding during fine-tuning, CodeSSM fails to retain its pre-trained syntactic and semantic knowledge. Consequently, when a task requires both short and long range understanding, CodeSSM’s representations fail.

% As discussed above, CodeSSM captures more syntactic and semantic properties of code than RoCoder on short lengths. We observe that the gap between CodeSSM and RoCoder increases with increasing context length. CodeSSM captures better syntactic and semantic properties at longer lengths. 

While previous work, such as \citet{spade}, has argued that SSMs are poor at capturing short-range dependencies, our work is the first to provide evidence for this in a multilayer SSM-based model. This complementarity -- SSMs for global context and Transformers for local dependencies -- provides potential explanation for the success of hybrid models \cite{ren2025samba, mamba2}. However, the underlying mechanics remain opaque, prompting two critical questions: 1) Why do some layers capture more code properties? 2) Why does code understanding decrease on the type inference task? 
%3) What range of token relations CodeSSM capture? 
To answer these questions we designed the kernel analysis.

\begin{figure*}[h!]
\centering
    \begin{subfigure}[b]{0.99\textwidth}
     \includegraphics[width=0.5\linewidth]{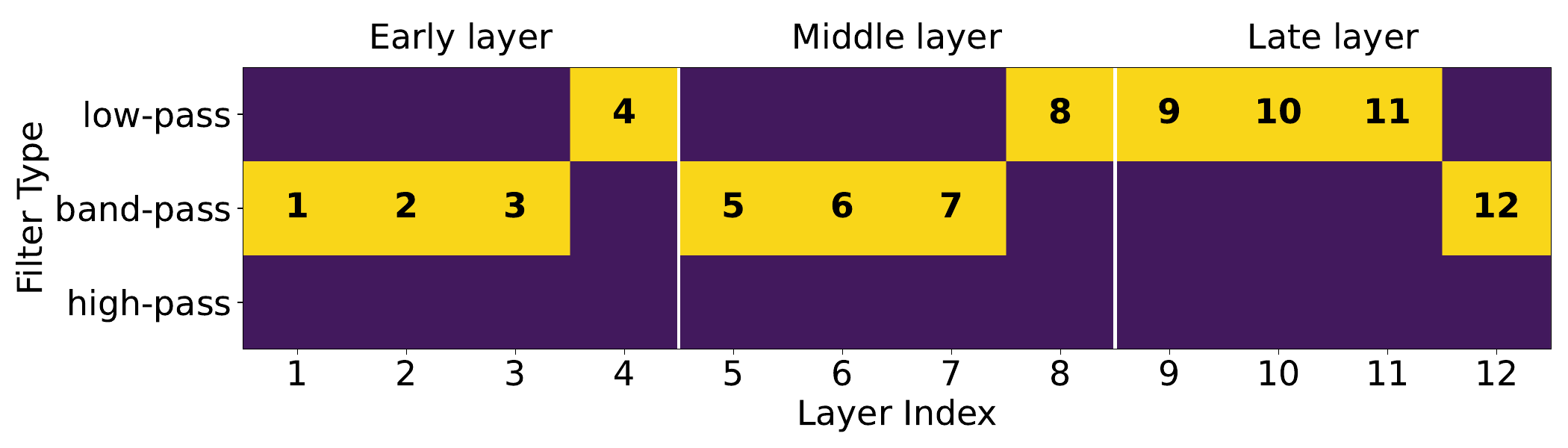} 
     \includegraphics[width=0.5\linewidth]{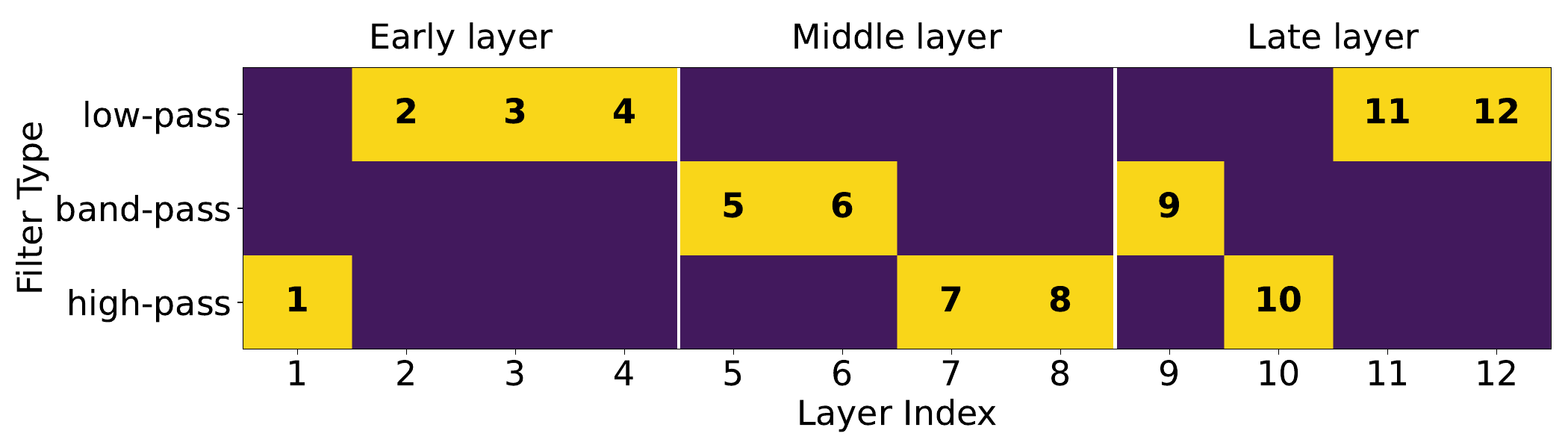}
     \caption{CodeSSM}
     \label{fig: filter_codessm}
    \end{subfigure}
    \begin{subfigure}[b]{0.99\textwidth}
     \includegraphics[width=0.5\linewidth]{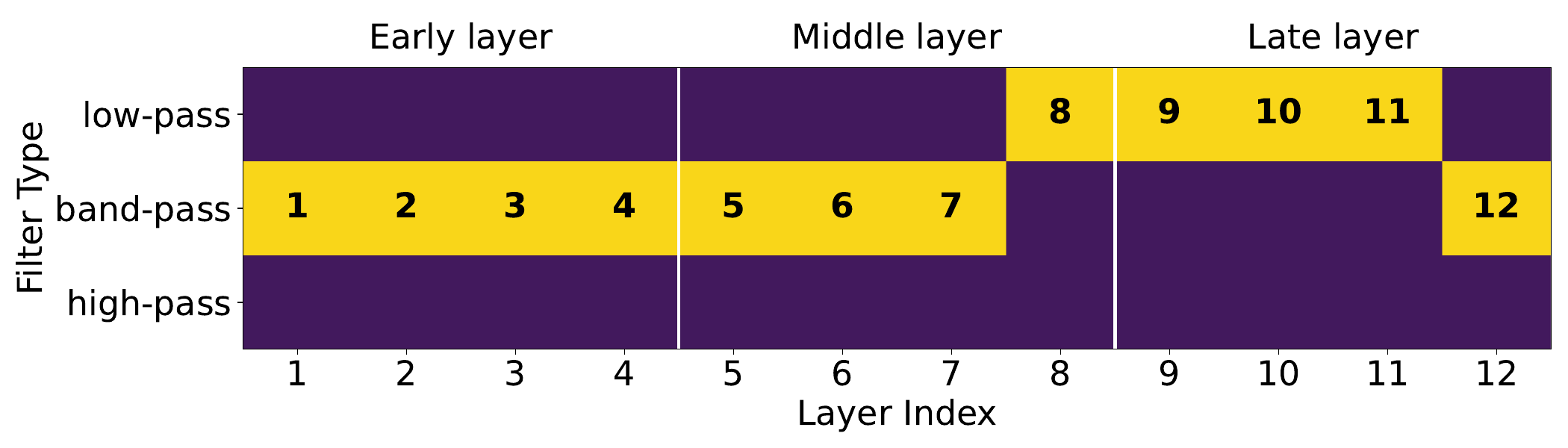} 
     \includegraphics[width=0.5\linewidth]{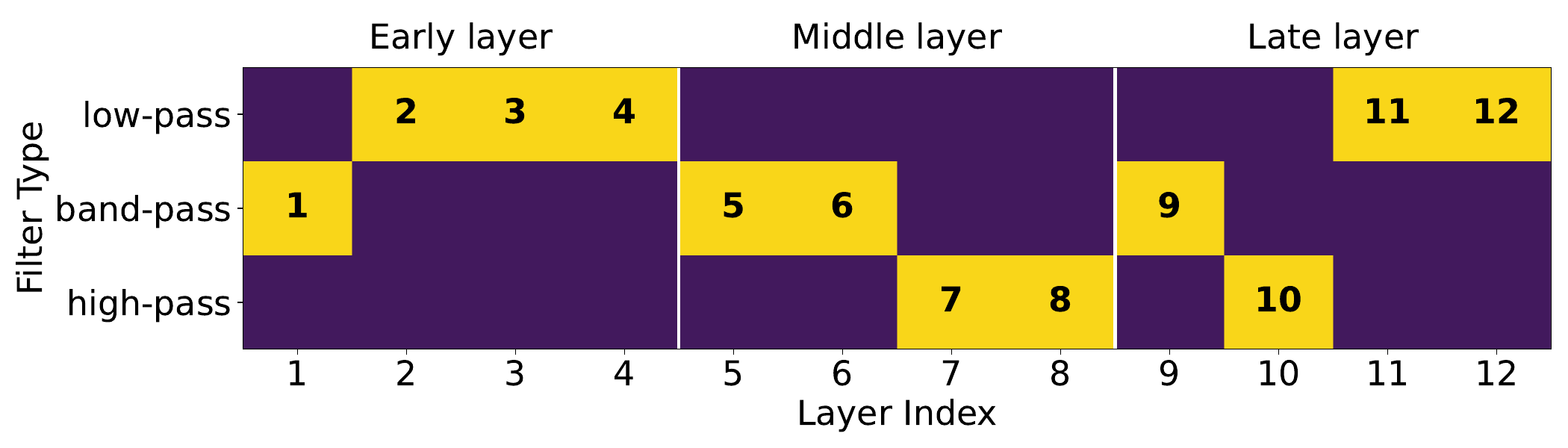}
     \caption{CodeSSM\_sqa}
     \label{fig: filter_codessmsqa}
    \end{subfigure}
     \begin{subfigure}[b]{0.99\textwidth}
     \includegraphics[width=0.5\linewidth]{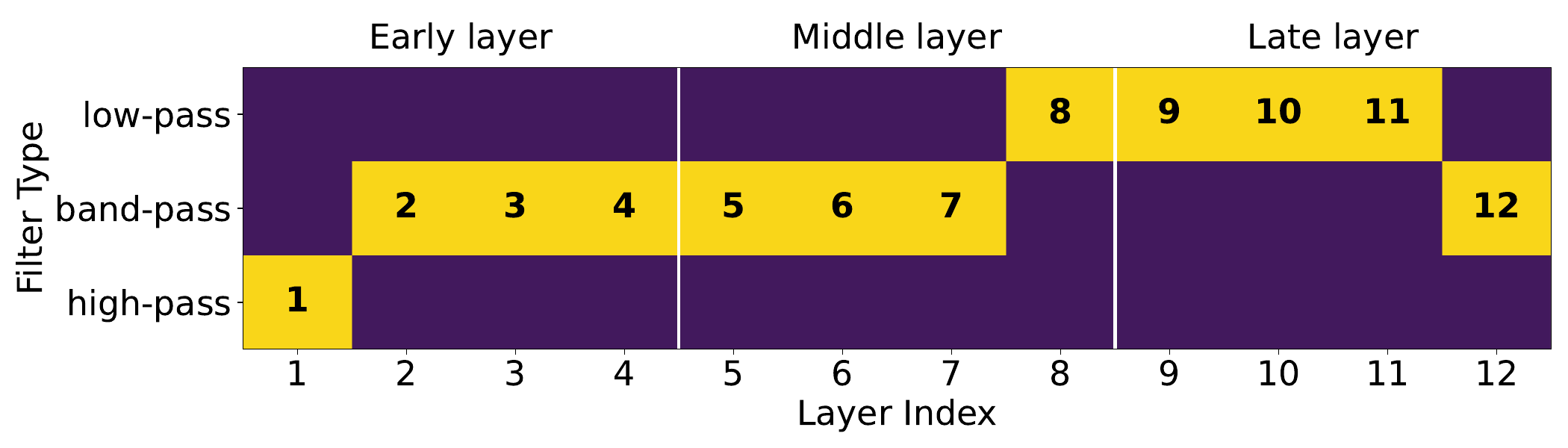} 
     \includegraphics[width=0.5\linewidth]{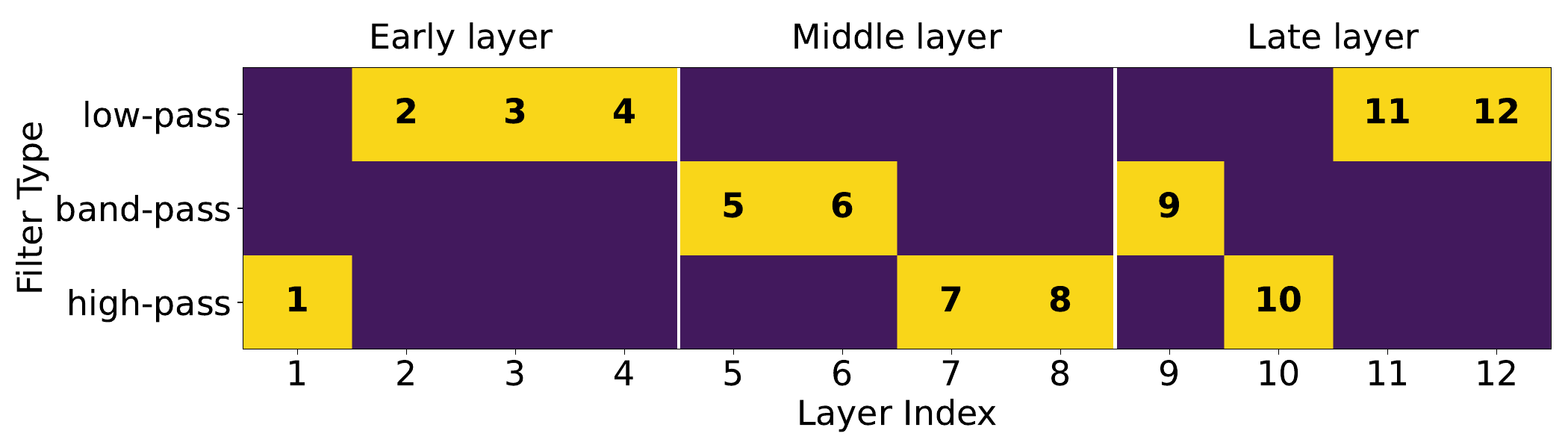}
     \caption{CodeSSM\_typeinf}
     \label{fig: filter_codessmtypeinf}
    \end{subfigure}
    \begin{subfigure}[b]{0.99\textwidth}
     \includegraphics[width=0.5\linewidth]{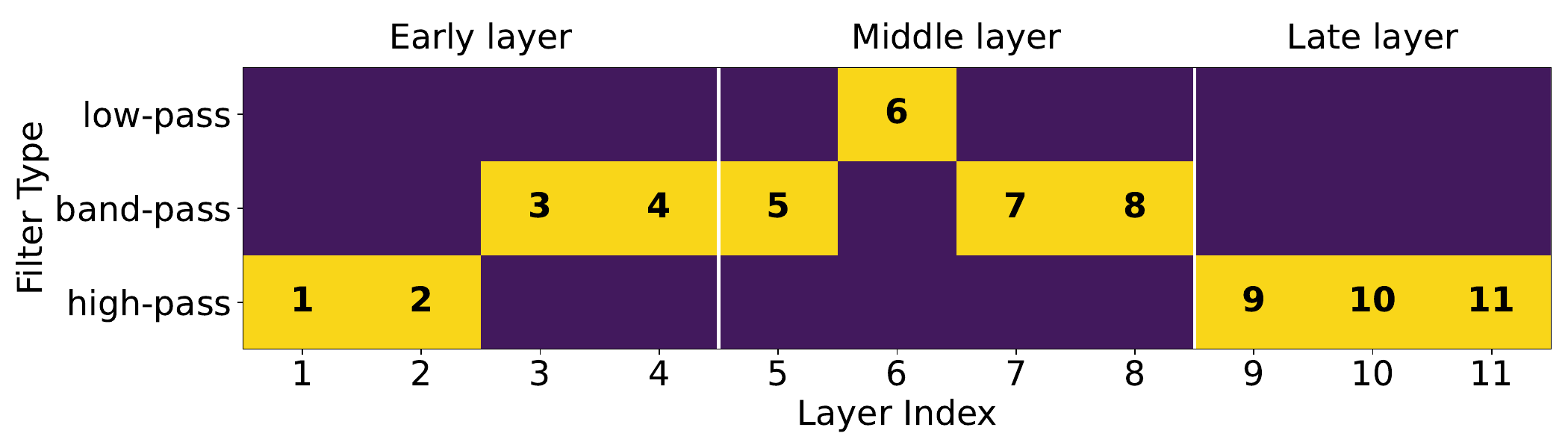} 
     \includegraphics[width=0.5\linewidth]{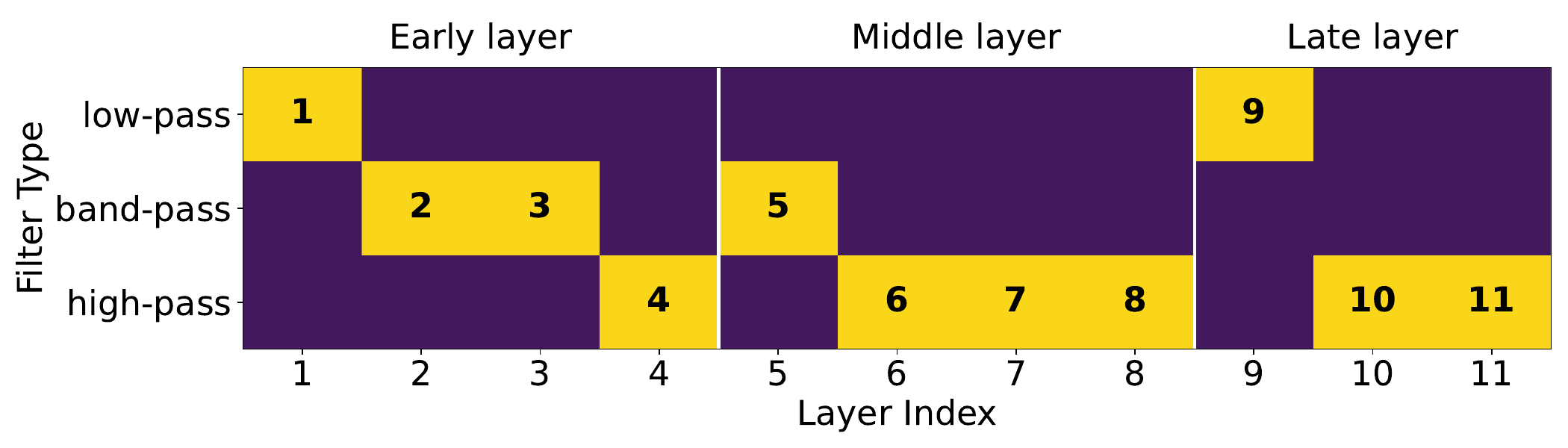}
     \caption{CodeSSM-HF}
     \label{fig: filter_codessmhf}
    \end{subfigure}
  \caption{Layer wise filter classification of forward (left) and backward kernel (right) of CodeSSM and its variants.}  
\end{figure*}
\section{SSM Kernel Analysis}
To systematically analyze the token dependencies captured by state-space models, we propose \textit{SSM-Interpret}, a novel framework that categorizes SSM kernels based on their spectral characteristics. By mapping kernels to the frequency domain, this framework allows us to infer whether a model prioritizes local (high-frequency) or global (low-frequency) token interactions (\cref{fig: code-freq}). We use the framework to analyze CodeSSM which uses S4D \cite{s4d} with a single kernel shared across all channels. The size of the kernel equals the context length. We set the context length to 4096, however, the framework is context length agnostic (see Appendix \ref{kernel-agnostic}). Each layer has two SSM kernels (one forward and one backward as shown in \cref{cssm}). The \textit{SSM-Interpret} is applicable to complex NPLR and DPLR S4 implementations \cite{s4} as well.

\subsection{The SSM-Interpret Kernel Analysis} \label{ssm-interpret}
The core objective of our kernel analysis is to characterize the behavior of SSM kernels in terms of the range of dependency captured. To achieve this, we extract the forward and backward kernels from all 12 layers of CodeSSM and analyze their spectral properties via the Fourier transform. By examining which frequency bands are amplified or attenuated, we classify kernels into low-pass (long-range), high-pass (short-range), or band-pass categories, following established links between spectral behavior and dependency range \cite{ravikumar2026analysislongrangedependency}.%: Low-pass kernels correspond to long-range dependency capture, whereas high-pass kernels emphasize local dependencies.

Prior approaches rely on the dominant frequency (frequency with the highest magnitude) for categorization \cite{ravikumar2026analysislongrangedependency}. However, a kernel may exhibit a sharp peak in one spectral region while the majority of its energy resides elsewhere, leading to misclassifications. To mitigate this, we propose a robust classification strategy based on total spectral energy distribution. %rather than peak magnitude. 
We employ two complementary metrics: Spectral Centroid (SC) \cite{dsp_spectralcentroid} and Low-to-High Frequency Energy Ratio (LHFR) \cite{lhfer}. Reliance on the spectral centroid alone risks mirroring the pitfalls of dominant frequency analysis by ignoring secondary high-energy regions \cite{spectral_centroid_dominantfreq}. However, our empirical evaluation demonstrates that combining the spectral centroid with LHFR yields a robust and perceptually accurate classification of kernel behavior.

\textbf{SC} is the weighted mean of all the frequencies in the frequency domain representation of a signal, where the weights are the magnitude of each frequency component. Often described as the spectrum’s “center of gravity,” the SC identifies the point around which the spectral energy is concentrated. A higher centroid value indicates a concentration of energy towards the high-frequency region. Formally, we calculate the SC as:
\begin{equation}\label{eq: sc}
    \text{SC} = \frac{\sum_{n=0}^{N-1} f(n)\, X(f(n))}
                 {\sum_{n=0}^{N-1} X(f(n)}
\end{equation}
where $n$ denotes the index of the frequency component, $f(n)$ is the frequency at index $n$, and $X(f(n))$ is the magnitude of the Fourier transform at $f(n)$. Based on the computed SC, we classify kernel behavior using the following thresholds, derived from the observed frequency distribution of CodeSSM kernels:
\begin{equation}
\begin{split}
\textbf{Low-pass:} \quad & \text{Centroid} < \frac{1}{3} \cdot 0.5 \;\approx\; 0.16 \\
\textbf{High-pass:} \quad & \text{Centroid} > \frac{2}{3} \cdot 0.5 \;\approx\; 0.33 \\
\textbf{Band-pass:} \quad & \text{otherwise}
\end{split}
\end{equation}
Here, $0.5$ represents the normalized highest representable frequency (in cycles per sample).

\textbf{LHFR} measures the balance of spectral power by comparing the energy concentrated in the lower versus upper frequency bands, i.e., $\text{LHFR} = \frac{E_{\text{low}}}{E_{\text{high}}}$, where $E_{\text{low}}$ is the sum of magnitudes within the bottom 10\% of the frequency spectrum, and $E_{\text{high}}$ is the sum within the top 40\%. This ratio serves as an indicator of whether the kernel focuses on global structure (low-frequency dominance) or local ones (high-frequency dominance). We classify the kernels based on the LHFR as follows:
\begin{equation}
   \begin{split}
 \textbf{Low-pass:} \quad & \text{LHFR} \gg 10 \\
 \textbf{High-pass:} \quad & \text{LHFR} \ll 1 \\
 \textbf{Band-pass:} \quad & \text{otherwise}
\end{split}
\end{equation}
\textbf{Kernel Categorization.}
We combine the SC and LHFR metrics to categorize the SSM kernels into low-pass, high-pass, or band-pass filters. \cref{fig: scnlhfrfilter_codessm} illustrates the resulting classification for CodeSSM. We established the specific thresholds for both metrics through a rigorous qualitative analysis of the kernel spectral profiles of all the models and our chosen values minimizes the outliers. 
Detailed justifications for these criteria are provided in Appendix \ref{sc_LHFR}.

\subsection{Kernel Analysis Results}\label{res_kernel}
\textbf{Why do some layers capture more code properties?}
The Forward and backward kernels of certain CodeSSM layers exhibit complementary frequency responses. For instance, in layer 8, the forward kernel acts as a high-pass filter and the backward kernel as a low-pass filter, indicating that most of the forward kernel’s energy lies in the high-frequency range, whereas the backward kernel concentrates in the low-frequency range. 

This complementary filter behavior suggests that using two directionally distinct kernels enables the model to capture richer contextual information from both past and future tokens as well as the far away and nearby tokens when making predictions.
Notably, layers displaying such complementary forward–backward kernel patterns tend to capture the most code-related properties. For example, layer 8 achieves the best performance on the siblings and edge prediction tasks, while layer 10 performs best on the distance task (see \cref{fig: hiddenrepr_pretr}). Both layers exhibit complementary filter characteristics (layer 8: forward–high-pass, backward–low-pass; layer 10: forward–low-pass, backward–high-pass; see \cref{fig: filter_codessm}).

In contrast, layers where both the forward and backward kernels emphasize the same frequency range tend to perform poorly. For instance, layer 11 of CodeSSM-TypeInf, which exhibits low-pass behavior in both kernels, shows reduced performance on the sibling and DFG tasks.

\textbf{Why does code understanding decrease on the type inference task?}
In the fine-tuned models, the forward kernel behavior in the initial layers shifts toward higher frequencies compared to the pretrained models. For instance, in \cref{fig: filter_codessmtypeinf}, the forward kernels of layers 1 and 4 transition to high-pass and band-pass filters, respectively. Additionally, the frequency response of both forward and backward kernel of layer 1 lies in the same region of the spectrum (high-pass) due to which the kernel misses a wide range of code properties which lie in the other region of the spectrum.
The layers with complementary filter behavior (layer 8 and layer 10 of CodeSSM\_typeinf) still capture more code properties (\cref{fig: hidden repr finetuned_codessm}) but the proportion of code properties captured reduces significantly due to early high frequency shift.

Because the outputs of the forward and backward kernels are multiplied in the time domain (see \cref{cssm}), they become convolved in the frequency domain. When both kernels exhibit high-pass characteristics in the initial layers (as in \cref{fig: filter_codessmtypeinf}), this convolution amplifies the high-frequency response, potentially further biasing the model toward short-range dependencies. Moreover, emphasizing short-range dependencies too strongly in the early layers limits the model’s ability to learn long-range dependencies, as reflected in the drop in accuracy at distances 5 and 6 shown in \cref{fig: distancelayer8910-codessmtypeinf}.

As discussed in \cref{hidden_analysis}, the type inference task depends on both short- and long-range dependencies. Transformers can flexibly adapt to short-range syntactic patterns, but CodeSSM lacks this flexibility due to its use of only two SSM kernels per layer. In \cref{fig: filter_codessmtypeinf}, we observe that CodeSSM shifts toward learning short-range dependencies, yet without sufficient capacity to do so effectively. \citet{verma-etal-2025-codessm} hypothesized that CodeSSM’s decreased performance on the type inference task stems from its inability to capture short-range dependencies. Our filter behavior and hidden-representation (distance 2 and 3 in \cref{fig: distancelayer8910-codessmtypeinf}) analyses are consistent with this hypothesis.

\section{Analysis-Driven Improvements}
Our analysis of hidden representations and SSM kernels has uncovered the strengths and critical limitations in the CodeSSM architecture. Leveraging these insights, we propose targeted architectural modifications which yield significant performance gains across code understanding tasks.

\subsection{Proposed Modifications}
Our analyses reveal that CodeSSM struggles with tasks that require understanding of short- and long-range dependencies due to lack of encoding high-frequency. While hybrid architectures interleaving Transformer layers offer a solution \cite{spade, ren2025samba}, they also have the quadratic complexity of self-attention.

\textbf{CodeSSM-HF.} Given that the limitation is due to the handling of high-frequency information, we instead propose an efficient alternative which we refer as \texttt{CodeSSM-HF}: introducing a dedicated high-frequency path in parallel to the low-frequency SSM path (see \cref{cssm}). This path employs a 1D Convolutional Neural Network (CNN) with a kernel size of 3 \cite{cnn_highfreqbias}, providing a strong inductive bias for short-range dependencies (attending to the current, preceding, and succeeding tokens) analogous to local attention heads in Transformers.
To offset the parameter increase from the CNN, we employ grouped convolutions (group size 8) and reduce the depth to 11 layers. These constraints ensure CodeSSM-HF, has slightly fewer parameters than the original CodeSSM. As shown in Table \ref{div-table}, incorporating a high-frequency path yields consistent performance gains on the SQA and type inference tasks, confirming the critical role of high frequency information across tasks. The performance gain is also observed on the adversarial NLCodeSearch task \cite{codexglue}.

\textbf{Multi-kernel.} In \cref{res_kernel}, we hypothesized that CodeSSM's performance on type inference might be hindered by having just two SSM kernels per layer. To evaluate the hypothesis, we train a model with $1024$ kernels (i.e., equal to the input dimensions)(referred as \texttt{CodeSSM-1024k}). This is similar to previous works on SSMs \cite{s4d, s4}.
%as well as works introducing frequency and phase based positional embeddings in transformers \cite{Rope, fope}. => Move this to Appendix with a more detailed description of similarity between multi-kernel and rope/fope
While having $1024$ kernels improves the performance on NLCodeSearch and type inference, the performance remains same on SQA and the performance on NLCodeSearch is lower than CodeSSM-HF (Table \ref{div-table}).  

To understand the trade-off between increasing capacity and performance, we pretrained CodeSSM with kernels 1, 4, 8, ..., 512, 1024 on a small dataset. The pretraining performance was best for 8 kernels. Subsequently, we 
pretrained a model with 8 kernels (\texttt{CodeSSM-8k}). This variant achieves the best performance on all tasks. In this model, each kernel is shared across 128 dimensions. These results suggest that increasing capacity while maintaining shared
transformation across some dimensions, is optimal for better code understanding.

\subsection{Kernel Analysis of CodeSSM-HF}
Since CodeSSM-HF retains the single-kernel SSM block structure, we directly compare its kernels against the baseline CodeSSM using our \textit{SSM-Interpret} analysis. Figure \ref{fig: filter_codessmhf} shows the kernel classification of CodeSSM-HF (see Appendix \ref{8kernel-analysis8} for analysis of CodeSSM-8k). We observe two distinct spectral shifts in the CodeSSM-HF kernels. 

First, the HF model exhibits a greater prevalence of high-pass filters, particularly in the early layers, unlike CodeSSM. In these layers, we observe a complementary spectral behavior where high-pass filters are paired with low- or band-pass filters. Moreover, unlike CodeSSM kernels, which typically display a single dominant frequency which decays on each side, CodeSSM-HF kernels often possess multiple crests across frequency bands (see Figure %\ref{ind kernels} and 
\ref{more kernels}). This spectral heterogeneity allows early layers to simultaneously capture both long- and short-range dependencies, correcting the “blind spots” identified in \ref{res_kernel}. Consequently, this simple architectural modification enhances the SSM’s ability to model rich token-level dependencies, leading to improved overall performance.

% \begin{figure}[h]
%     \centering
%     \begin{subfigure}[b]{\columnwidth}
%     \includegraphics[width=0.48\linewidth]{figures/individual_kernels/CodeSSM_Layer_2_forward.pdf}
%     \includegraphics[width=0.48\linewidth]{figures/individual_kernels/CodeSSM_Layer_3_backward.pdf}
%     \label{cssm_kernel}
%     \caption{CodeSSM Kernels}
%     \end{subfigure}
%     \begin{subfigure}[b]{\columnwidth}
%     \includegraphics[width=0.48\linewidth]{figures/individual_kernels/hf_Layer_2_forward.pdf}
%     \includegraphics[width=0.48\linewidth]{figures/individual_kernels/hf_Layer_3_backward.pdf}
%     \label{all hf_kernel}
%     \caption{CodeSSM-HF kernels}
%     \end{subfigure}
%     \caption{SSM Kernel of layer 2 forward (left) and layer 3 backward (right). The dominant frequency is represented with a circle (red) and the spectral centroid is represented by a triangle (green)}
%     \label{ind kernels}
% \end{figure} 
%=> removing this from the main paper to fit in page limit

Second, the final two layers of CodeSSM-HF shift predominantly toward high-pass behavior. This contrasts with CodeSSM, where later layers show low- or band-pass characteristics. This shift toward high-pass behavior in the later layers (reflecting a focus on short-range dependencies) is intuitively consistent with effective hierarchical representation learning: after the early layers perform extensive global mixing through the SSM convolution kernel (low-pass), long-range context is effectively “localized” into the hidden states. Consequently, the deeper layers only need to model short-range dependencies.

\begin{table}[t!] \centering
\caption{Results on NLCodeSearch, SQA and Type Inference tasks as mean over 3 different seeds (Standard Deviation in subscript). The best performance is in bold.}
\label{div-table}
\begin{center}
\begin{tiny}
\begin{sc}
\begin{tabular}{lccccccr}
\toprule
Model & NLCodeSearch & SQA & Type Inference \\
 & (MRR) & (MRR) & (F1) \\
\midrule
CodeSSM  & $25.31_{0.51}$ & $76.02_{0.08}$ & $59.24_{0.45}$ \\
CodeSSM-hf  & $30.26_{0.37}$ & $78.33_{0.10}$ & $60.04_{0.36}$\\
CodeSSM-8k & $\mathbf{31.30}_{0.25}$ & $\mathbf{79.80}_{0.25}$ & $\mathbf{61.22}_{0.22}$\\
CodeSSM-1024k & $28.19_{0.41}$ & $76.01_{0.12}$ & $60.38_{0.39}$ \\
\bottomrule
\end{tabular}
\end{sc}
\end{tiny}
\end{center}
\vskip -0.1in
\end{table}

\section{Conclusion}
In this work, we presented the first systematic analysis of hidden representations and convolution kernels in multi-layer SSMs. Our empirical results reveal a nuanced landscape: while SSMs fundamentally outperform Transformers in capturing global code properties, they exhibit  weaknesses in modeling short-range dependencies, particularly in tasks like type inference. To diagnose this failure of SSMs, we proposed a novel interpretability framework to analyze the convolutional kernel of SSMs, which showed a shift towards short range dependency understanding during fine-tuning. Guided by these insights, we introduced architectural refinements that significantly narrow this performance gap. 

We anticipate that the introduced framework will serve as a foundation for future interpretability research. Additionally, with the architectural improvements SSMs can be scaled and adapted for generative tasks in future work.

% In this work, we presented the first systematic analysis of hidden representations and convolution kernels in multi-layer SSMs. The hidden representation analysis revealed that while a pretrained SSM model is significantly better at capturing code structure compared to transformers, SSM is unable to retain some of these structure when fine-tuned. SSM is especially prone to forgetting code structures on tasks that require both long and short range dependencies. To diagnose this failure of SSMs, we proposed a novel interpretability framework to analyze the convolutional kernel of SSMs. The kernel analysis showed a shift towards short range dependency understanding during fine-tuning. Guided by these insights, we introduced architectural improvements that significantly improve the performance of CodeSSM. 

%Beyond these model improvements, our \textit{SSM-Interpret} analysis establishes a rigorous methodology for dissecting the spectral behavior of state-space architectures. 

\section*{Limitations}
Broadly, our work has the following limitations.

First, the SSM-interpret framework considers several threshold values for classifying the kernel as low-pass, high-pass, or band-pass, and for calculating LHFR. Although these threshold values are empirically validated, they may not be optimal for some kernels or other models. Thus, future research can refine classification thresholds for the spectral centroid and the low-to-high frequency energy ratio by accounting for the kernel’s frequency response, thereby improving the robustness of frequency-based kernel classification.

Second, we did not perform the analysis of generative models. Although there are no S4D-based generative models, our findings show that this absence may not be accidental. Similar to type inference, generative tasks involve a large vocabulary and require modeling both short- and long-range dependencies simultaneously. Thus, the architectural modifications we proposed can guide future research in developing S4/S4D-based generative models. 

Third, we performed the hidden representation analysis for lengths up to 8k. Currently, models are also being evaluated for lengths up to 32k. However, we limited our analysis to 8k due to resource constraints and the small pre-training context of CodeSSM and RoCoder (256). Our analysis still covers a 32x length extrapolation. Based on our hypothesis and observations, CodeSSM and its variants will show significantly better code understanding with length extrapolation than transformer models.

\section*{Ethical Considerations}
The aim of our work is to analyze and understand the success and failure of SSMs with respect to code properties. The analysis does not raise any ethical risks. In our work, we have used models whose weights are publicly available and are trained on publicly available datasets. We have also trained models on publicly available data. It is not clear whether the models can leak personally identifiable information if present in the pretraining dataset, but we acknowledge that such potential risk exists. 
%for CodeSSM and upto 2k for RoCoder as RoCoder consumes very high memory and it was not possible to complete this with the resources we have

%\section*{Acknowledgments}

\bibliography{custom}

\appendix

\section{Additional Related Works} \label{related mamba}
\begin{figure*}[!]
\centering
     \includegraphics[width=0.99\linewidth]{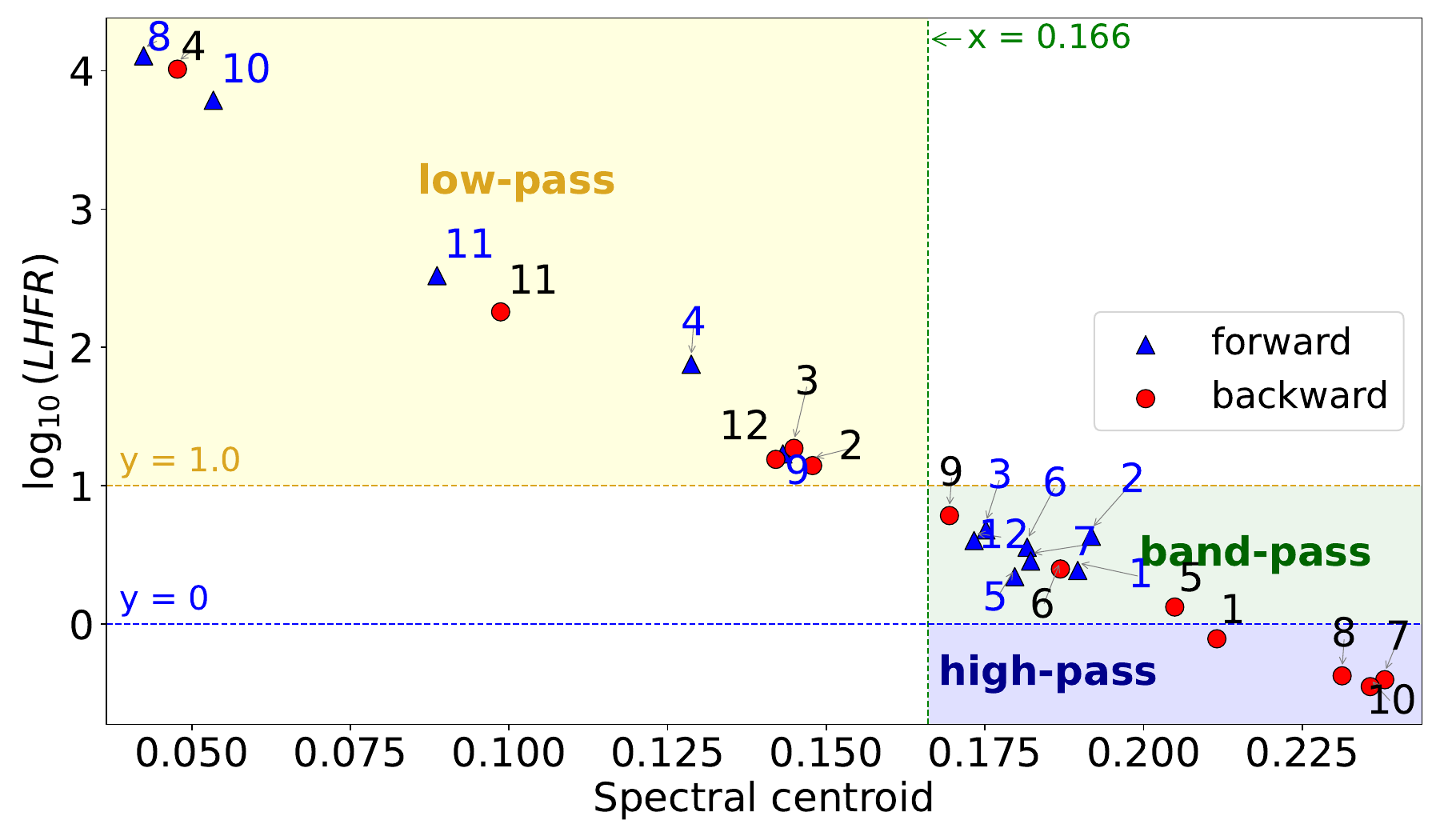} 
     \caption{Filter classification of each layer of the CodeSSM on the basis of spectral centroid and LHFR.}
     \label{fig: scnlhfrfilter_codessm}
\end{figure*}   

\textbf{Frequency Domain Analysis.}
%Previous works on frequency-domain analysis have mainly focused on feature extraction in convolutional neural networks and on filter analysis that depends on specific inputs. \citet{CNN_ztransform} evaluate the kernels of a convolutional neural network from a spatial frequency perspective using the Z-transform where they classify the kernels based on the median of the amplitude spectrum. 

%\citet{frequencyregularization} shows that regularization enforces a strong spectral bias towards low frequencies. \citet{s4nd} filters out the high-frequency components to prevent aliasing in a multi-dimensional state space model. \citet{cnn_highfreqbias} shows that CNNs tend to learn more high and mid-frequency patterns than low-frequency ones. Frequency-domain analysis has also been used to understand length generalization in transformers \cite{fope}, but it is limited to positional embedding analysis only. In contrast to these works, ours is the first to develop a framework for frequency-domain analysis of SSM kernels. The framework can also be used to analyze convolutional neural networks.

Prior research on frequency-domain analysis has primarily focused on feature extraction within Convolutional Neural Networks (CNNs) or input-dependent filter analysis. For instance, \citet{CNN_ztransform} evaluated CNN kernels via the Z-transform to classify them based on their amplitude spectrum medians. Additionally, a more recent work \cite{ravikumar2026analysislongrangedependency} focuses on the analysis of CNNs and single layer SSMs using time and frequency domain analysis. The classification criteria used by \citet{ravikumar2026analysislongrangedependency} is based on dominant frequency. The work also shows how long-range dependency understanding varies across different architectures of SSMs. However, we have talked about the disadvantages of relying on dominant frequency in Section \ref{ssm-interpret}. 

In the context of generalization, \citet{frequencyregularization} demonstrated that regularization enforces a strong spectral bias towards low frequencies. Similarly, regarding architectural constraints, \citet{s4nd} filtered out high-frequency components to prevent aliasing in multi-dimensional state space models, while \citet{cnn_highfreqbias} found that CNNs exhibit a tendency to learn high- and mid-frequency patterns over low-frequency ones. Frequency-domain analysis has also been applied to Transformers to understand length generalization \cite{fope}, though this work was restricted to positional embeddings. In contrast to these approaches, ours is the first to develop a framework specifically for the frequency-domain analysis of SSM kernels, a framework that remains applicable to CNNs as well.

%Recent works analyzing selective SSMs are discussed in detail in \ref{related mamba}.
\textbf{Explainability in SSMs.}
Numerous studies have conducted both theoretical and empirical analyses of selective state space models (SSM), particularly Mamba. These investigations have identified several notable limitations of Mamba. For example, \citet{mamba2025achilles} demonstrate that Mamba relies on non-linear convolution to retrieve relevant information, and that this non-linearity introduces an asymmetry bias, which impedes the model's ability to recognize symmetrical patterns and relationships. The authors propose straightforward mitigation strategies, such as incorporating residual paths or gating around the convolution, and report promising improvements in SSM models. However, their analysis is restricted to synthetic tasks. \citet{jafari2024mambalrp} apply Layer-wise Relevance Propagation (LRP) to Mamba and identify specific architectural components responsible for generating unfaithful explanations. They further introduce more interpretable variants of Mamba, termed MambaLRP. Additional studies have employed attention-based explainability techniques for Mamba \cite{explainingmoderngatedlinearrnns}. \citet{hiddenattention_mamba} argue that Mamba can be interpreted as an attention-based model and provide a theoretical comparison between the mechanisms underlying Mamba and those of attention. In their analysis, they reformulate Mamba layers as self-attention and conduct attention-based evaluations. Furthermore, \citet{repeatmetransformersbetter} and \citet{Mamba_retrival} report that Mamba exhibits difficulties with input copying tasks.

% Several works have done theoretical and empirical analysis of selective SSM (Mamba). These works highlight some significant limitations of Mamba. \citet{mamba2025achilles} show that Mamba tend to rely on non-linear convolution for retrieving relevant information and non-linear convolution introduces an asymmetry bias which makes it difficult for the model to recognize symmetrical patterns and relationships. They also propose simple mitigation (e.g., residual paths / gating around the convolution) and show good potential of improving SSM models. However, this analysis is limited to synthetic tasks.  \citet{jafari2024mambalrp} applies Layer-wise Relevance Propagation (LRP) to Mamba and identify specific components in the Mamba architecture which cause unfaithful explanations. They also propose more interpretable variants of Mamba - MambaLRP. Several works have used attention based explainability techniques for Mamba \cite{explainingmoderngatedlinearrnns}. \citet{hiddenattention_mamba} show that Mamba can be viewed as attention-based models and theoretically compared the underlying mechanism in Mamba to that of attention. For the analysis, they reformulate Mamba layers as self attention and perform attention analysis. Additionally, \citet{repeatmetransformersbetter} and \citet{Mamba_retrival} show that Mamba struggles with input copying tasks. 

\citet{qi-etal-2024-smr} employed event-triggered control to conduct a theoretical analysis of the stability of SSMs. \citet{stablessm} demonstrated that SSMs lacking reparameterization are subject to the "curse of memory," a limitation also observed in Recurrent Neural Networks. They introduced a novel reparameterization that enhances model performance compared to the approach in \citet{s4}. \citet{stateillusion} established that SSMs, like Transformers, are incapable of representing complex state-tracking problems. The authors further showed that the expressive power of SSMs can be increased by utilizing input-dependent transition matrices. Additionally, \citet{nishikawa2025state} found that SSMs, when combined with specific non-linearities, exhibit dynamic token selection abilities comparable to those of Transformers.

% \citet{qi-etal-2024-smr} used event-triggered control to perform theoretical analysis of stability of SSMs. \citet{stablessm} showed that SSMs without any reparameterization suffers from "curse of memeory" similar to Recurrent Neural Networks. They proposed a new reparameterization which improves the model performance over the reparametrization used in \citet{s4}. \citet{stateillusion} proved that SSMs, like transformers, cannot express hard state-tracking problems. The authors also showed that the expressive power of SSMs can be improved using input-dependent transition matrices. \citet{nishikawa2025state} showed that SSMs combined with specific non-nonlinearities have dynamic token selection abilitis similar to transformers.

\begin{figure*}[h!]
\centering
     \includegraphics[width=0.99\linewidth]{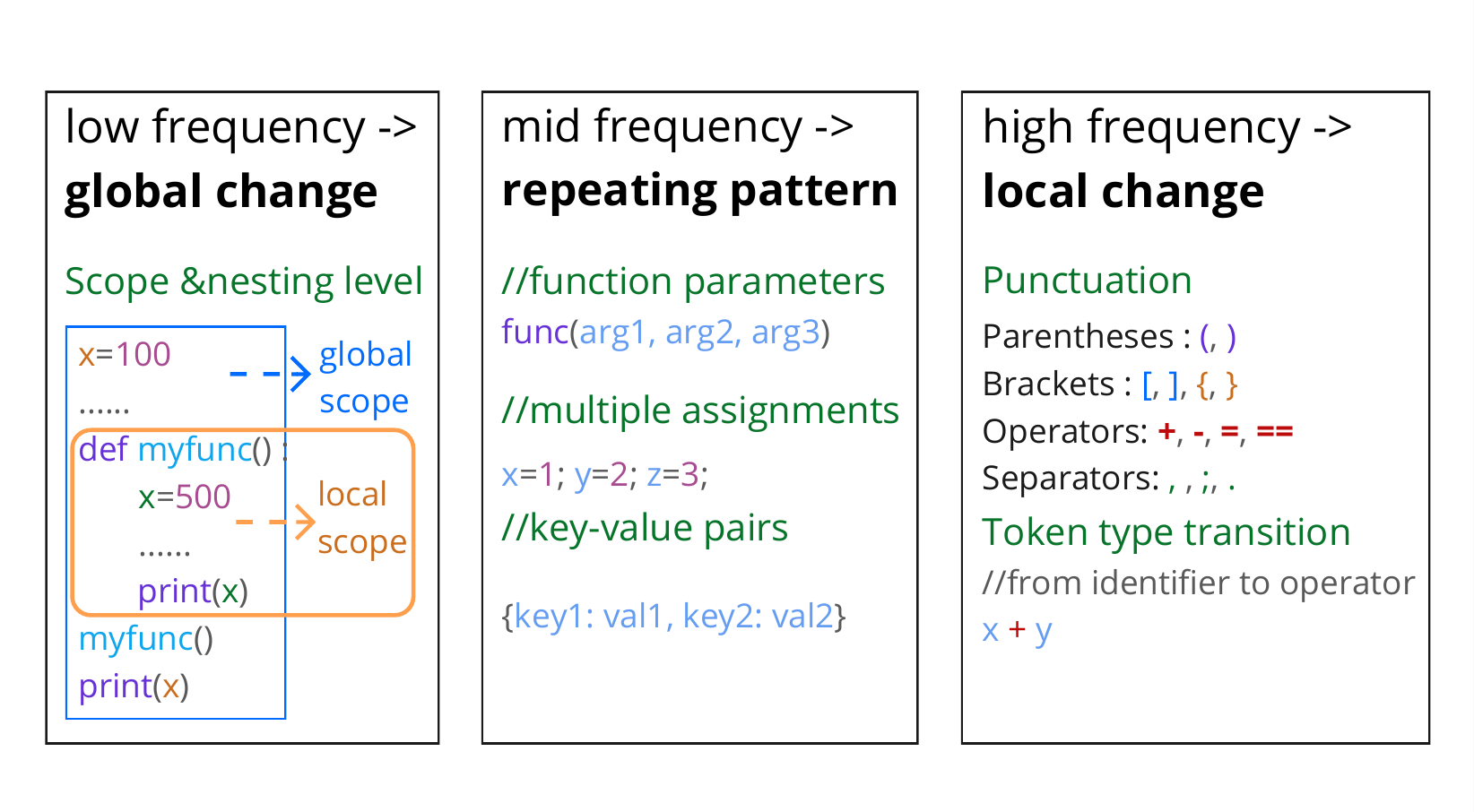} 
     \caption{Relationship between frequency ranges and code patterns.}
     \label{fig: code-freq}
\end{figure*}  

\section{Kernel Analysis}
\subsection{Kernel Classification}\label{sc_LHFR}
We analyze both the forward and backward kernels of CodeSSM across all 12 layers, classifying their behavior based on the frequency ranges they amplify or attenuate. Kernels are categorized as low-pass, high-pass, or band-pass, reflecting whether they primarily capture long-range dependencies or focus on local interactions \cite{ravikumar2026analysislongrangedependency}. Low-pass kernels correspond to long-range dependency capture, whereas high-pass kernels emphasize local dependencies. The relationship between frequency and code token representation is illustrated in Figure \ref{fig: code-freq}.

 In the kernel analysis, we characterize the frequency response of the filters based on the concept that originates from signal processing. The kernel weights represent the contribution of the token at each time step in the sequence to the current token prediction. Since these weights are learned parameters rather than data samples, we compute the Fourier Transform of the kernel weights using a sample rate fs = 1.0, meaning one sample per unit time. Because frequency components are normalized, the resulting frequencies are relative rather than absolute. We therefore measure the relative energy distribution across frequencies to classify filters.

In traditional signal analysis, the cutoff frequency is used for non‑normalized responses where frequency is measured in cycles per second. However, since our kernel frequencies are normalized, a specific cutoff frequency is not meaningful. Instead, we use the spectral centroid and the Low-to-High Frequency Ratio (LHFR) as proxy measures for cutoff frequency.
While LHFR can depend on the choice of cutoff or median frequency, we found that using the median often introduced numerous outliers. To address this, we qualitatively analyzed the kernels and experimented with multiple threshold values. The most stable results—yielding no outliers across all CodeSSM model kernels—were achieved by defining low- and high-frequency energy thresholds at 10\% and 40\%, respectively.

We also performed multiple ablation studies varying the spectral centroid and LHFR thresholds to refine the classification criteria. For the CodeSSM model and its variants, the threshold values reported in the paper provided the most consistent and robust classification, minimizing outliers across all experiments. The final kernel classification results using the selected criteria are shown in Figure \ref{fig: scnlhfrfilter_codessm}.

\subsection{Effect of Kernel Length}\label{kernel-agnostic}
The Spectral Centroid (SC) and LHFR are computed over normalized frequencies and are invariant to absolute kernel length. We observe a change of less than 0.8\% on SC when the kernel length is increased from 256 to 4096. The filter classifications are therefore stable across sequence lengths.

Importantly, the kernel classification remains the same for kernels of size 4096 and 256. Thus, the same conclusions hold for smaller or larger kernels under a given training setup. This means that SSM-Interpret generalizes to longer-context scenarios: as sequence length scales, the metrics remain meaningful and the framework remains applicable without modification.

The filter classification in paper is on kernel length 4096.% We provide classification with kernel length of 256 here.

\subsection{Spectral Behaviour of SSM Kernels} \label{Spectral}
In addition to the filter classification of the kernels, we also show the spectral power of the kernels as a heatmap in \cref{fig: heatmap}. The layers with warm colors (high spectral power) at lower frequencies and cool colors (low spectral power) at higher frequencies show low pass behavior. A constant cool color (low spectral power) across low to high frequencies shows high pass behavior. A slight change in color (small change in spectral power) from low to high frequencies show band pass behavior. 
We can also see a lot of warm colors in CodeSSM-8k at lower frequencies which shows a higher spectral power towards lower frequencies. Additionally, we observe a lot of variation in spectral power at different frequencies, which shows that each kernel focus on different characteristics of code.

\subsection{Kernels of CodeSSM-8k}\label{8kernel-analysis8}
CodeSSM-8k is the best performing CodeSSM variant and has 8 kernel per layer. We studied the kernels of this model. The kernels for layer 1 are shown in \cref{fig: kernel_8} and mean of all 8 kernels of each layer is shown in \cref{fig: heatmap}. Some characteristics of these kernels stand out.

First, the kernels model complex relations across tokens, similar to CodeSSM-HF model. Second, we observe redundancies and range shifts. For example, the third and fifth forward kernels capture similar relations. But the third kernel encode it at a shorter range (higher spectral centroid) while the fifth kernel encode similar relations among tokens at longer distances. Similarly, the third and fifth backward kernels encode similar relations. But the third one allocates higher magnitude to lower frequency.

With just 8 kernels, the SSM block is able to model complex token relations. Increasing the number of kernels can result in more redundancies and does not contribute to the performance. Additionally, with significantly more kernels (for example, 1024), there is a higher chance of out of phase kernels in forward and backward paths which can cancel out the encoded token relations when the hidden representation of forward and backward paths are multiplied (equivalent to convolution in frequency domain). We believe this is the reason for lower performance of CodeSSM-1024kernel. 

% The kernels are classified as low pass, band-pass and high-pass based on the criteria discussed in \cref{ssm-interpret}. However the frequency response of the kernel is continuous and in order to under We also plot the spectral power of the 

\subsection{Regularization in CodeSSM}
The low- and high-pass filter behaviors of the SSM kernel also help explain other observations about CodeSSM \cite{verma-etal-2025-codessm}. Prior studies have reported a high-frequency bias in CNNs \cite{cnn_highfreqbias}, but our analysis finds no such bias in CodeSSM. Regularization techniques such as dropout and L2 have been shown to mitigate high-frequency bias \cite{frequencyregularization}. However, since the SSM in CodeSSM is not inherently biased toward high frequencies, such regularization is unnecessary. Indeed, \citet{verma-etal-2025-codessm} report that incorporating dropout in CodeSSM actually decreases performance.

\begin{figure*}[!]
\centering
     \includegraphics[width=0.99\linewidth]{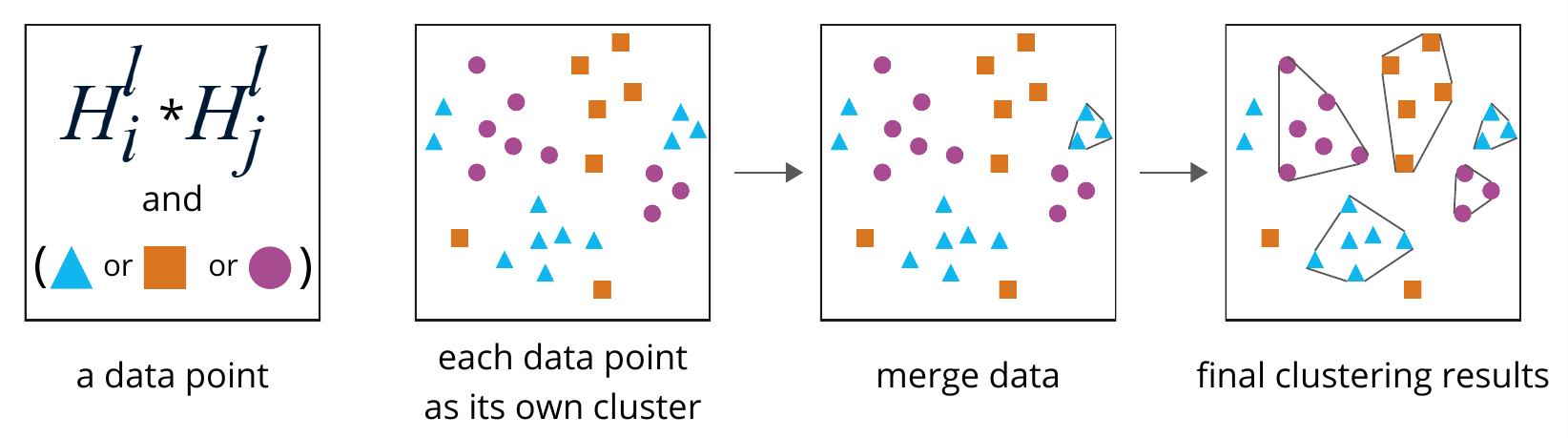} 
     \caption{Visual representation of DirectProbe.}
     \label{fig: dp}
\end{figure*}  

\section{Hidden Representation Analysis} \label{app: dp}
\subsection{DirectProbe}
DirectProbe takes as input a dataset of token representation pairs, each annotated with a relational label (e.g., the existence of an edge in a Data Flow Graph). The algorithm employs an agglomerative clustering approach: initializing each pair as a distinct cluster, it iteratively merges the nearest cl2usters, subject to the constraint that the convex hull of the merged cluster must not intersect with the convex hulls of clusters having different labels.
We provide a visual illustration of DirectProbe in Figure \ref{fig: dp}. Each data point in the input data consists of hidden representation $H^l_i$ and $H^l_j$ for tokens $i$ and $j$ at the output of layer $l$ and a label. As shown in Figure \ref{fig: dp}, DirectProbe starts with each hidden representation point as a separate cluster. Then nearest clusters with the same labels are merged provided the convex hull of the merged cluster does not overlap with any other cluster. The process results in non-overlapping clusters with each cluster having a single label. 

We perform the analysis with three tasks:

\textbf{Distance prediction.} The labels are distance in the AST and range from 2 (minimum possible distance in the tree) to 6. We ignore any token pairs with a distance longer than 6. The clustering is done over difference of the hidden representation.

\textbf{Siblings prediction.} The labels are $siblings$ and $not siblings$. The clustering is done over the concatenated hidden representations of the two tokens.

\textbf{Edge prediction.} The labels are $no edge$, $comes from$ and $computed from$. The clustering is done over the concatenated hidden representation of the two tokens.

On each task, we run the analysis once for each layer of the model.

\subsection{Models} \label{models}
We focus on encoder-only architectures as they have been shown to capture code syntax and semantics more effectively than significantly larger decoder-only counterparts \cite{anand-etal-2024-critical}. RoCoder adapts the BERT architecture by replacing absolute positional embeddings with Rotary Positional Embeddings (RoPE) \cite{Rope} to enable length generalization. CodeSSM is built upon the BiGS architecture \cite{bigs}, with its layer structure detailed in Fig. \ref{cssm}.

\section{Length Extrapolation} \label{length-extra}
\begin{figure*}[t]
\centering
    \begin{subfigure}[b]{0.99\textwidth}
     \includegraphics[width=0.33\linewidth]{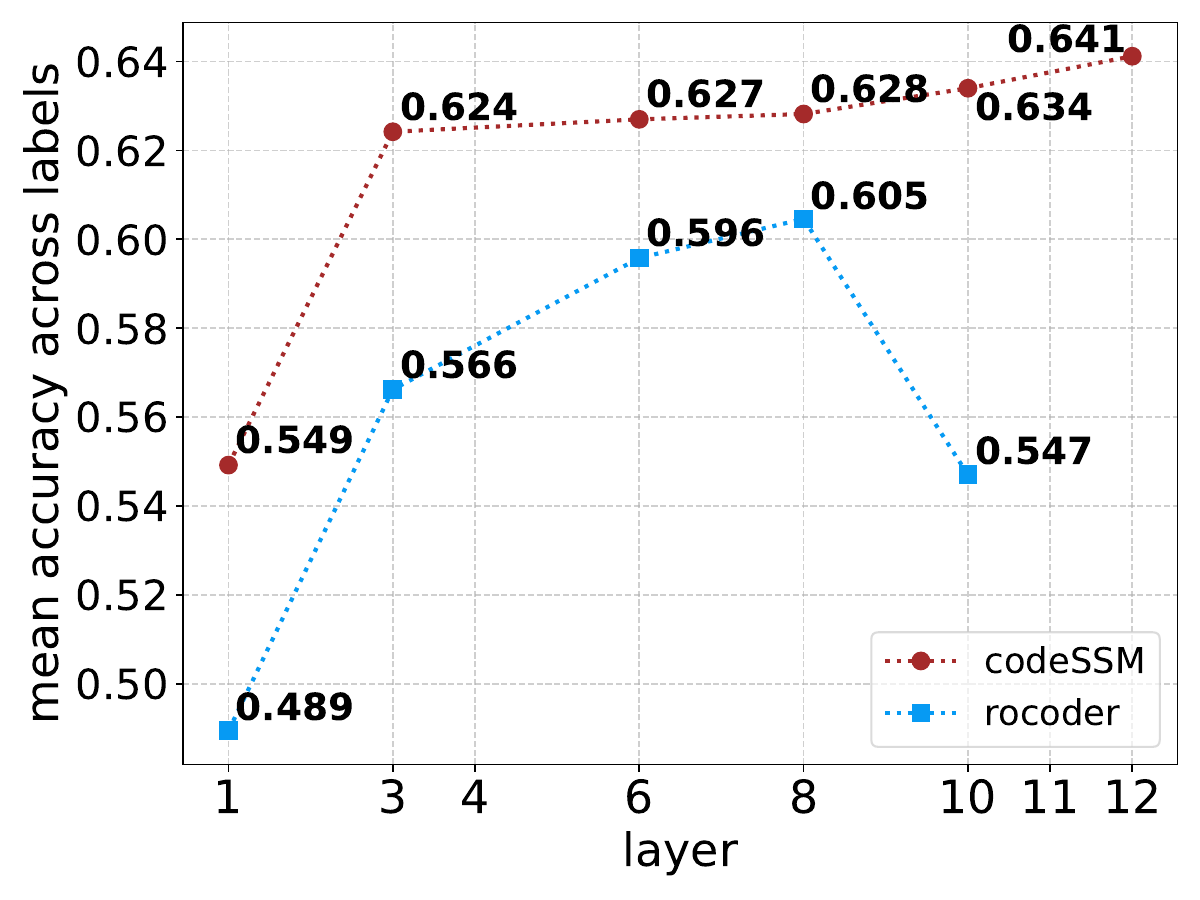} 
     \includegraphics[width=0.33\linewidth]{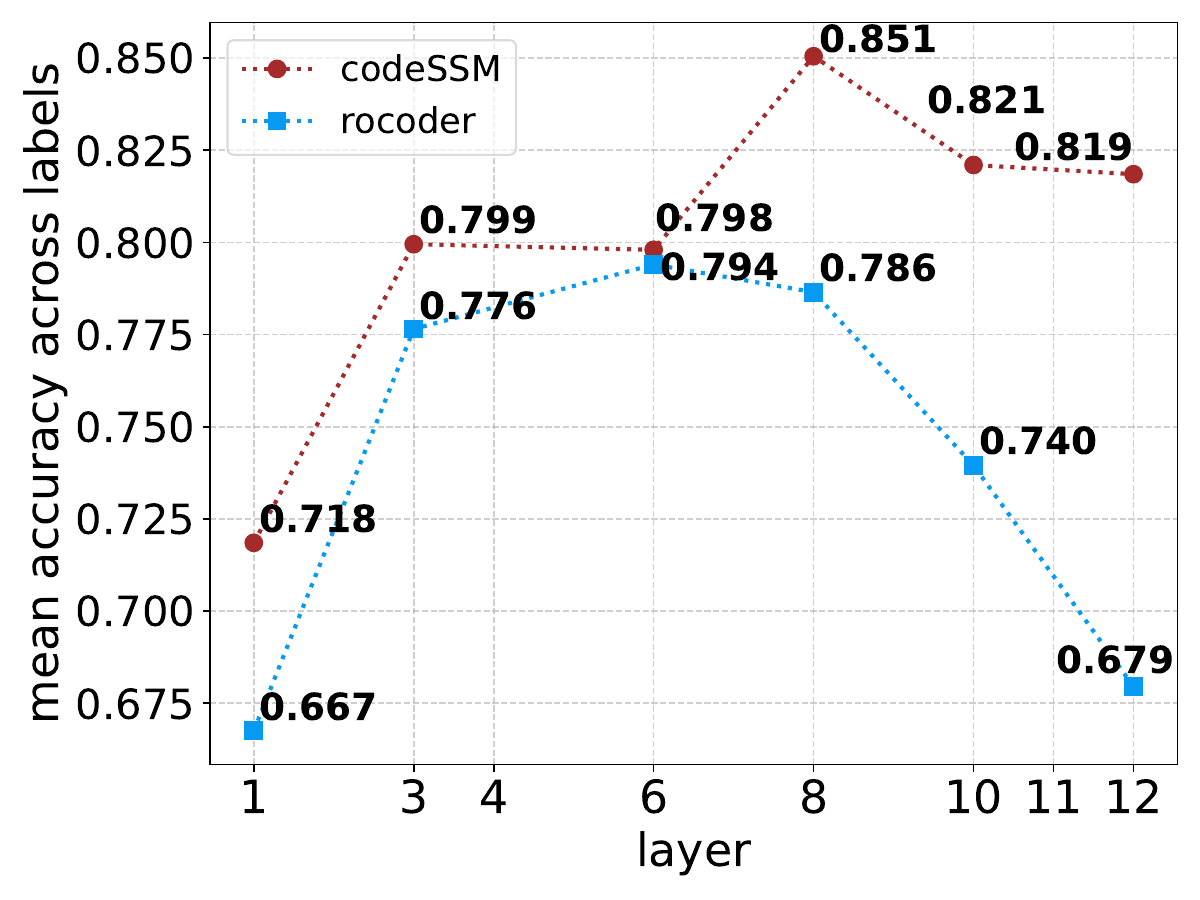}
     \includegraphics[width=0.33\linewidth]{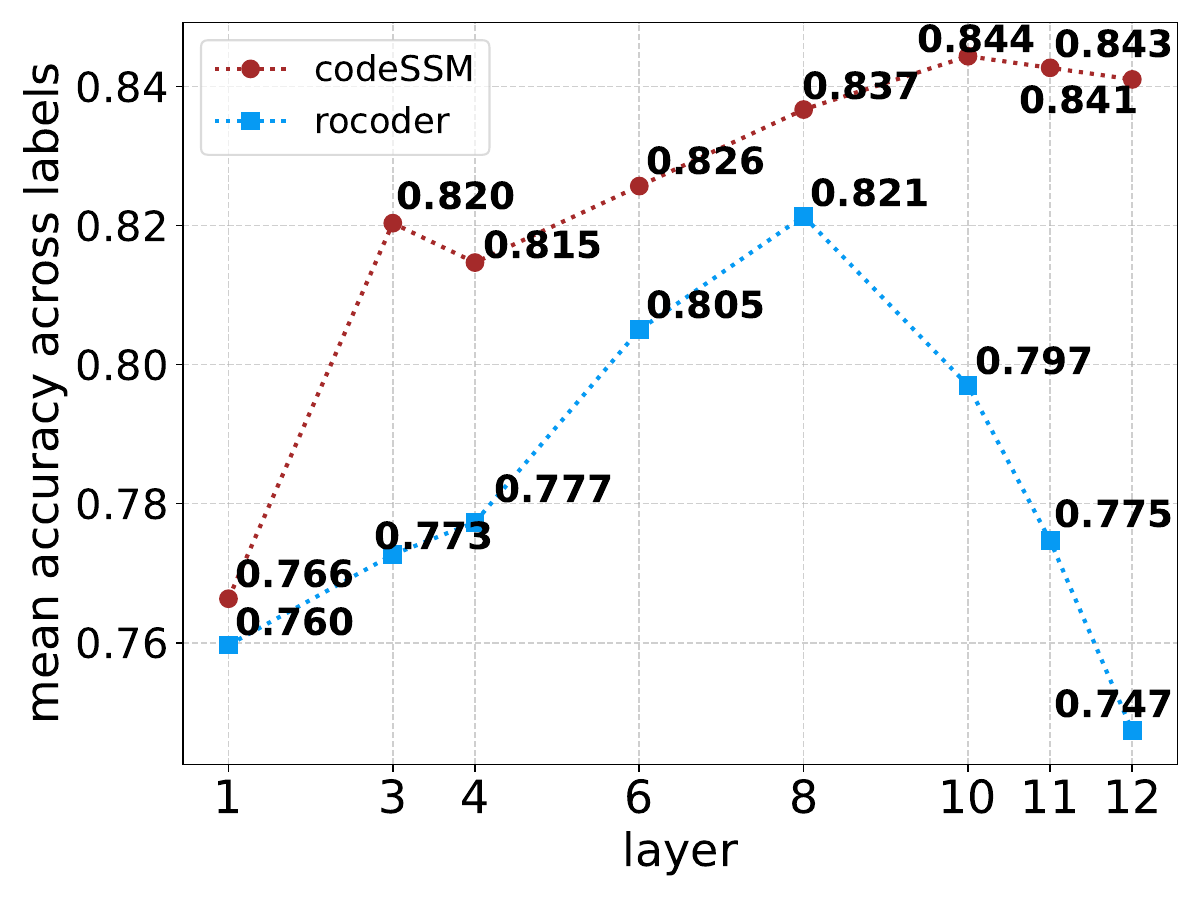}
    \end{subfigure}
    \caption{Comparison of hidden representation of CodeSSM and RoCoder on distance (left), siblings (center) and edge (right) prediction tasks for medium lengths.}
    \label{fig: hidden1k-2k}
\end{figure*}

\begin{figure*}[t]
\centering
    \begin{subfigure}[b]{0.99\textwidth}
     \includegraphics[width=0.33\linewidth]{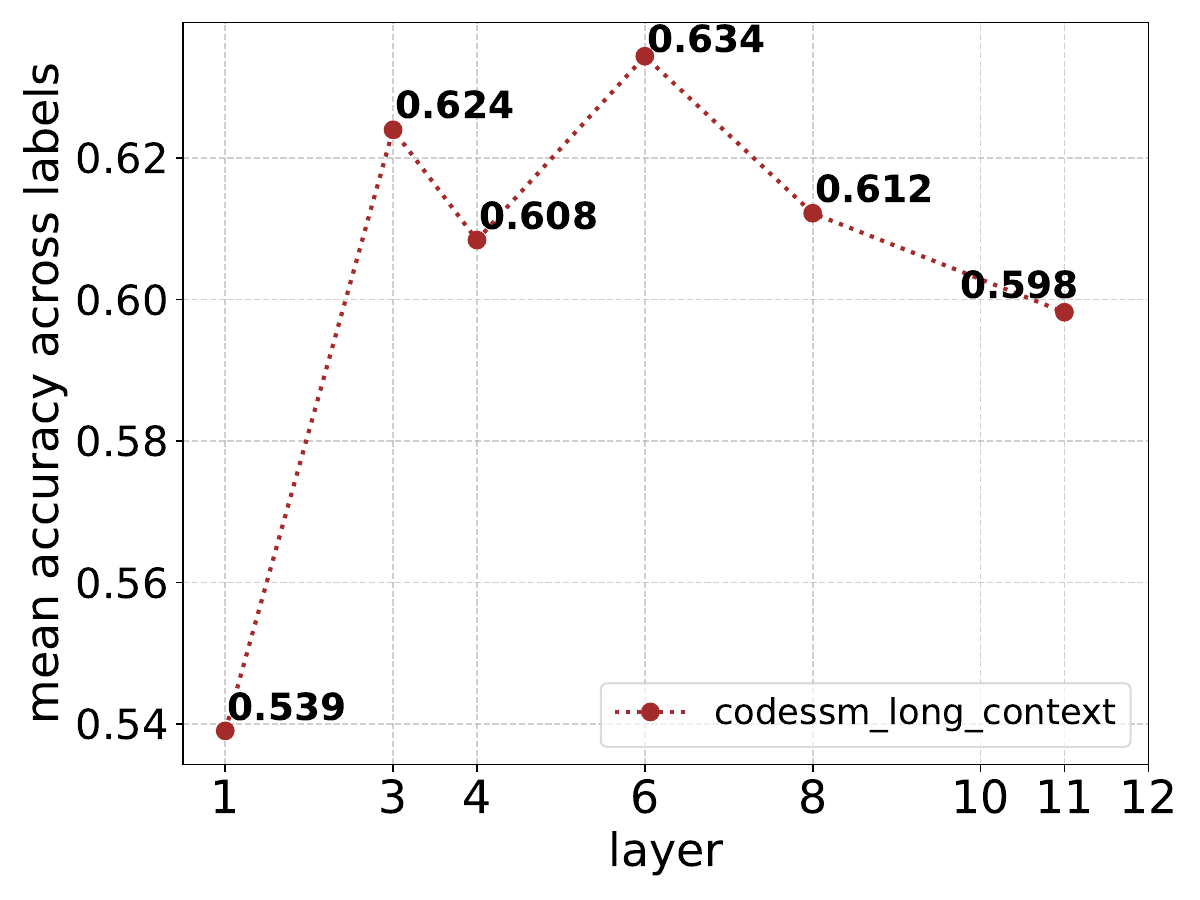} 
     \includegraphics[width=0.33\linewidth]{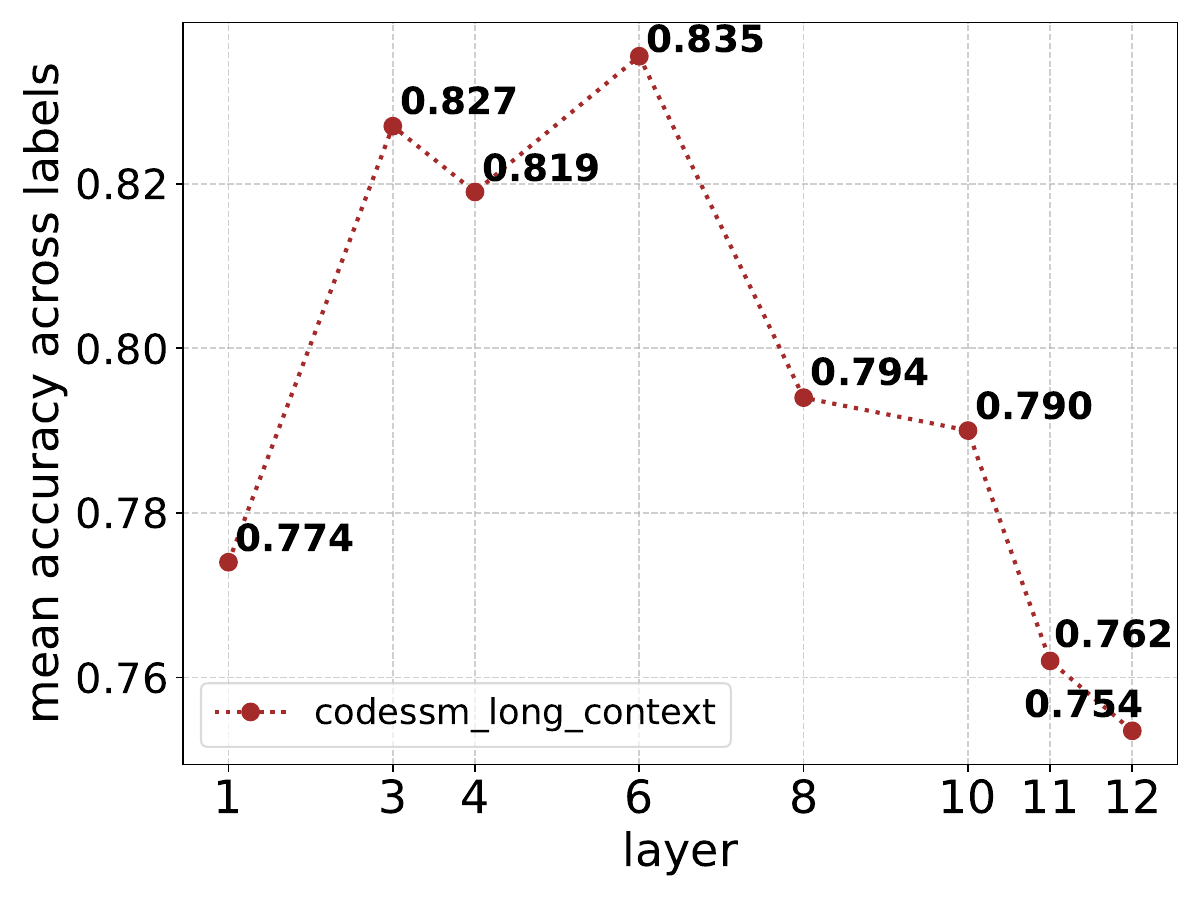}
     \includegraphics[width=0.33\linewidth]{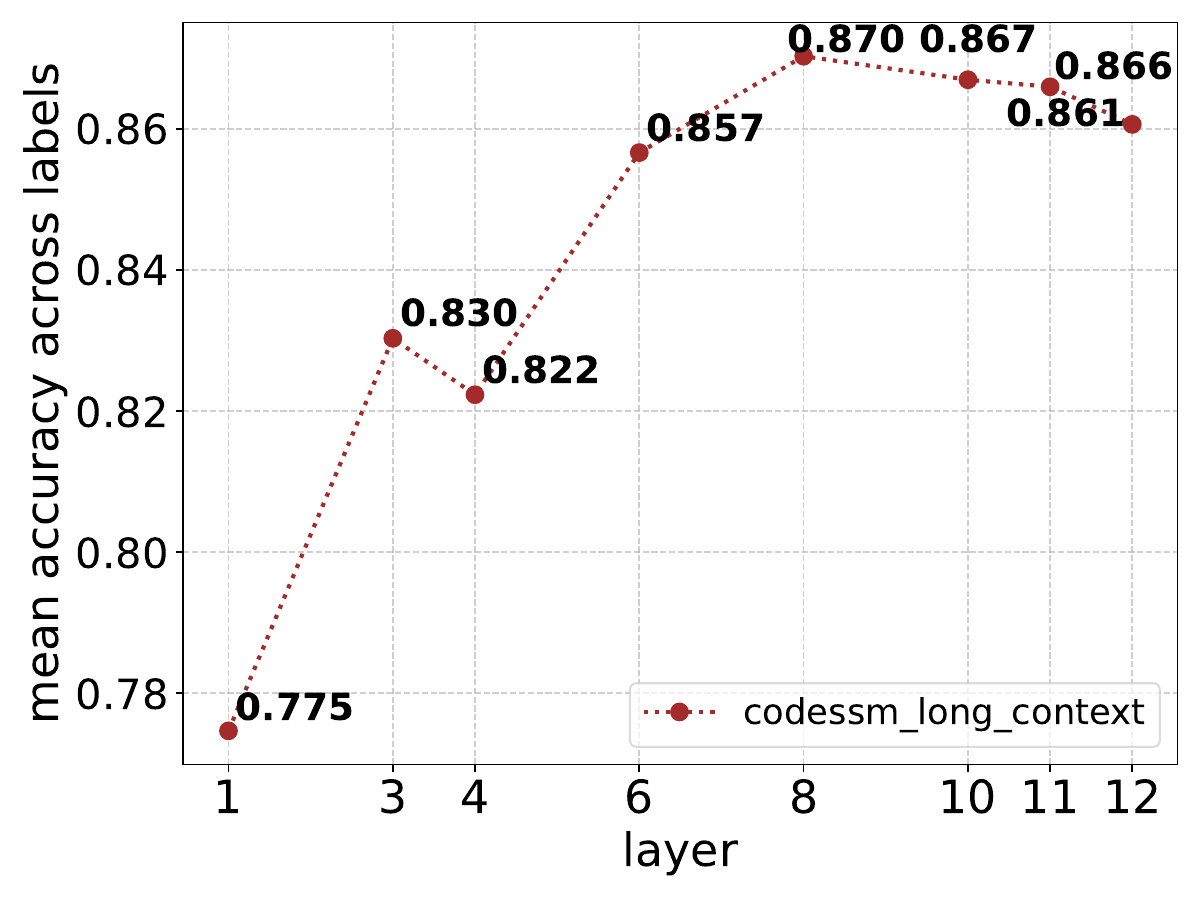}
    \end{subfigure}
    \caption{Comparison of hidden representation of CodeSSM and RoCoder on distance (left), siblings (center) and edge (right) prediction tasks for long lengths.}
    \label{fig: hiddenlong}
\end{figure*}

We performed direct probe analysis of CodeSSM and RoCoder on varying lengths: 1) Short (up to 500), 2) Medium (1k-2k), 3) Long (2k-8k). The results on short lengths are presented in \cref{fig: hiddenrepr_pretr}, which shows that CodeSSM captures more syntactic and semantic properties of code than RoCoder. The results for medium lengths for CodeSSM and RoCoder are shown in \cref{fig: hidden1k-2k}. The results for long length are only available for CodeSSM (shown in \cref{fig: hiddenlong}) because RoCoder consumes significantly more memory, and it was not possible to do the analysis with the resources we have. We observe that the gap between CodeSSM and RoCoder increases with increasing context length. CodeSSM captures better syntactic and semantic properties at longer lengths. 

\begin{figure*}[t]
\centering
    \begin{subfigure}[b]{0.99\textwidth}
     \includegraphics[width=0.33\linewidth]{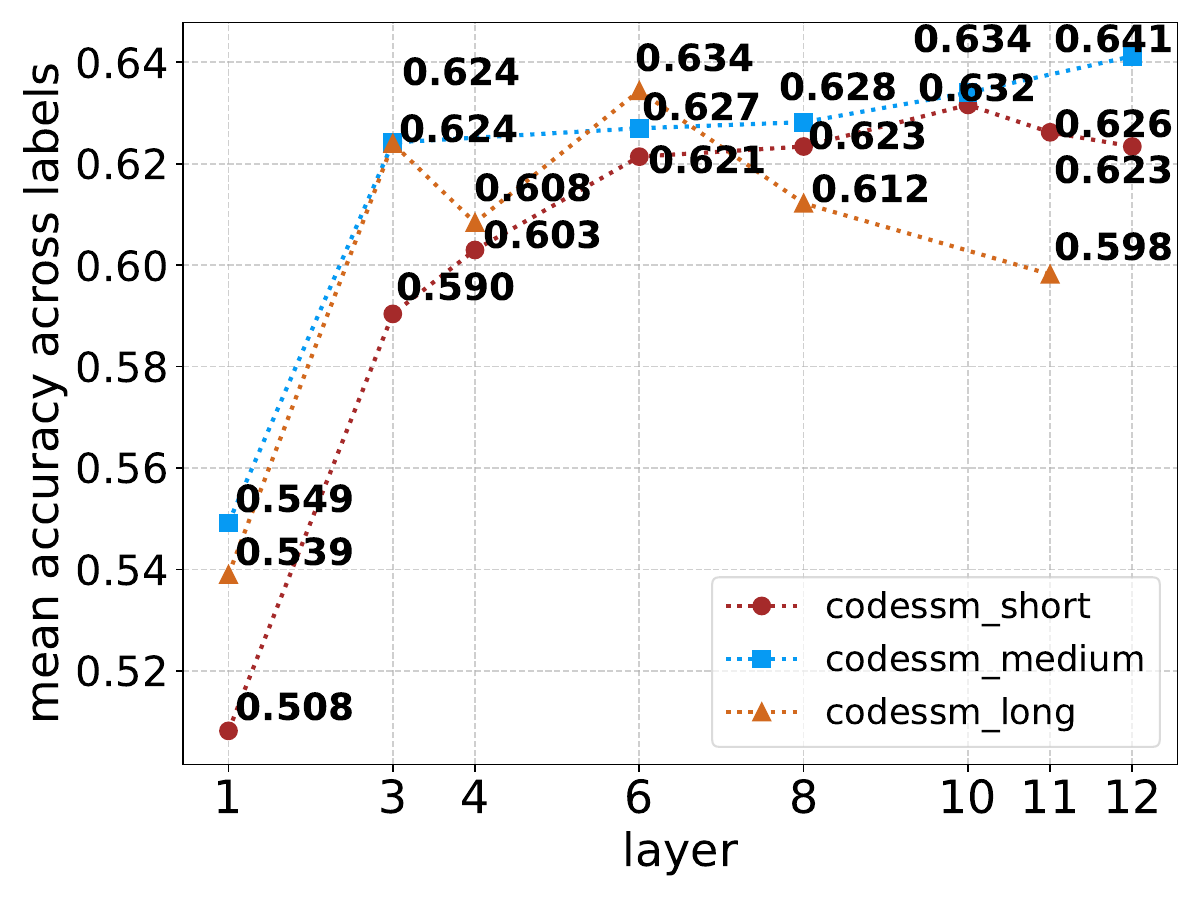} 
     \includegraphics[width=0.33\linewidth]{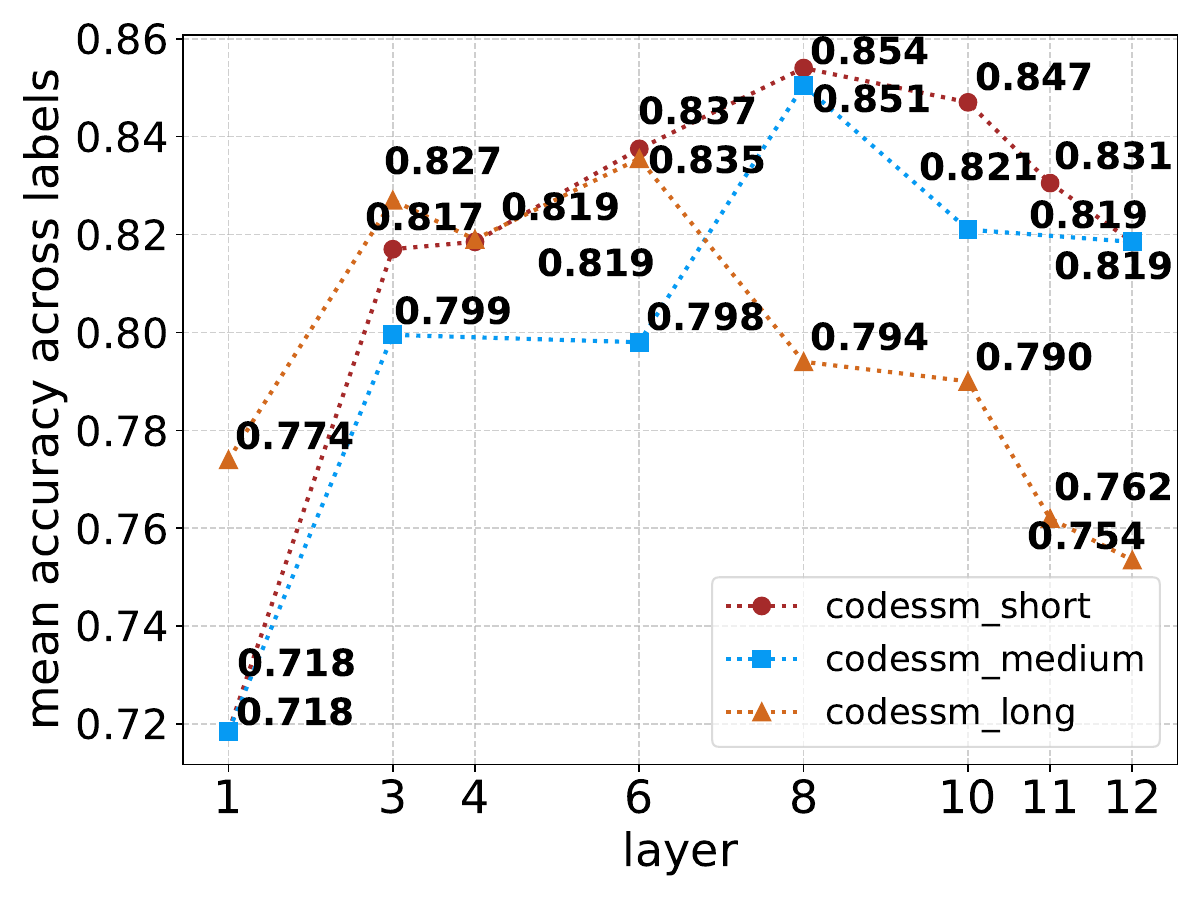}
     \includegraphics[width=0.33\linewidth]{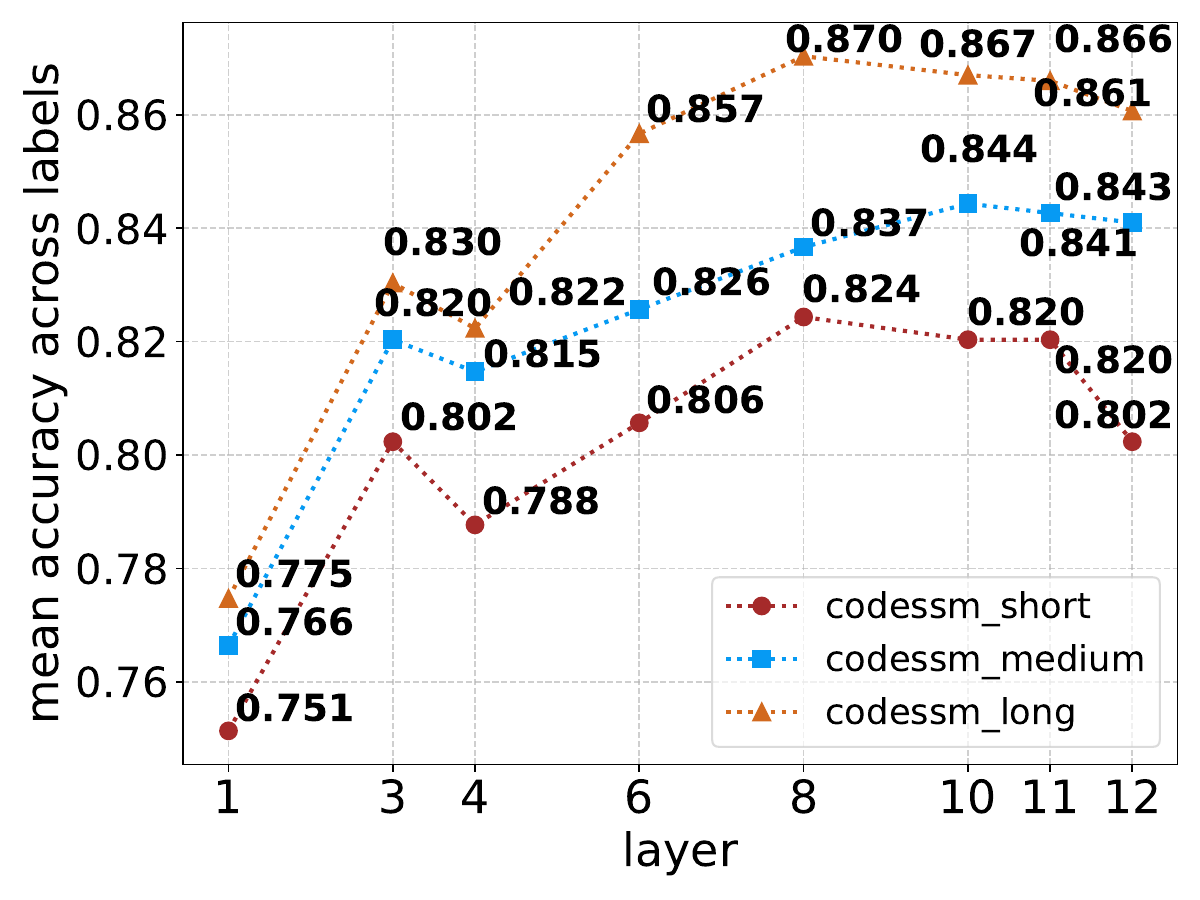}
    \end{subfigure}
    \caption{Comparison of hidden representation of CodeSSM on distance (left), siblings (center) and edge (right) prediction tasks with varying context lengths.}
    \label{fig: differentlen}
\end{figure*}

CodeSSM shows remarkable length extrapolation up to 8k context length, which is 32 times the pretraining length, while maintaining both syntactic and semantic performance. In \cref{fig: differentlen}, we observe that CodeSSM shows the best accuracy in capturing semantic properties (i.e., on the DFG task) for long lengths. The superior performance on the DFG task implies that CodeSSM captures long-range dependencies very well (especially for long lengths). While we observe that CodeSSM shows significantly better understanding of long-range dependencies at long lengths, it shows representational degradation in later layers for local dependencies (i.e., siblings tasks) at long lengths.

The exact mechanism that enables CodeSSM to extrapolate to such a large context is unclear. We hypothesize that it is due to the Discrete Fourier Transform (DFT) in the convolution operation. 

Numerous variants of Positional Embeddings have been proposed for the Transformer architecture. RoPE \cite{Rope} is the one most commonly used. However, RoPE provides about 2x extrapolation over pretraining context length. \citet{fope} showed that RoPE implicitly performs non-uniform DFT. However, RoPE suffers from spectral leakage due to a single frequency for each dimension. \citet{fope} proposed FoPE which uses a mixture of frequency across each dimension to avoid spectral leakage. FoPE provides upto 4x extrapolation over pretraining context size.

S4D performs explicit DFT in the convolution operation. Additionally, each dimension comprises of a mixture of frequency, just like FoPE. However, unlike FoPE, the basis of the frequency mixture is defined by the $A$ matrix, which has a special initialization -- HiPPO \cite{h3}. The HiPPO matrix is designed to improve the long-range memory and thus this special initialization might be the reason for the extreme length extrapolation. Positional Embedding derived from the HiPPO matrix might also help improve length extrapolation of Transformers. However, additional experiments is required to validate this claim robustly. Besides, the HiPPO initialization, the learnable parameters might also help SSM adapt the token relations to the data.

However, SSM alone is not sufficient for length extrapolation \cite{ravikumar2026analysislongrangedependency} and has not achieved success on practical real-world applications. In CodeSSM, the SSM routing comprises of two feed forward networks, one preceding and one following the SSM block. This architectural setup is similar to the previously proposed MLP-Mixer \cite{mlpmixer} and Conv-Mixer \cite{convmixer} architectures. Similar to these architectures, the CodeSSM model separates the channel-wise mixing (using feed forward layer) and the length-wise mixing (using S4D). CodeSSM differs in that the length-wise mixing is performed globally over the entire sequence length.

\section{Long Range Probing Tasks}
We would like to explain why our probing tasks capture long-range dependencies in the sense most relevant to code. Long-range dependency discussions in the literature are typically framed in terms of token distance. However, probing tasks for code are performed over structural representations — AST and DFG — because these capture the semantically meaningful relationships in programs, which do not necessarily align with token distance. A distance of 5–6 in the AST, or an edge in the DFG between two variable occurrences, can span a substantial number of tokens depending on nesting depth, loop structure, and branch complexity. Our distance prediction task explicitly evaluates performance at AST distances 2–6 (Figure 4), with distances 5–6 corresponding to structurally distant token pairs. Similarly, DFG edge prediction captures data-flow dependencies between variables that may be arbitrarily far apart in token space. We also present some examples from our dataset which show that our tasks require long-range dependency understanding. We present some examples for DFG in \cref{fig: dfgeg1} and \cref{fig: dfgeg2}. There are 1099 and 1486 tokens between two $positions$ variables and the two $start\_time$ variables respectively.

%%%%%%%%%%%%%%%%%%%%%%%%%%%%%%%%%%%%%%%%%%%%%%%%%%%%%%%%%%%%%%%%%%%%%%%%%%%%%%%
%%%%%%%%%%%%%%%%%%%%%%%%%%%%%%%%%%%%%%%%%%%%%%%%%%%%%%%%%%%%%%%%%%%%%%%%%%%%%%%

\section{Additional Evaluation Results}
\begin{figure*}[t]
\centering
    \begin{subfigure}[b]{0.99\textwidth}
     \includegraphics[width=0.33\linewidth]{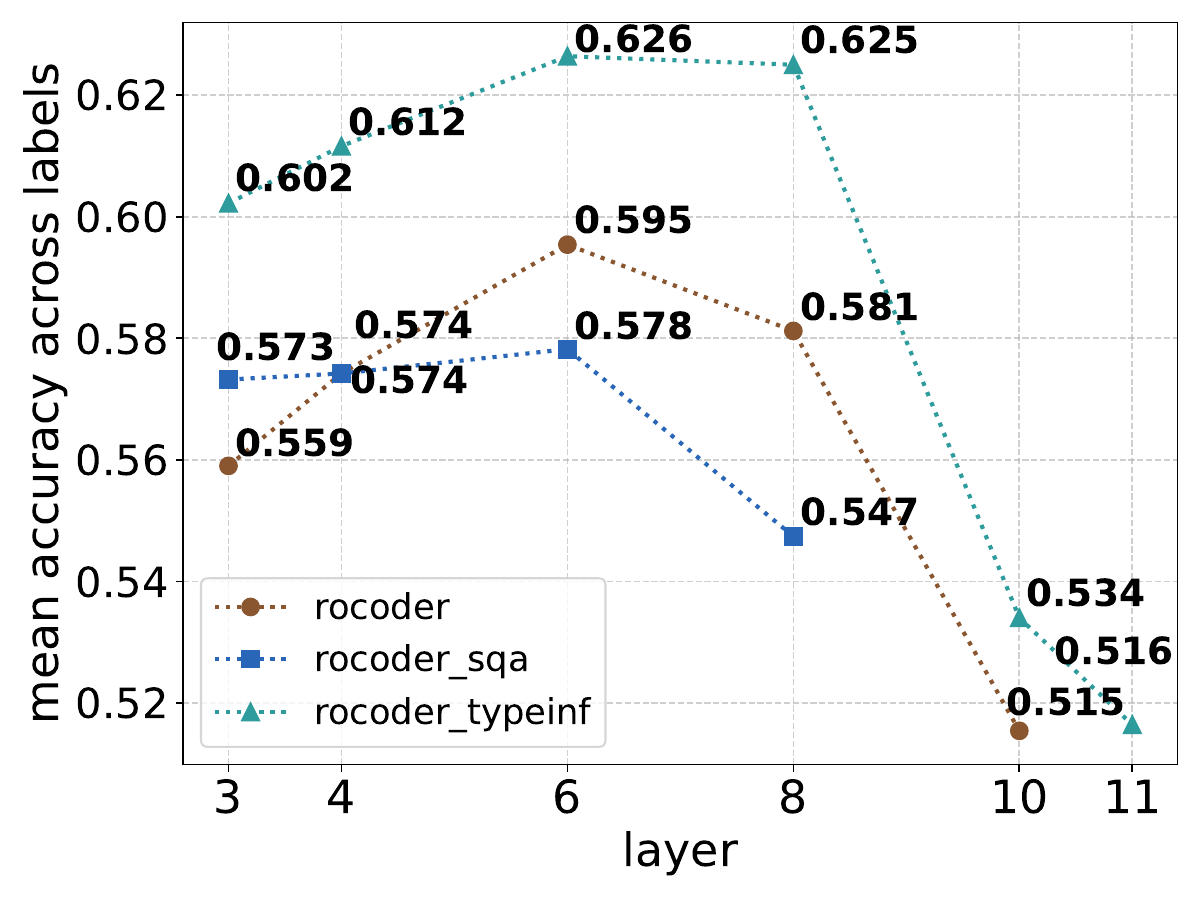} 
     \includegraphics[width=0.33\linewidth]{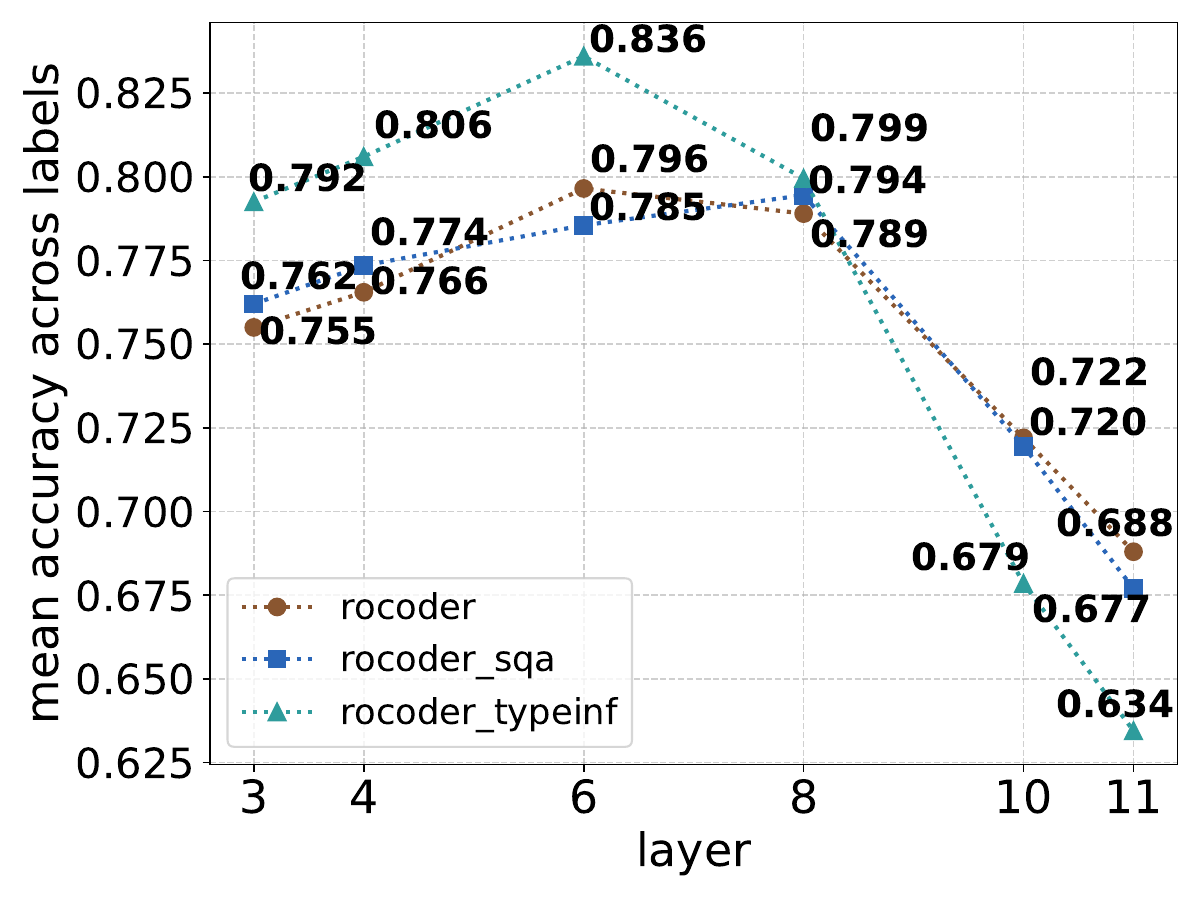}
     \includegraphics[width=0.33\linewidth]{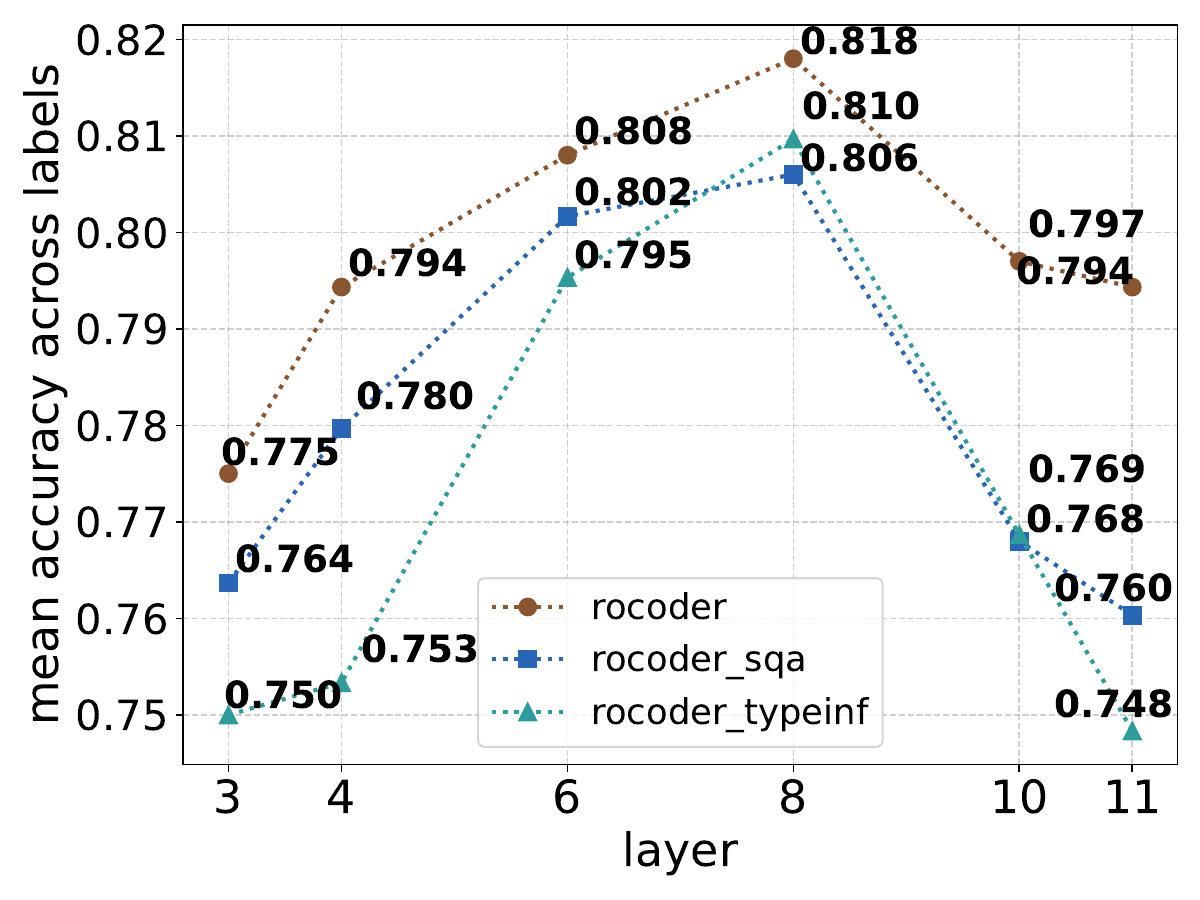} 
     \caption{Analysis results of RoCoder in comparsion with its finetuned variants.}
     \label{fig: hidden repr finetunedRocoder}
    \end{subfigure}
\end{figure*}

\begin{figure*}[htbp]
\centering
    \begin{subfigure}[b]{0.99\textwidth}
     \includegraphics[width=0.33\linewidth]{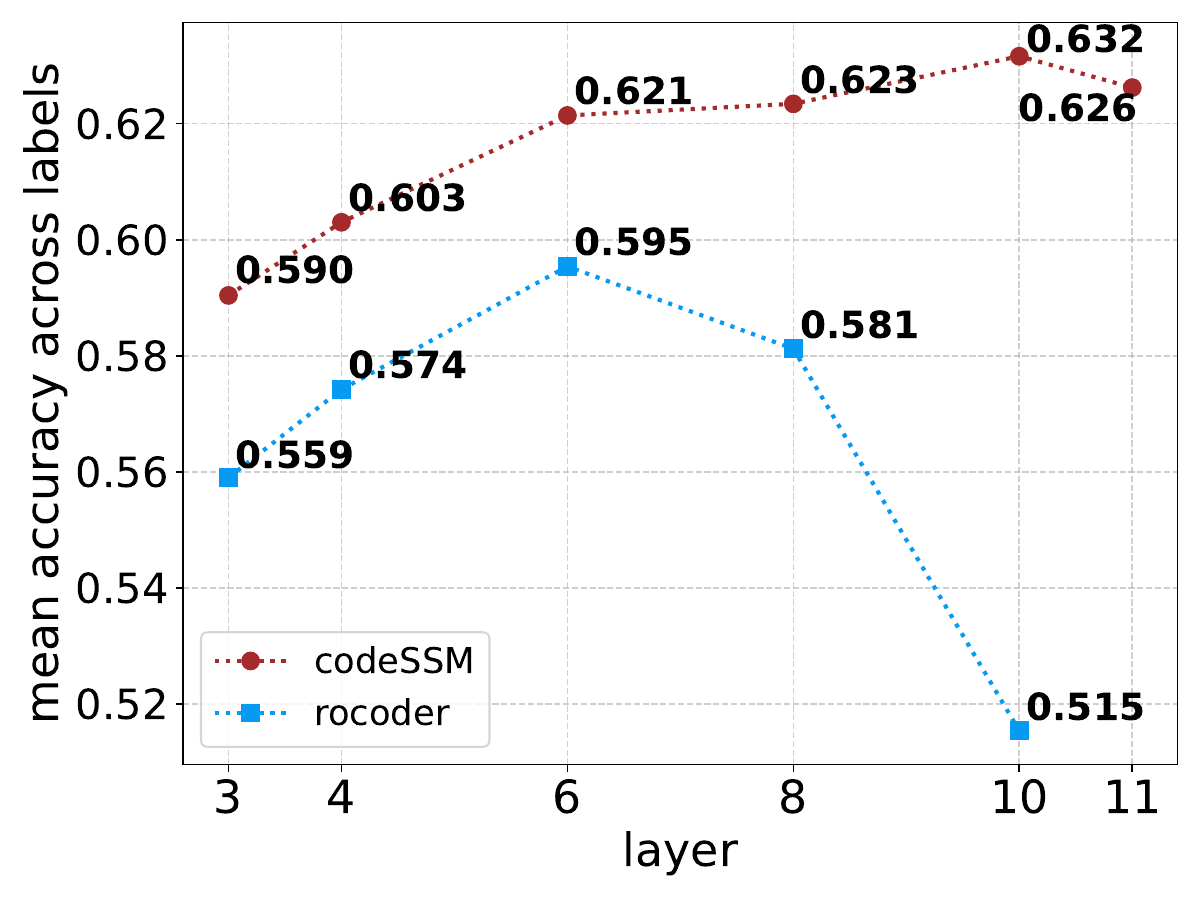} 
     \includegraphics[width=0.33\linewidth]{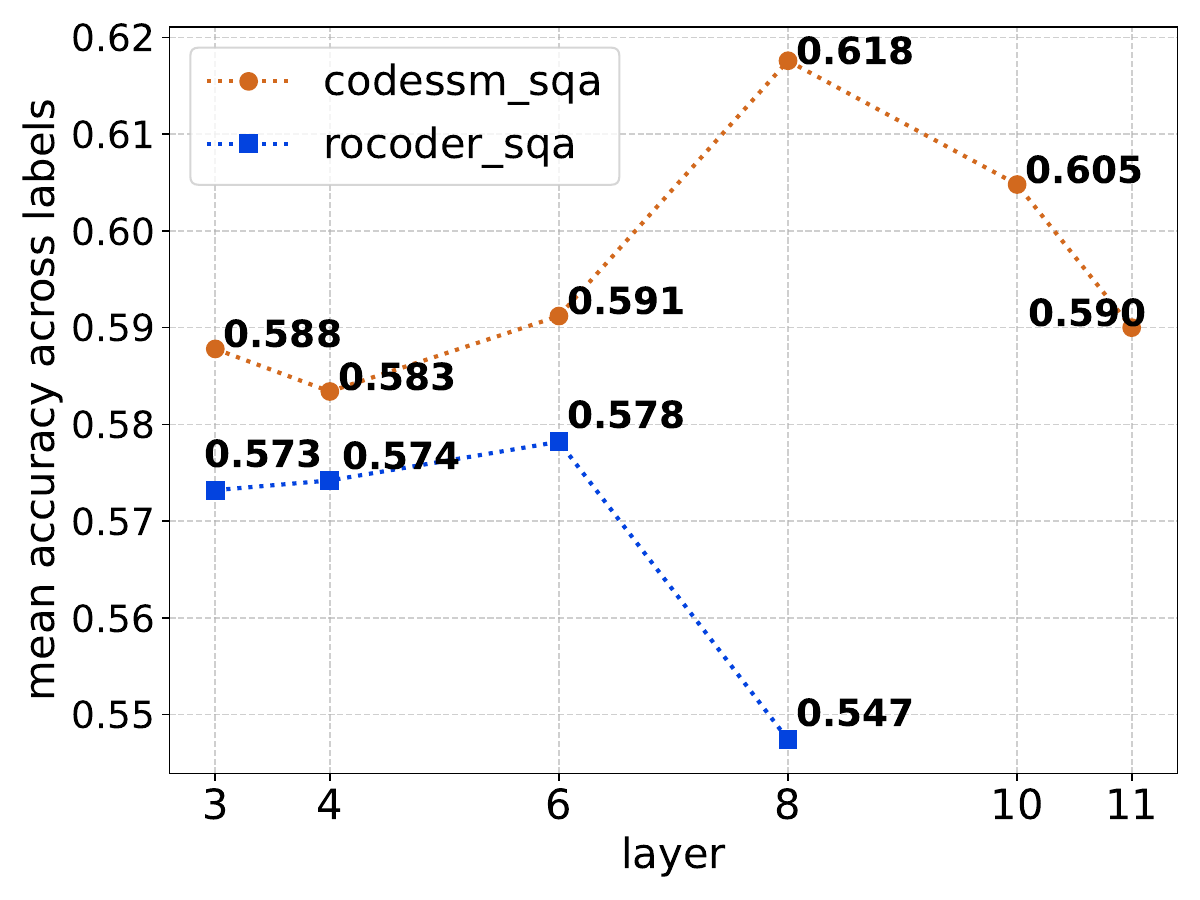}
     \includegraphics[width=0.33\linewidth]{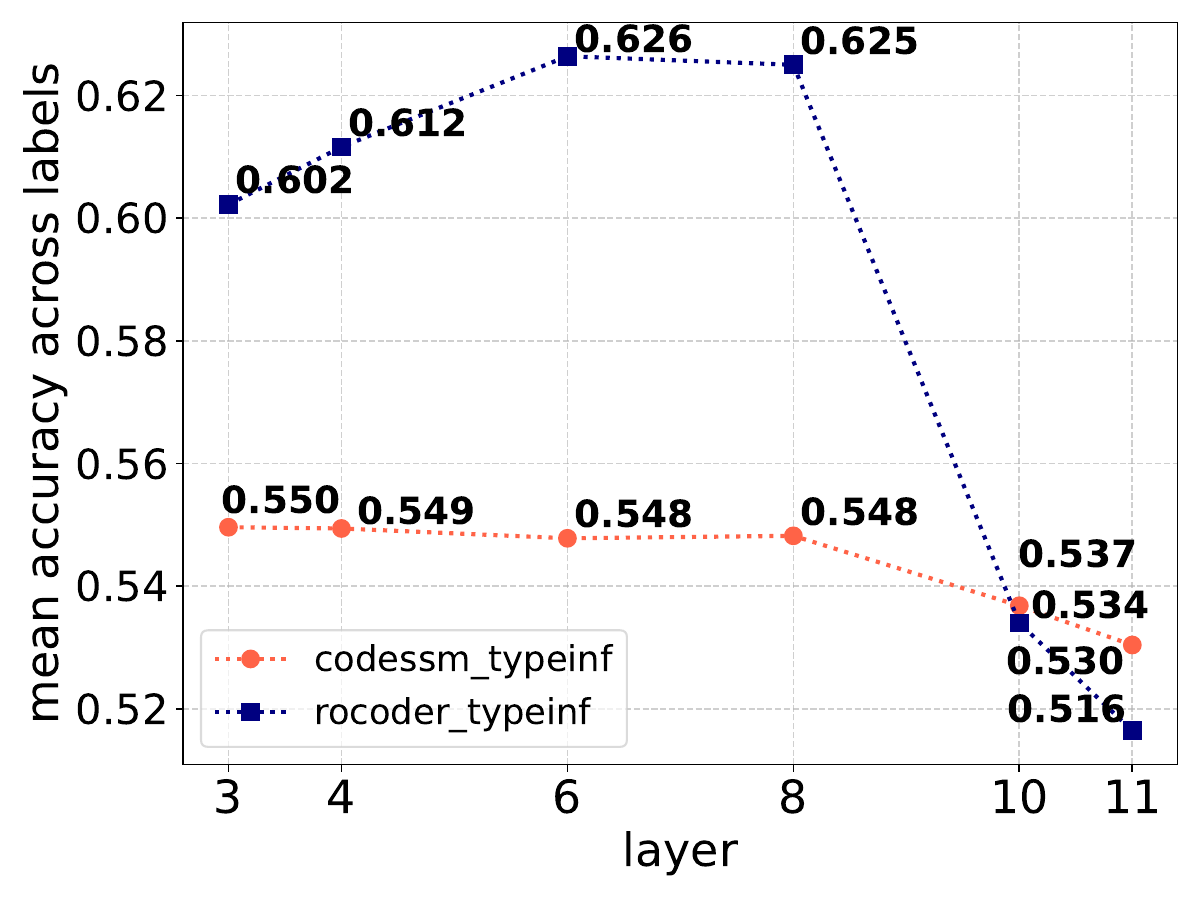}
     \caption{DirectProbe distance prediction task}
    \end{subfigure}
    \begin{subfigure}[b]{0.99\textwidth}
     \includegraphics[width=0.33\linewidth]{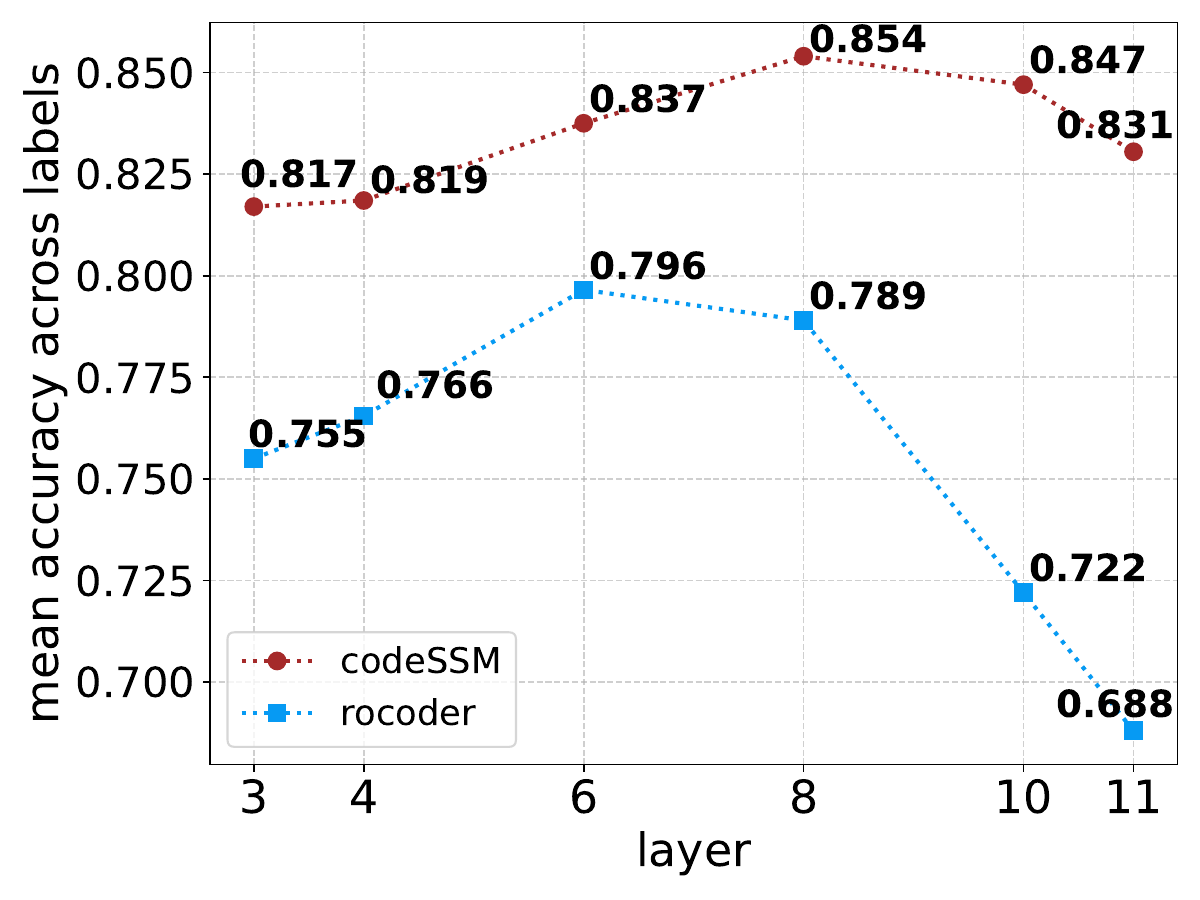} 
     \includegraphics[width=0.33\linewidth]{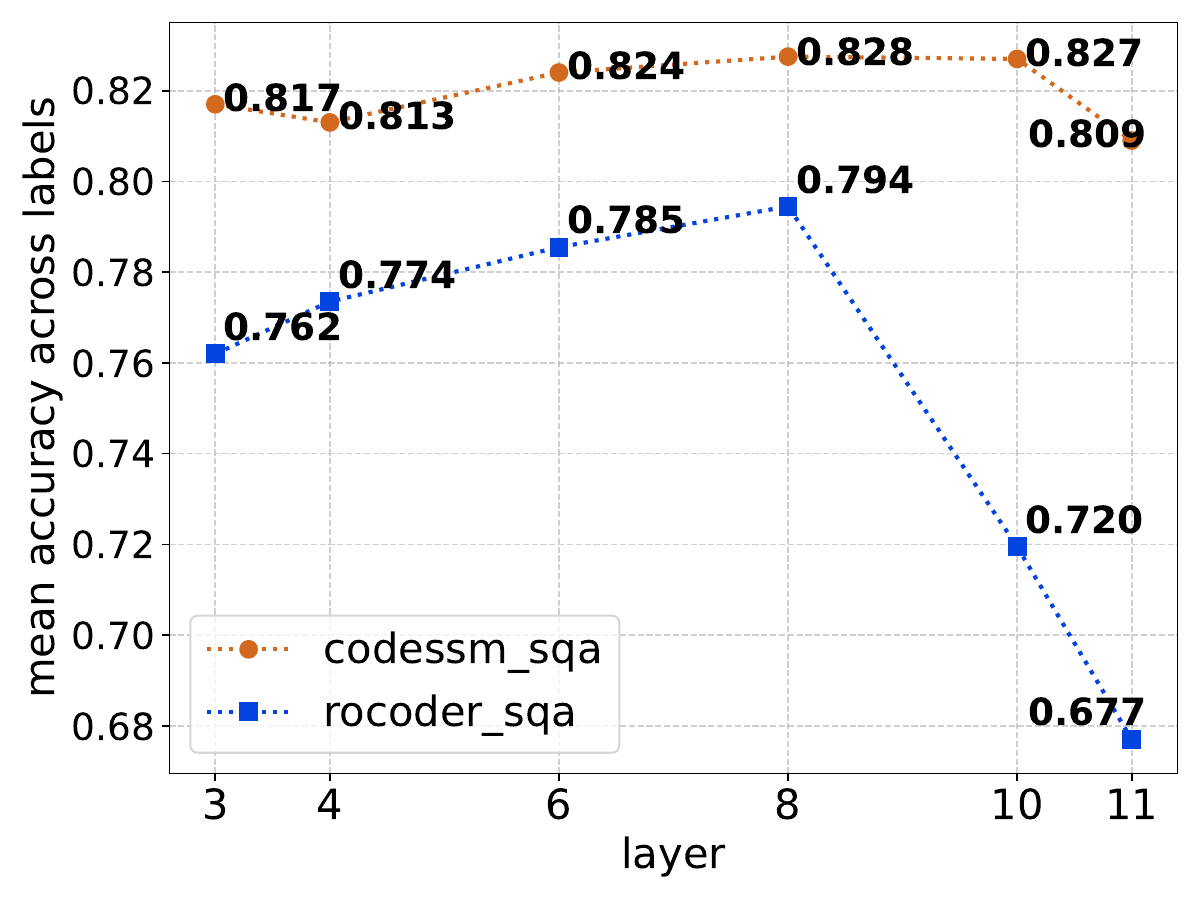}
     \includegraphics[width=0.33\linewidth]{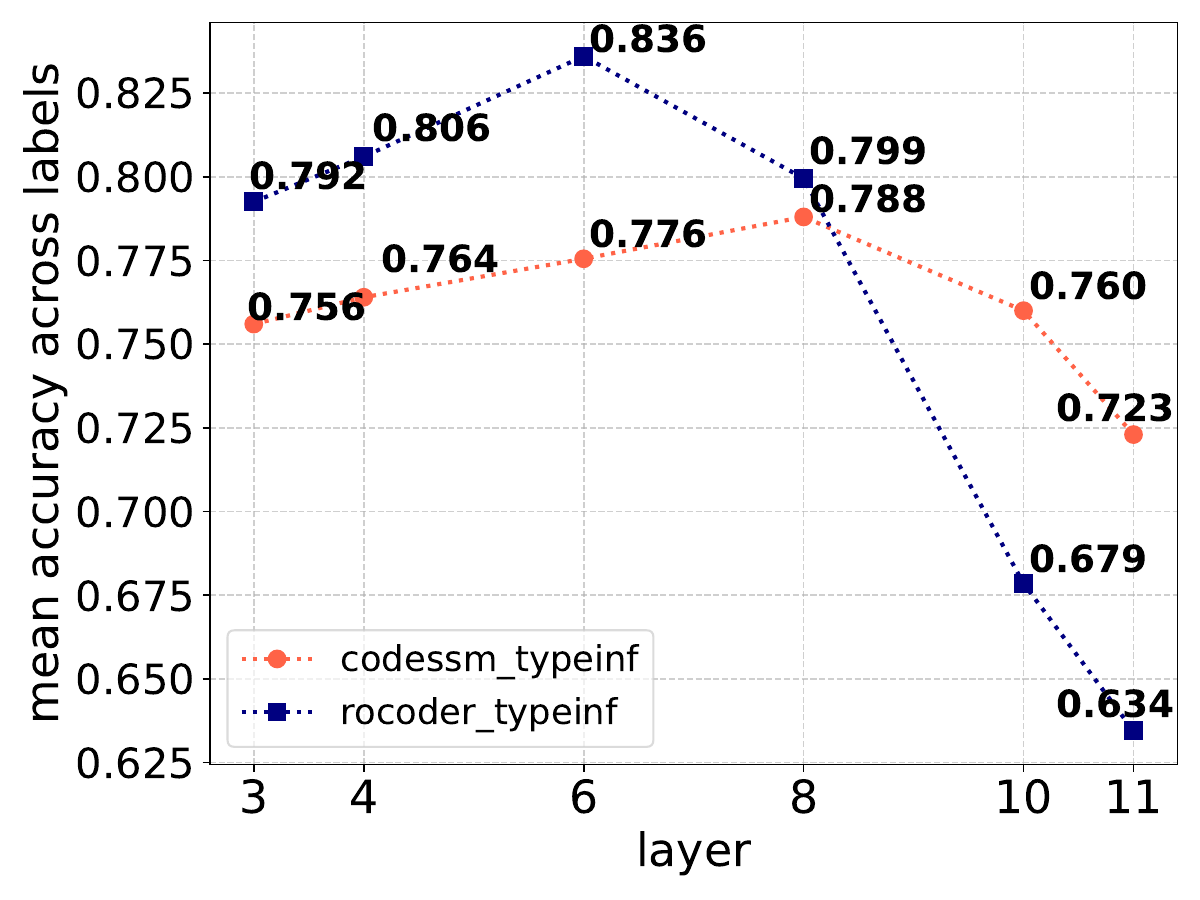}
     \caption{DirectProbe siblings prediction task}
    \end{subfigure}
    \begin{subfigure}[b]{0.99\textwidth}
     \includegraphics[width=0.33\linewidth]{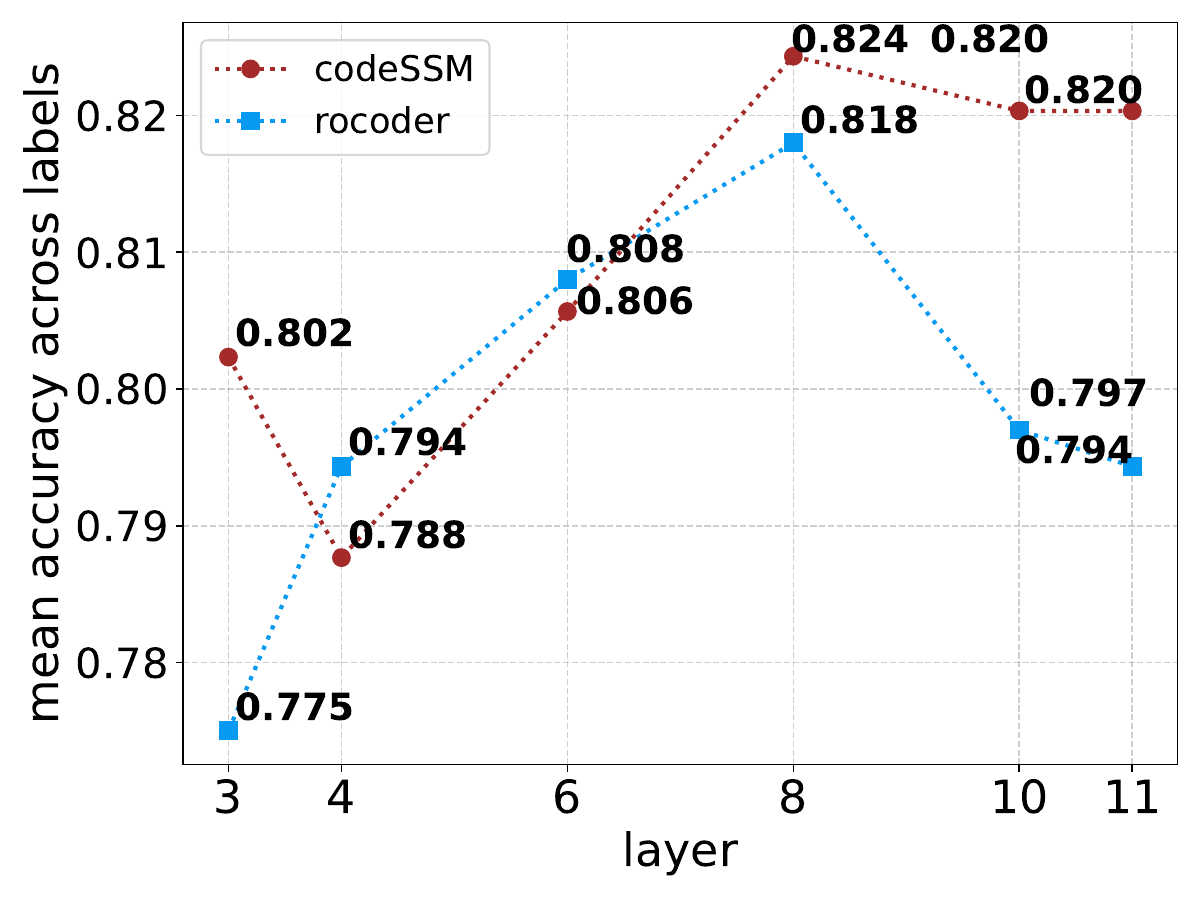} 
     \includegraphics[width=0.33\linewidth]{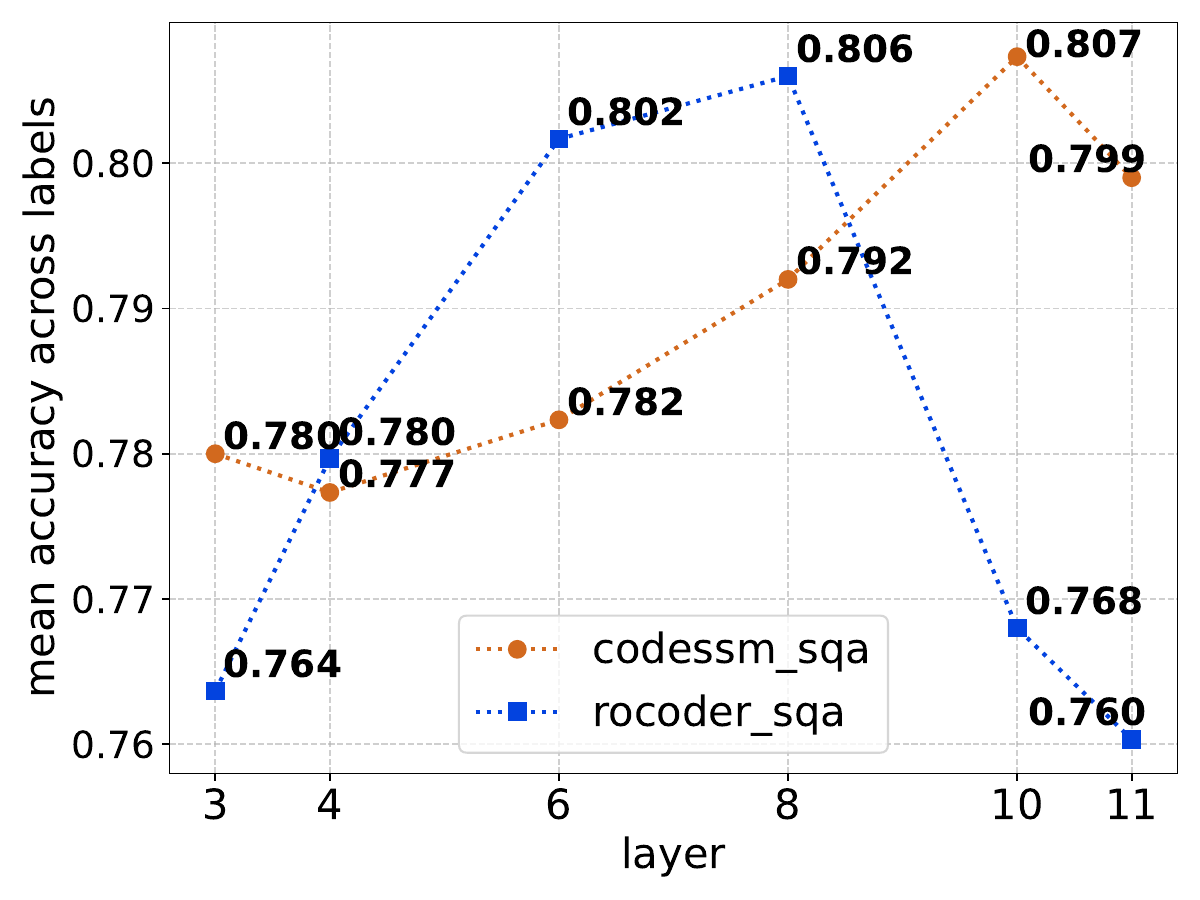}
     \includegraphics[width=0.33\linewidth]{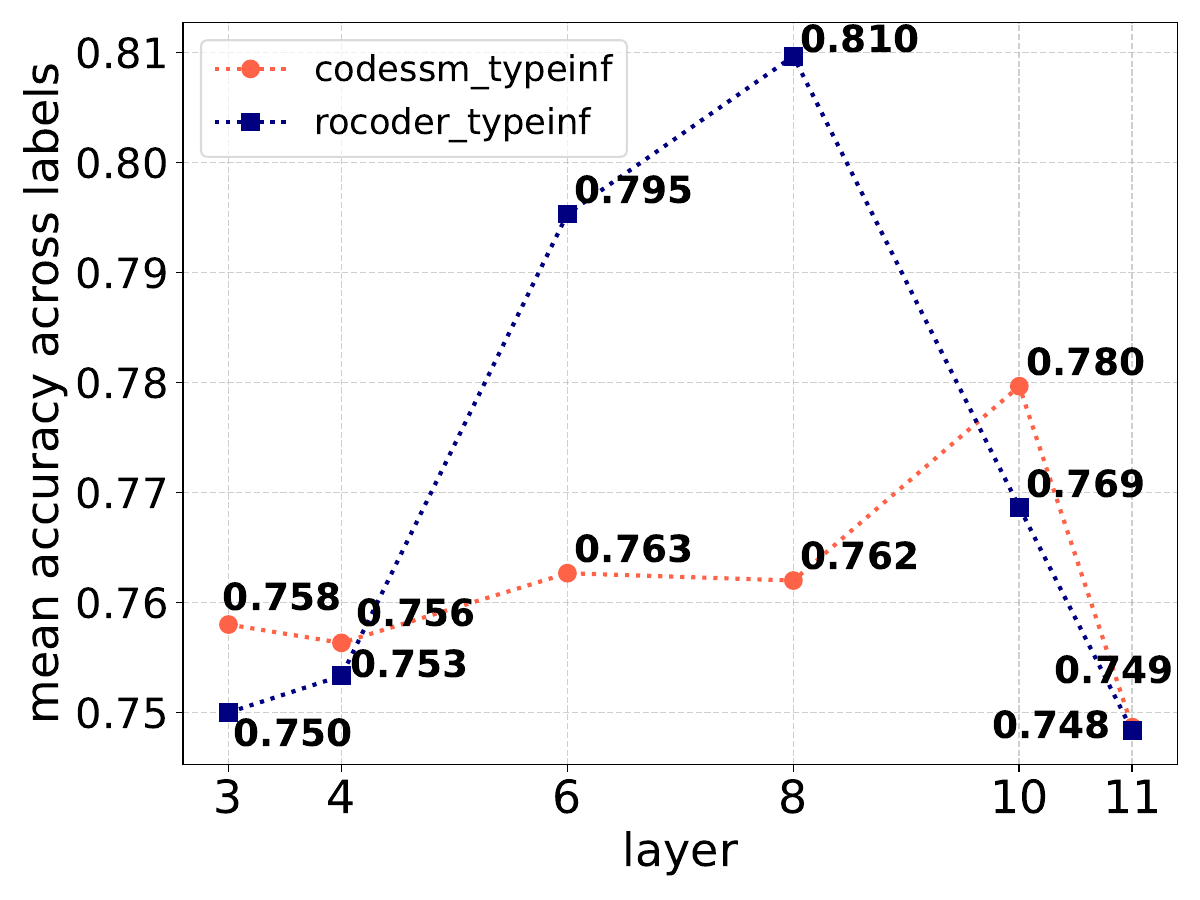} 
    \caption{DirectProbe DFG edge prediction task}
    \end{subfigure}
    
    \caption{Comparative analysis of hidden representation of CodeSSM and RoCoder (left), after finetuning on SQA (center) and after finetuning on type inference (right) for some layers.}
    \label{fig: hidden repr model comp}
\end{figure*}

\begin{figure*}[h]
\centering
    \begin{subfigure}[b]{0.99\textwidth}
     \includegraphics[width=0.5\linewidth]{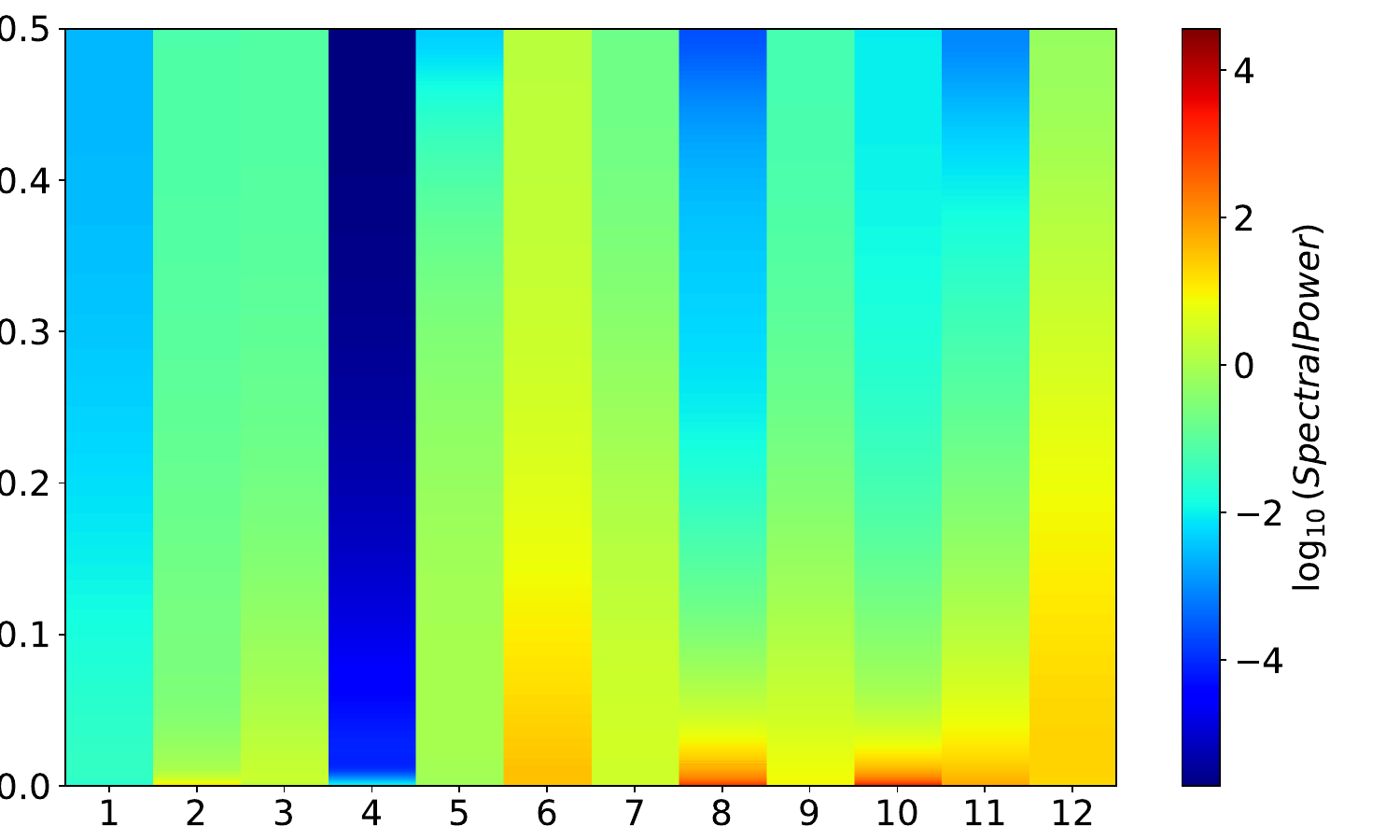} 
     \includegraphics[width=0.5\linewidth]{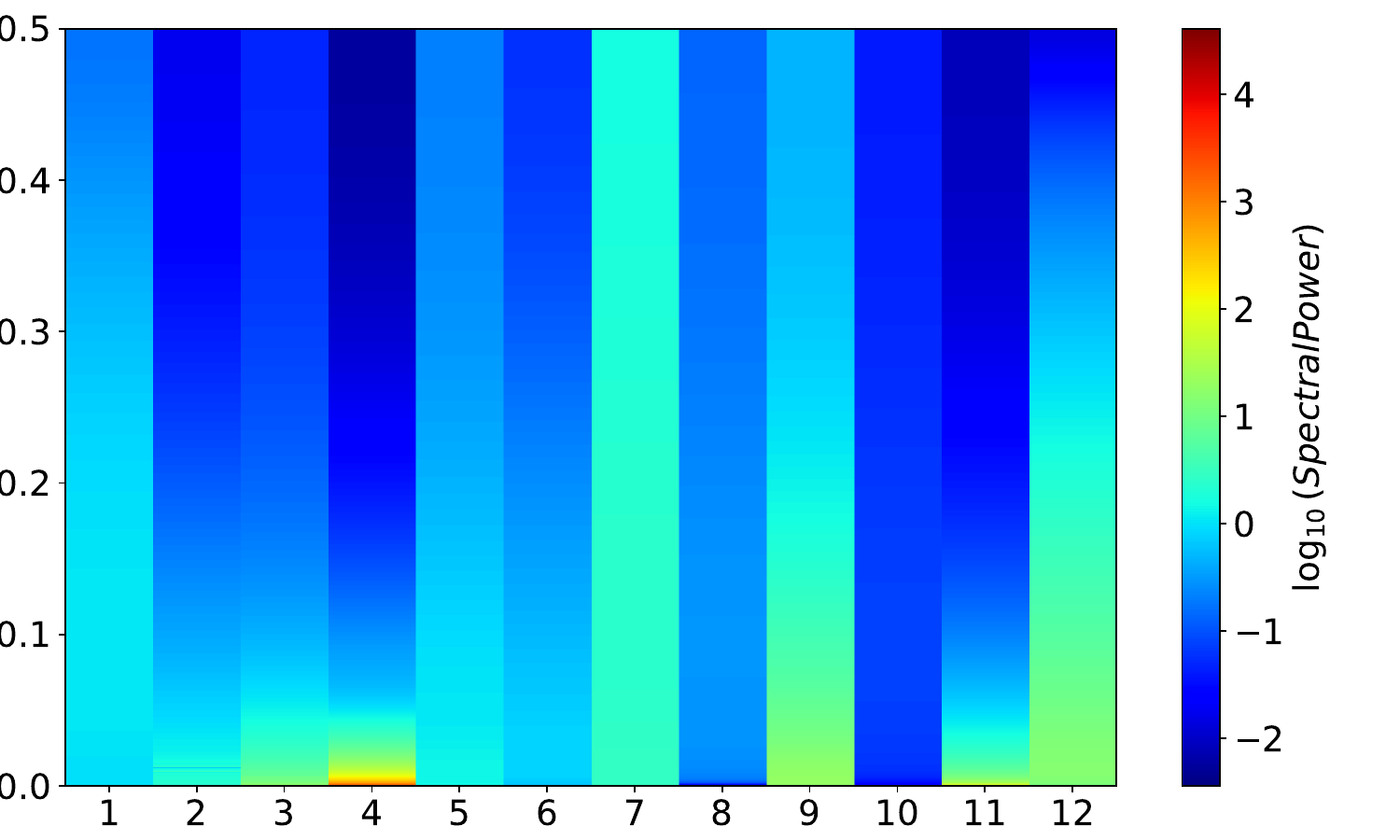}
     \caption{CodeSSM}
    \end{subfigure}
    \begin{subfigure}[b]{0.99\textwidth}
     \includegraphics[width=0.5\linewidth]{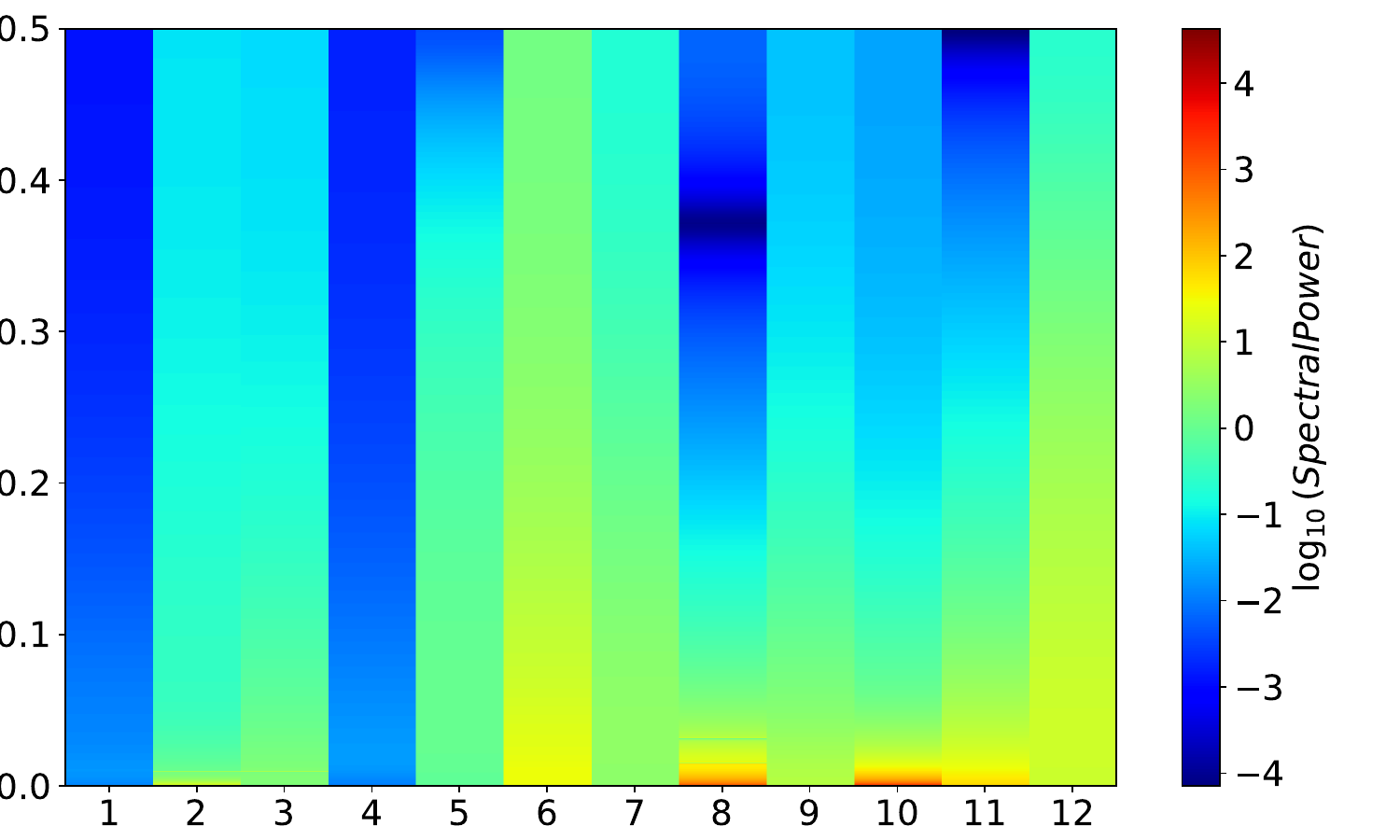} 
     \includegraphics[width=0.5\linewidth]{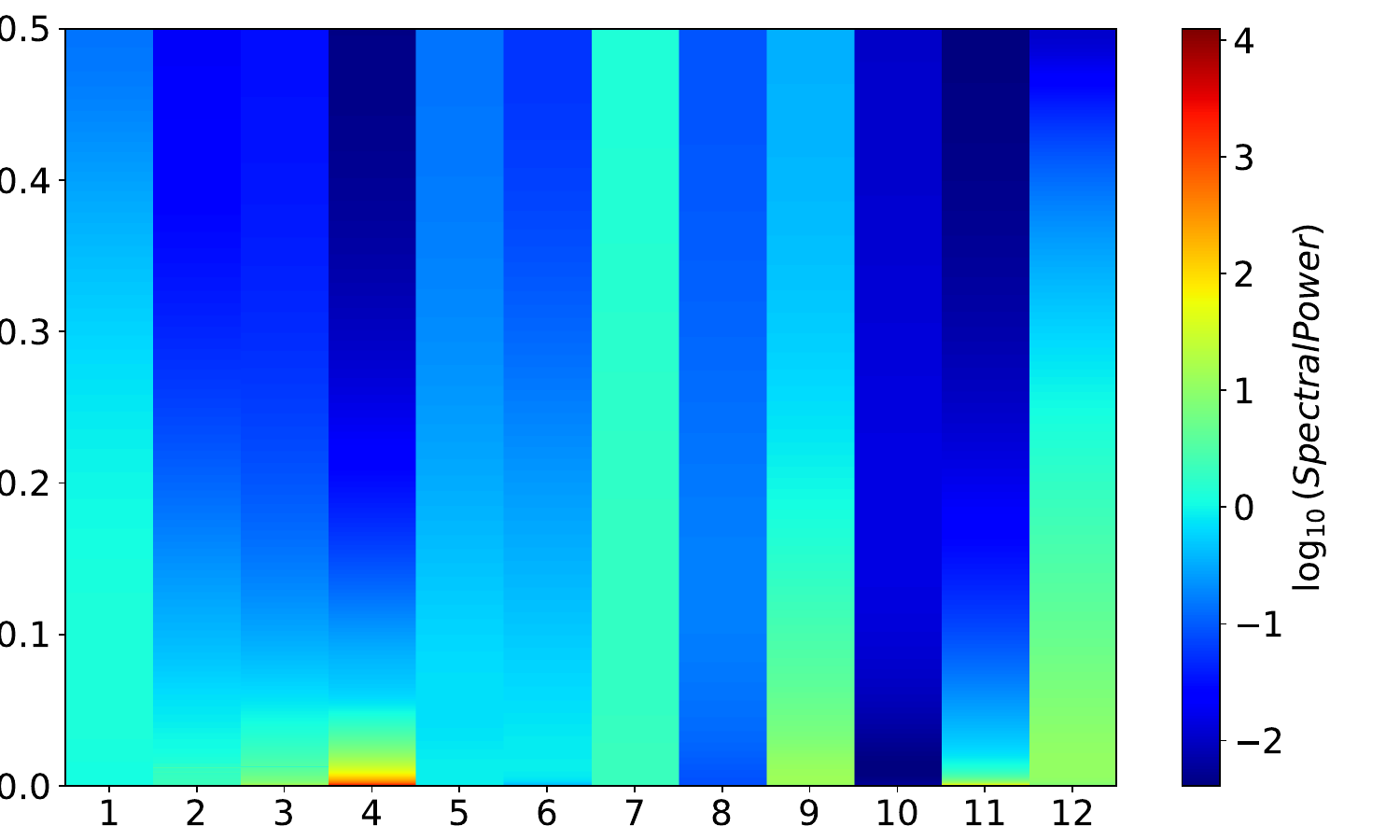}
     \caption{CodeSSM\_sqa}
    \end{subfigure}
    \begin{subfigure}[b]{0.99\textwidth}
     \includegraphics[width=0.5\linewidth]{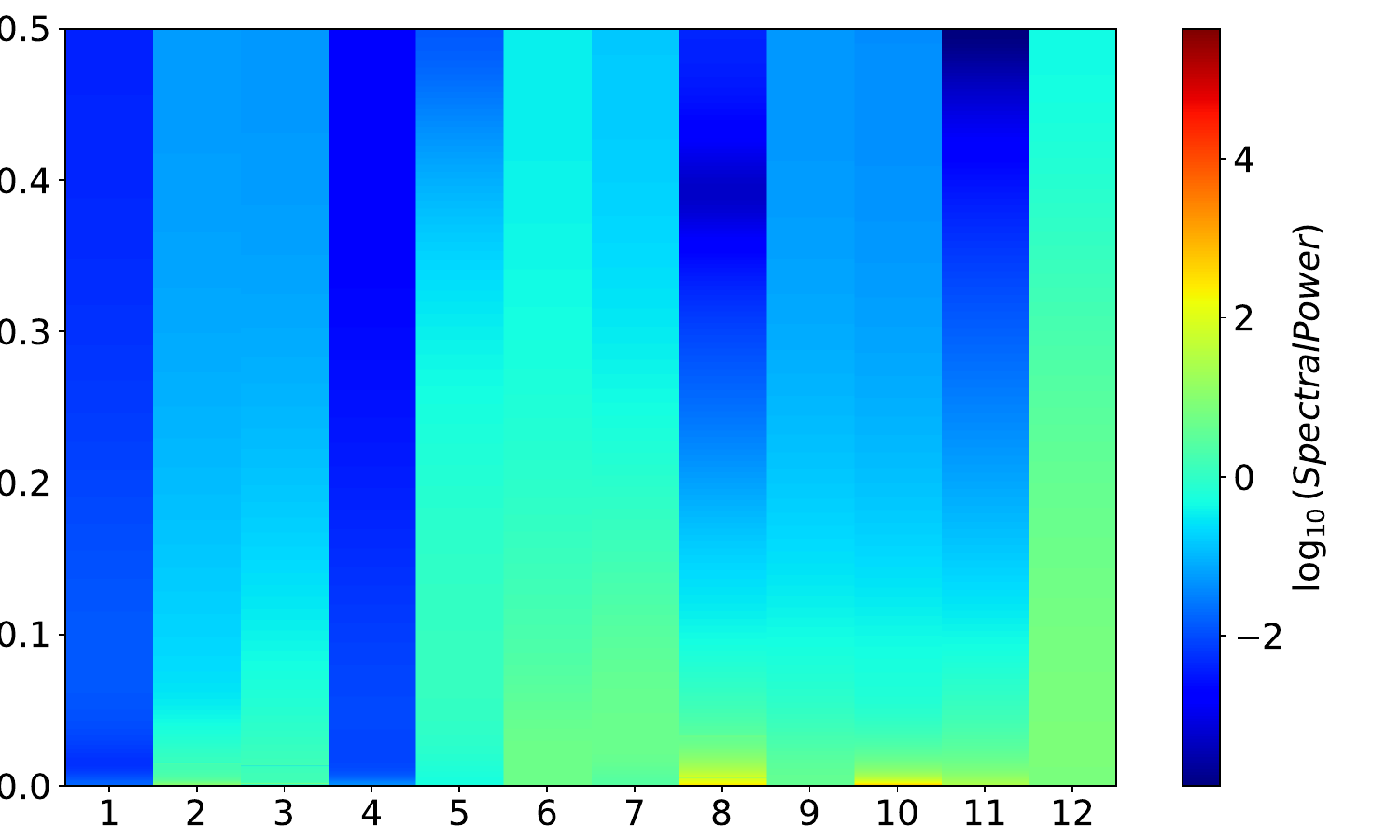} 
     \includegraphics[width=0.5\linewidth]{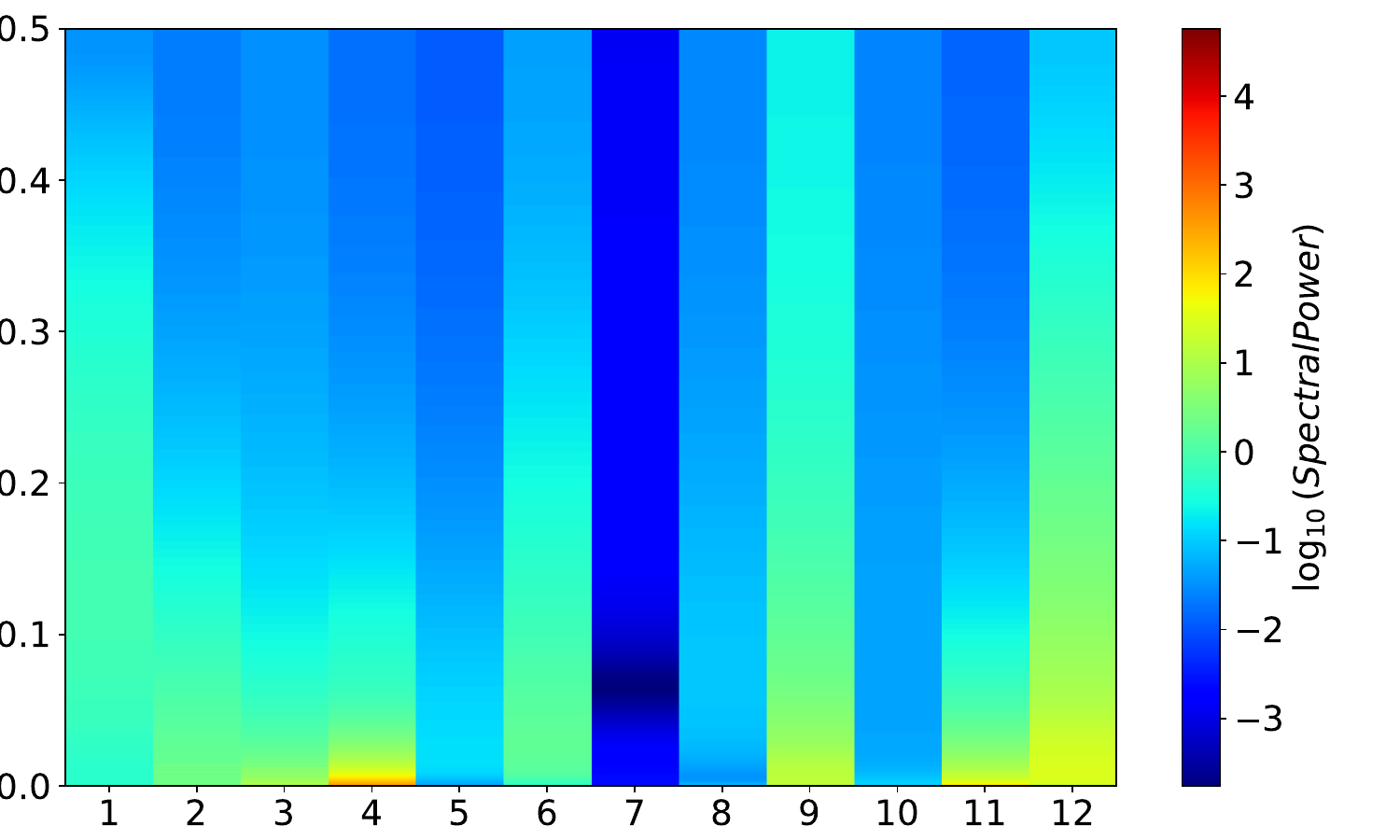}
    \caption{CodeSSM\_typeinf}
    \end{subfigure}
    \begin{subfigure}[b]{0.99\textwidth}
     \includegraphics[width=0.5\linewidth]{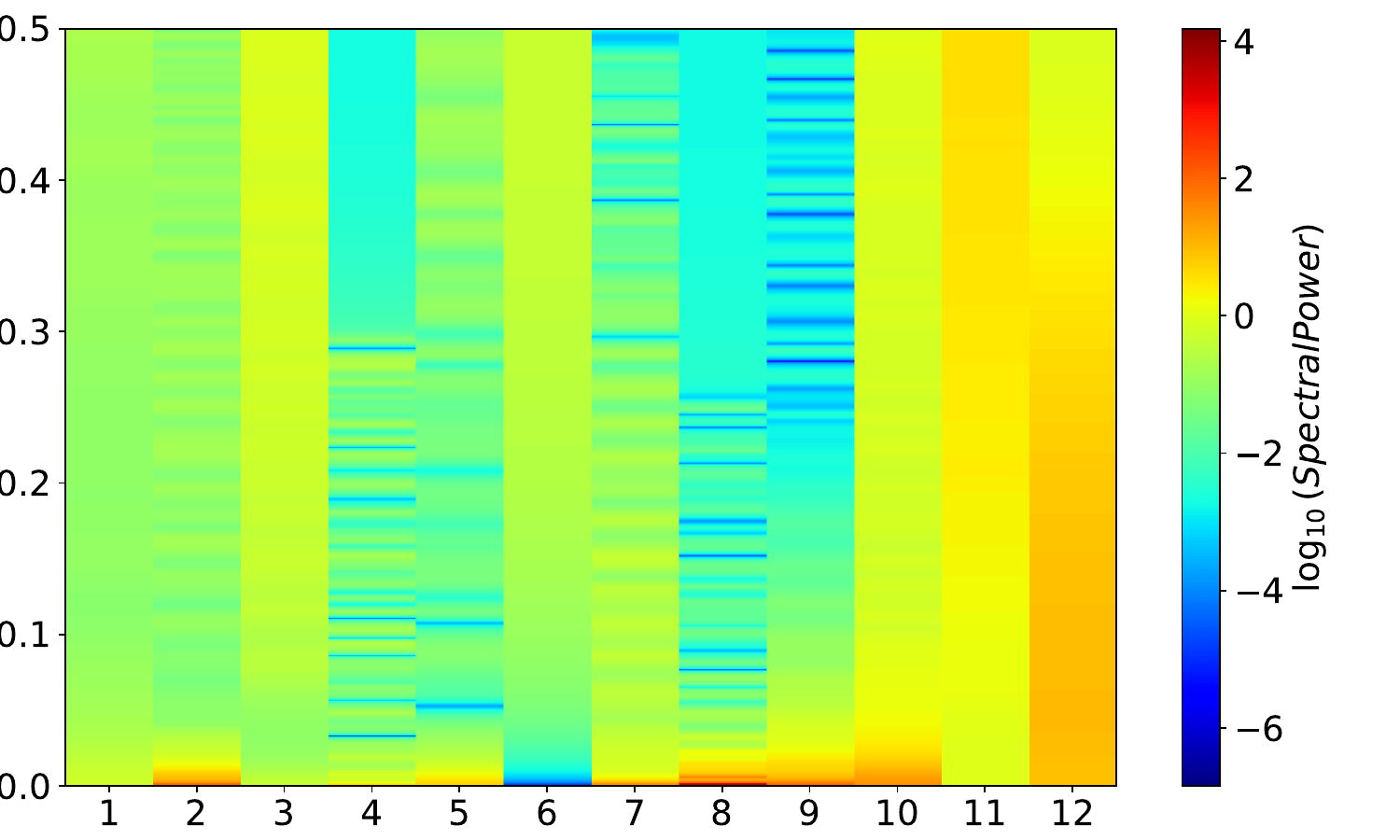} 
     \includegraphics[width=0.5\linewidth]{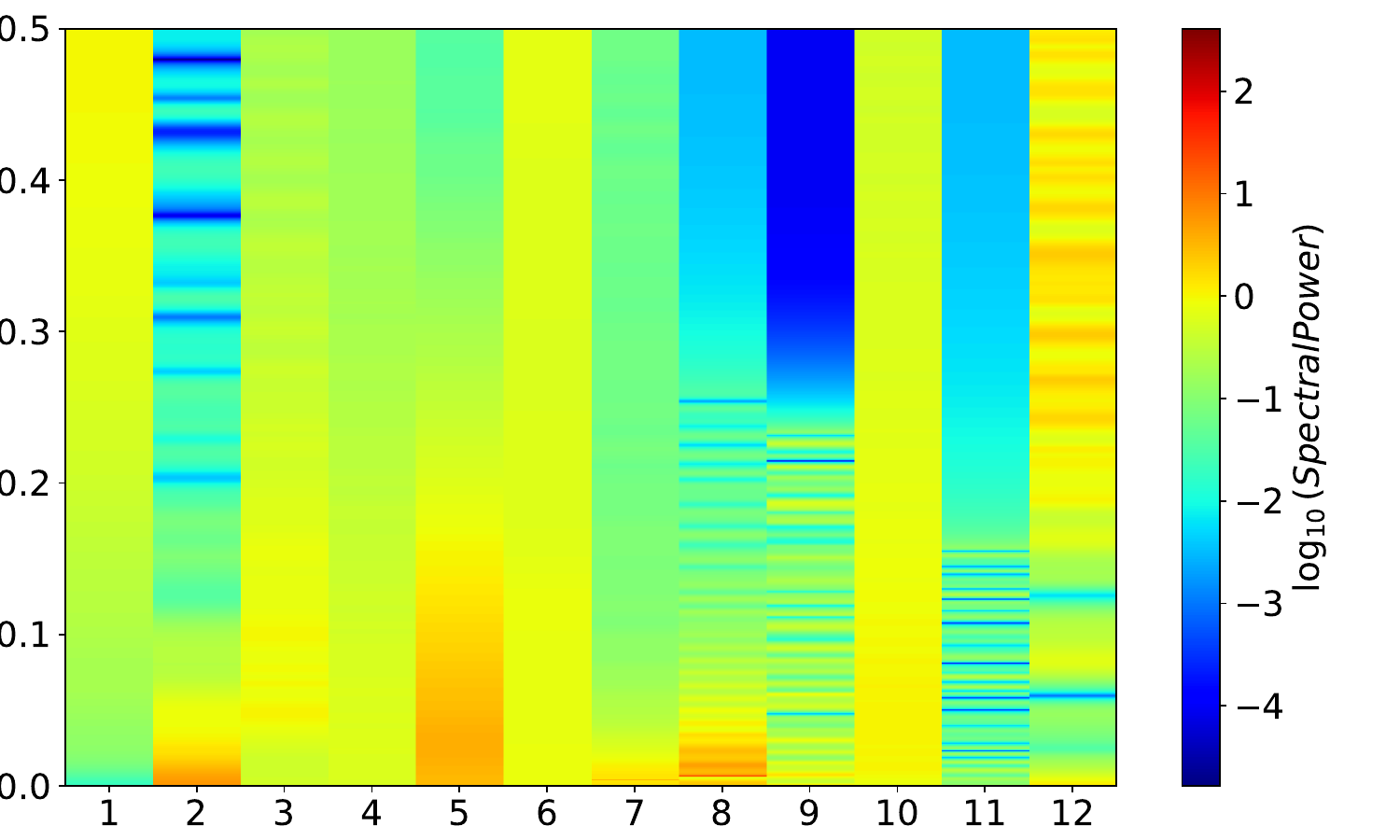}
    \caption{CodeSSM-8k}
    \end{subfigure}
    
    \caption{Heatmap visualization of spectral powers of CodeSSM, CodeSSM\_sqa, CodeSSM\_typeinf and CodeSSM-8k. For the CodeSS-8k model, we take the mean over all kernels. The visualization shows the richer token relations learnt in the case of CodeSSM-8k.}
    \label{fig: heatmap}
\end{figure*}

In this section, we present additional evaluation results. Figure \ref{fig: hidden repr model comp} shows the comparison between the two models and their fine-tuned version. We can observe the forgetting of code properties after finetuning on type inference. We also show the comparison of RoCoder with its finetuned variants in  \cref{fig: hidden repr finetunedRocoder}.

\begin{figure*}[h]
    \centering
    \begin{subfigure}[b]{\linewidth}
    \includegraphics[width=0.33\linewidth]{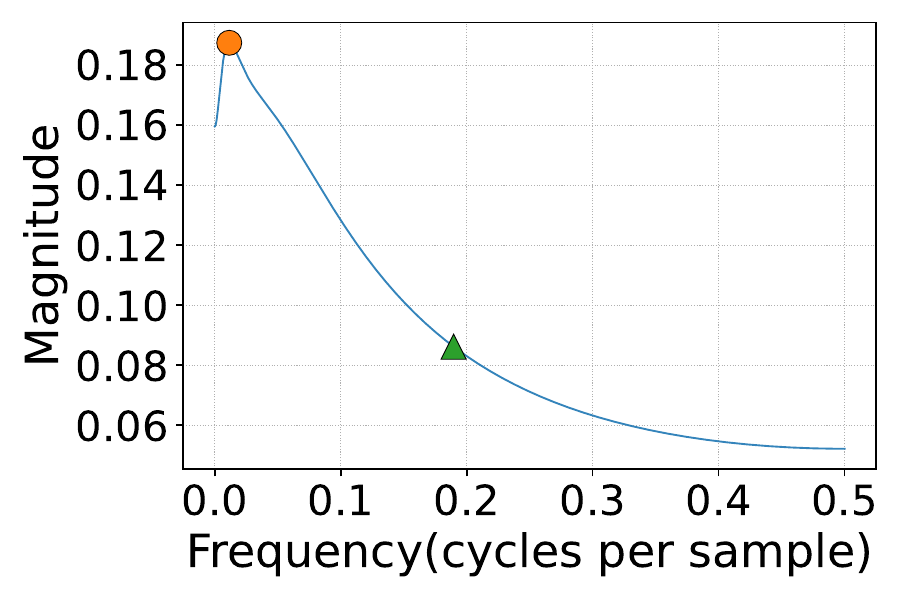}
    \includegraphics[width=0.33\linewidth]{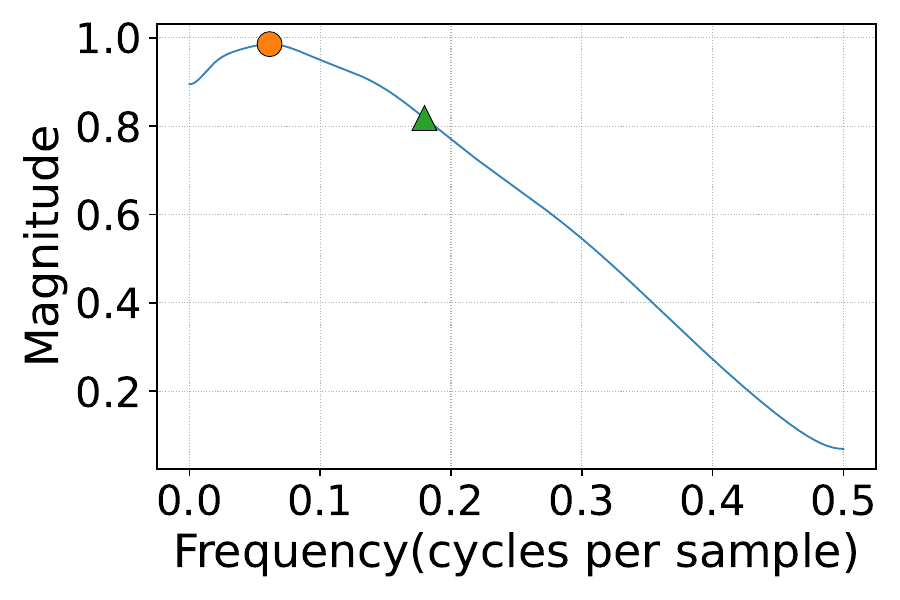}
    \includegraphics[width=0.33\linewidth]{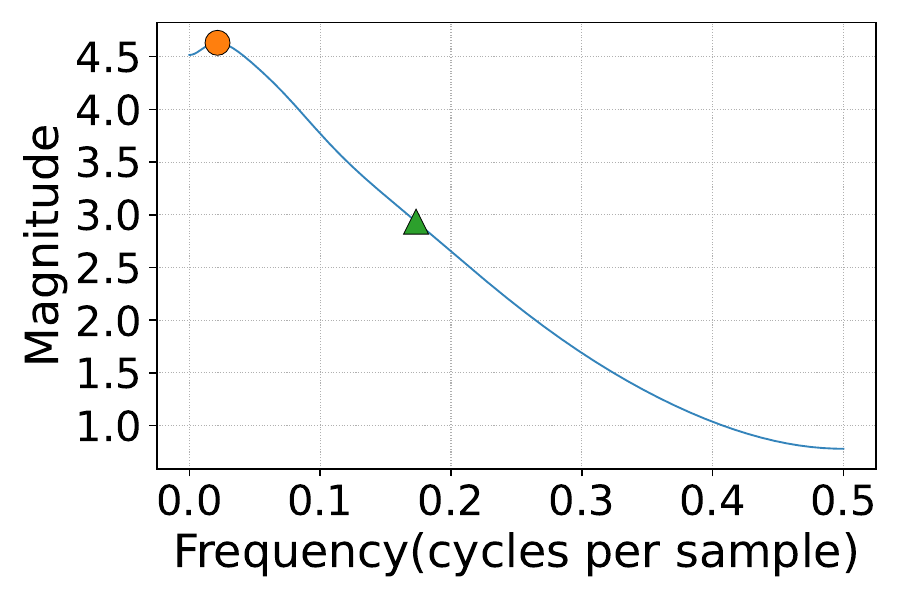}
    \includegraphics[width=0.33\linewidth]{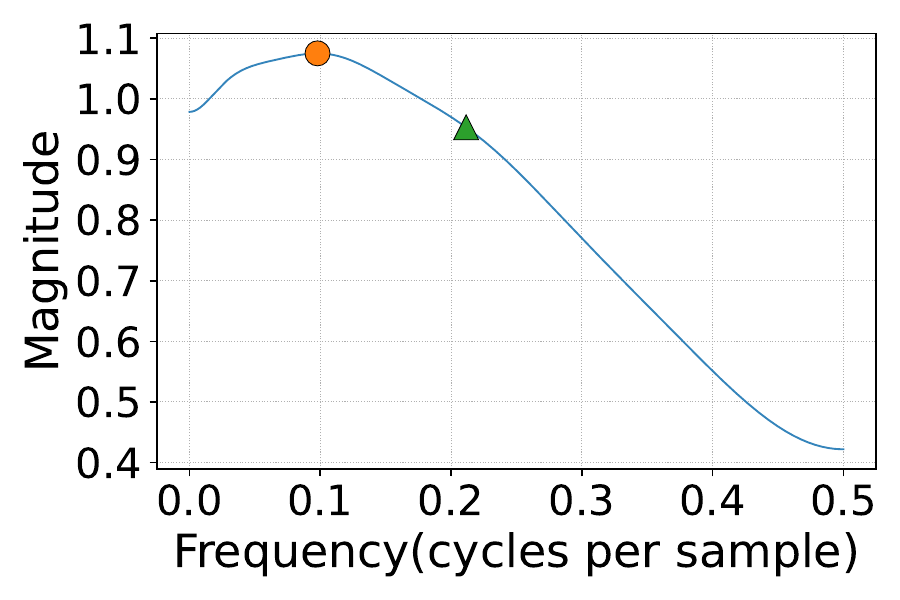}
    \includegraphics[width=0.33\linewidth]{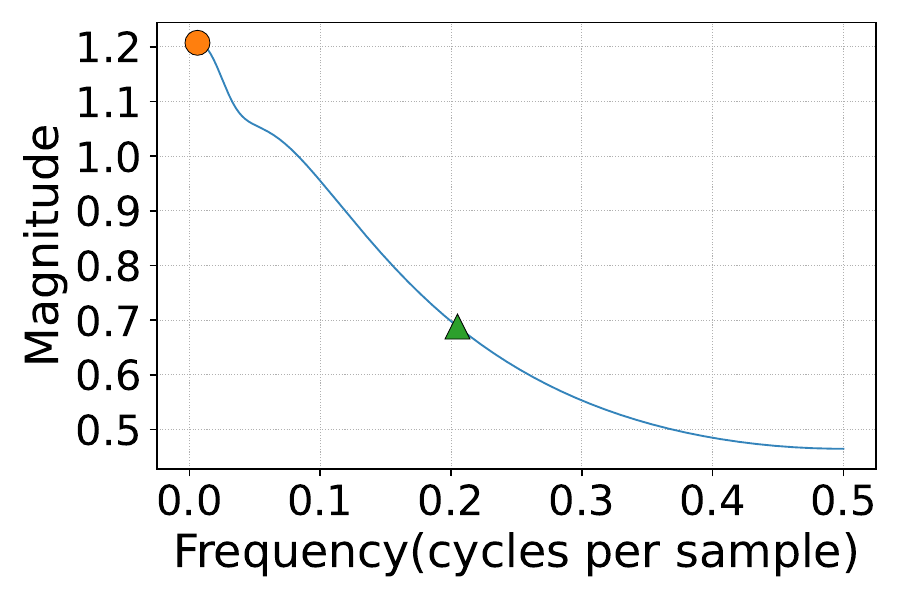}
    \includegraphics[width=0.33\linewidth]{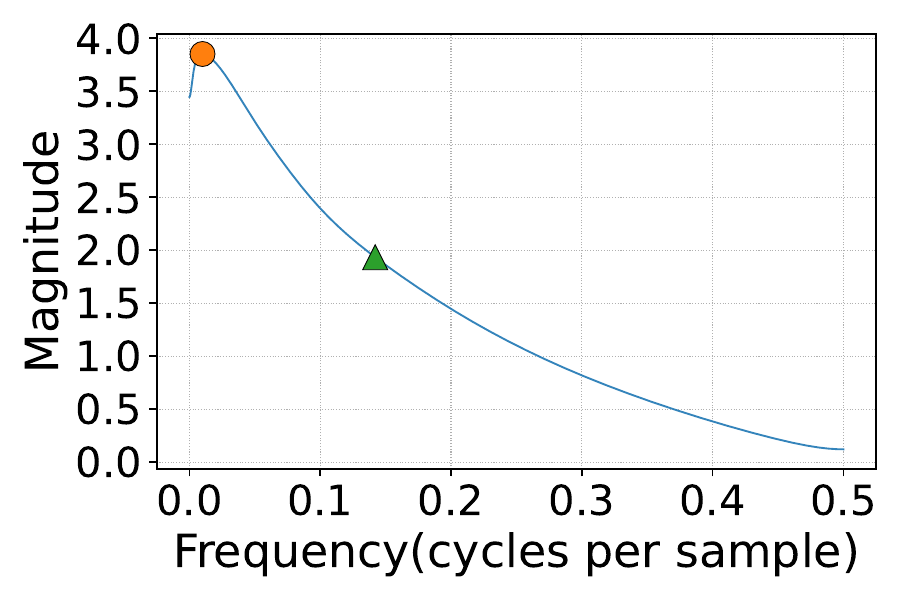}
    \label{cssm_kernel more}
    \caption{CodeSSM Kernels}
    \end{subfigure}
    \begin{subfigure}[b]{\linewidth}
    \includegraphics[width=0.33\linewidth]{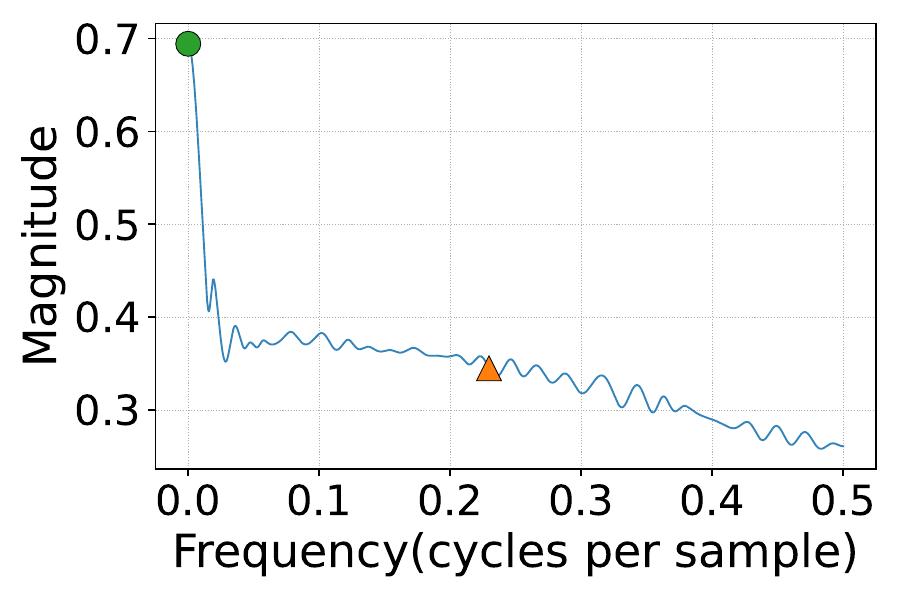}
    \includegraphics[width=0.33\linewidth]{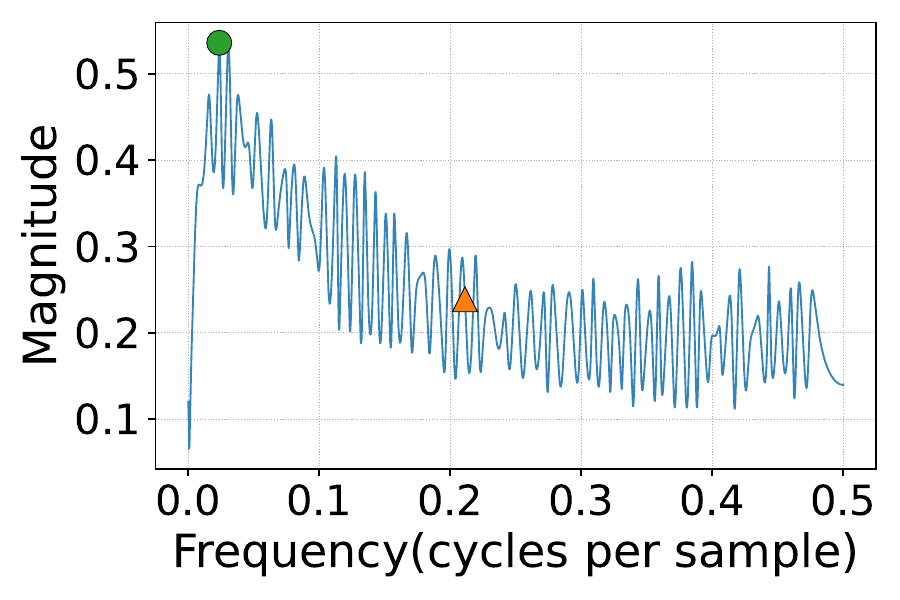}
    \includegraphics[width=0.33\linewidth]{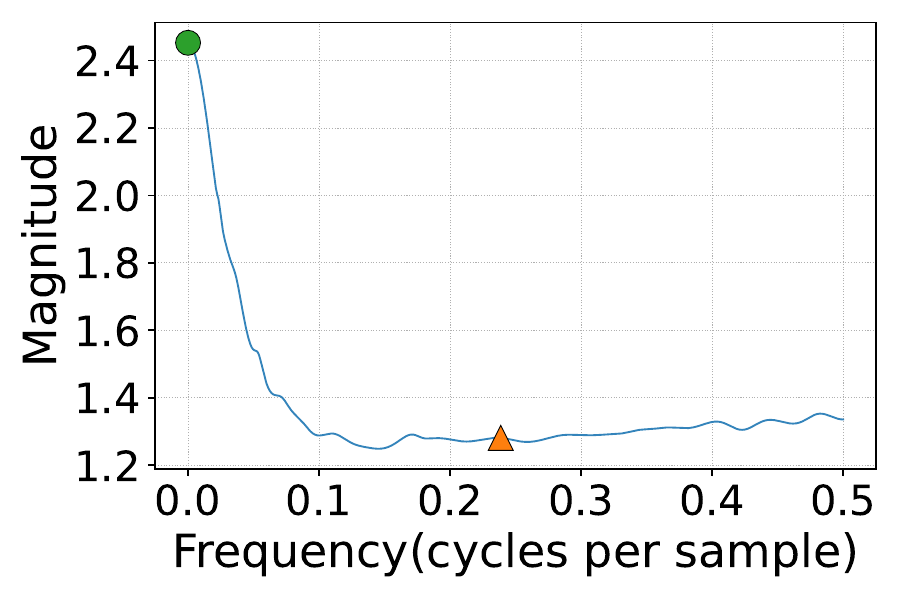}
    \includegraphics[width=0.33\linewidth]{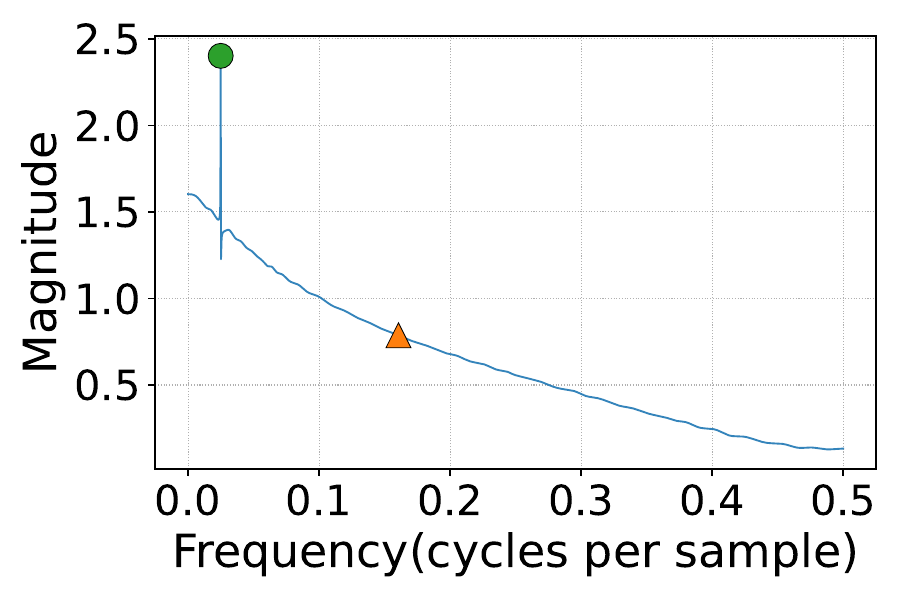}
    \includegraphics[width=0.33\linewidth]{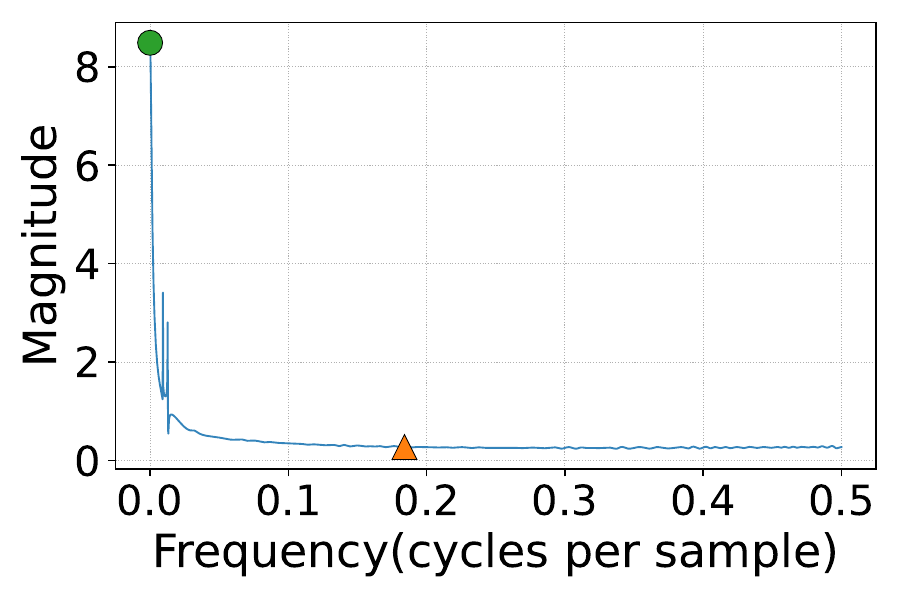}
    \includegraphics[width=0.33\linewidth]{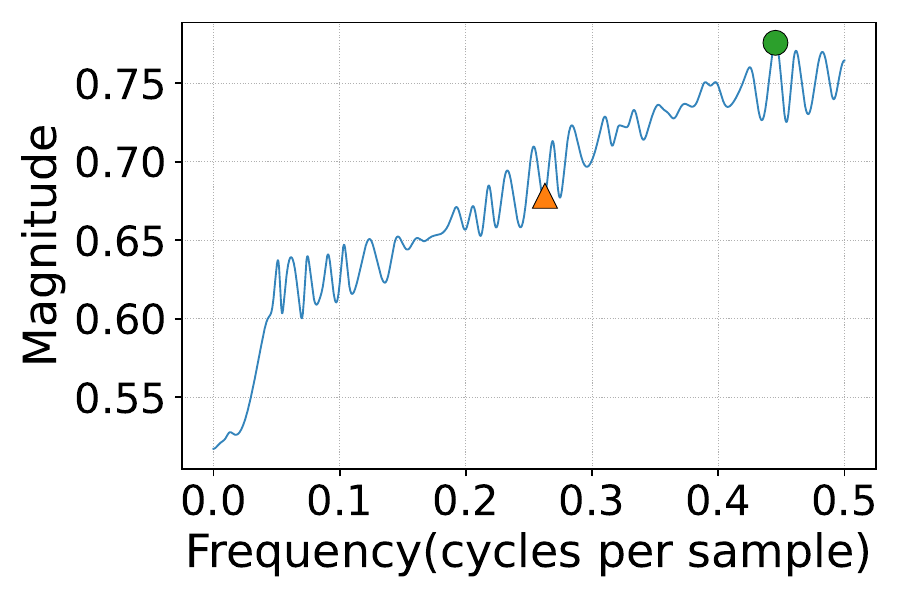}
    \label{hf_kernel more}
    \caption{CodeSSM-HF kernels}
    \end{subfigure}
    \caption{Additional kernel visualization showing kernels of layer 1 (left), layer 5 (center) and the last layer (right). In each sub-figure top row shows forward kernels and bottom row shows backward kernels. The visualization shows the richer kernels learned by CodeSSM-HF.}
    \label{more kernels}
\end{figure*}

\begin{figure*}
    \centering
    \begin{subfigure}[b]{\linewidth}
    \includegraphics[width=0.245\linewidth]{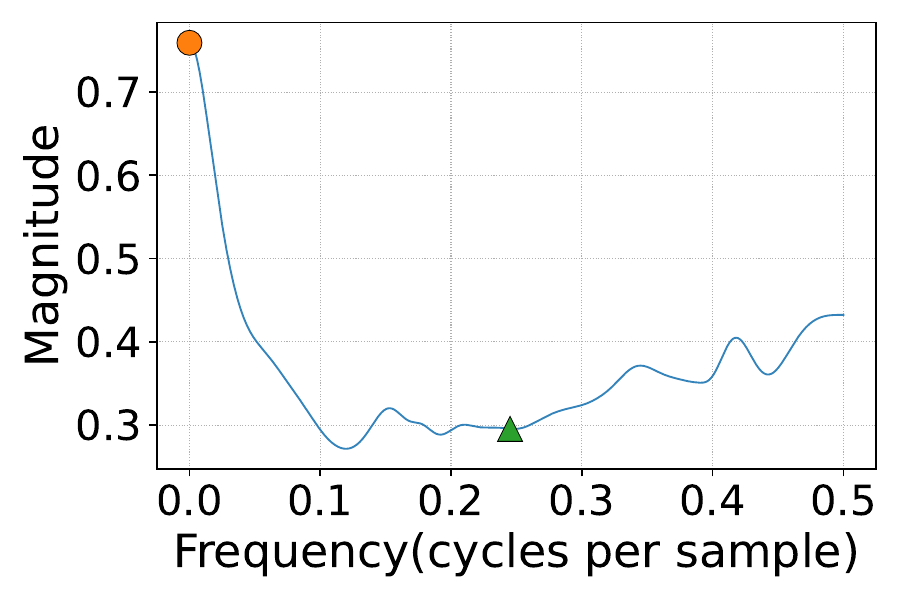}
    \includegraphics[width=0.24\linewidth]{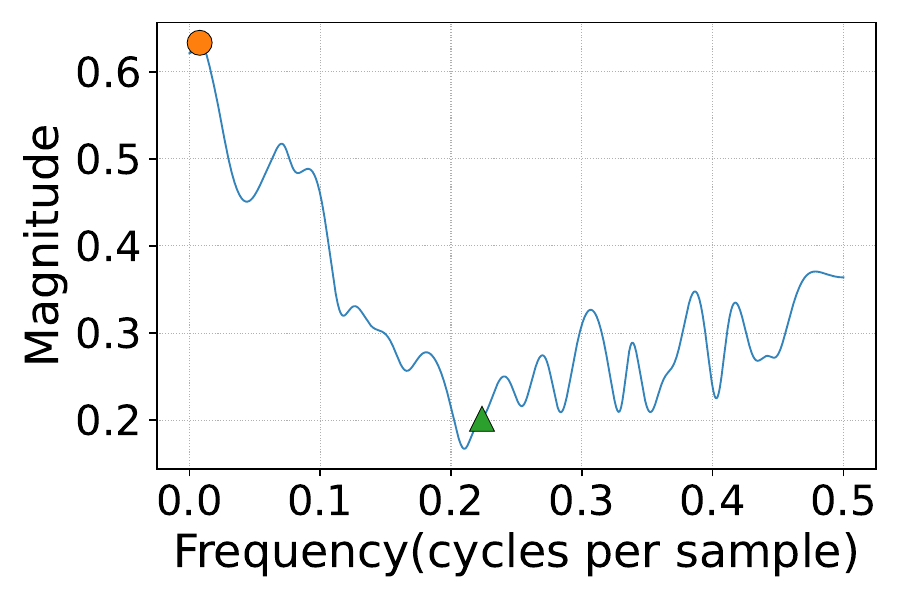}
    \includegraphics[width=0.245\linewidth]{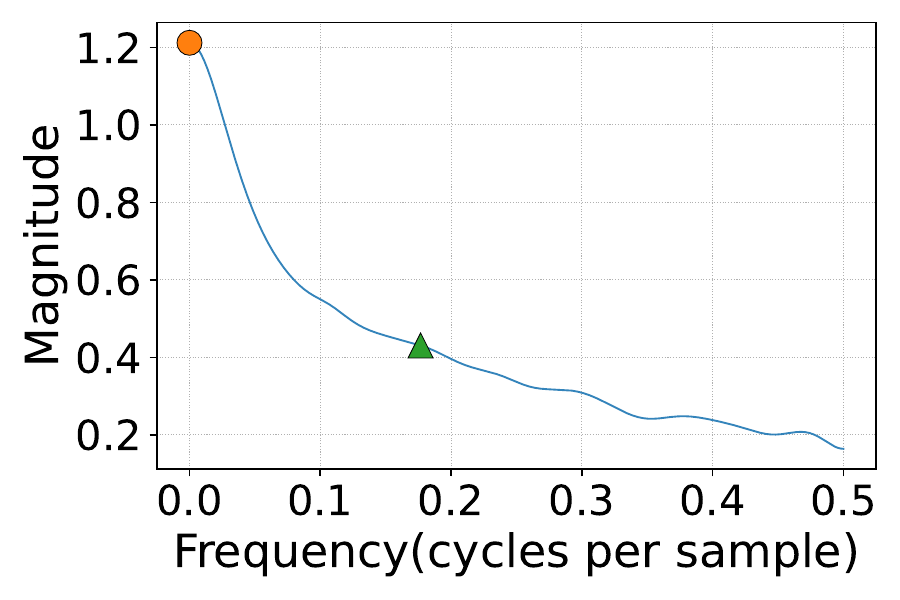}
    \includegraphics[width=0.245\linewidth]{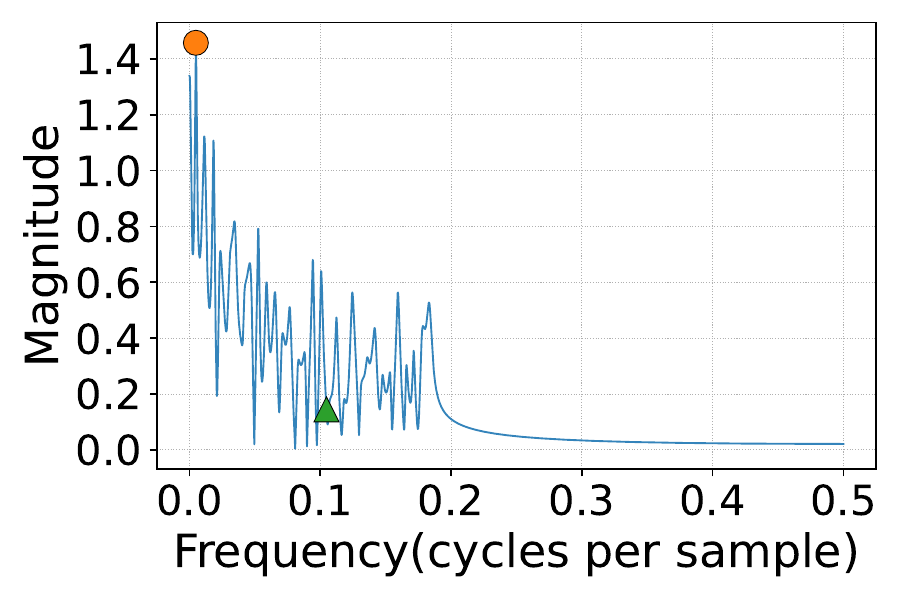}
    \includegraphics[width=0.245\linewidth]{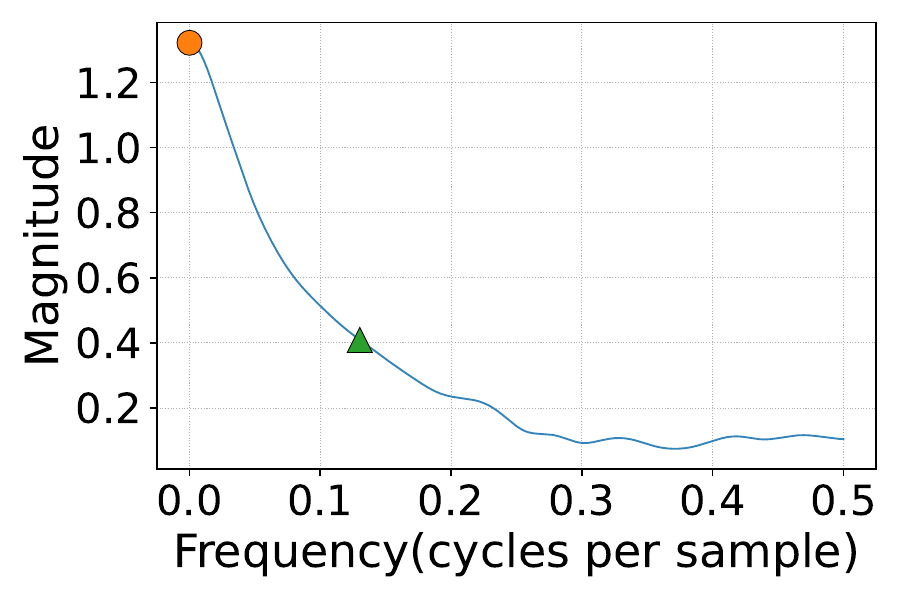}
    \includegraphics[width=0.245\linewidth]{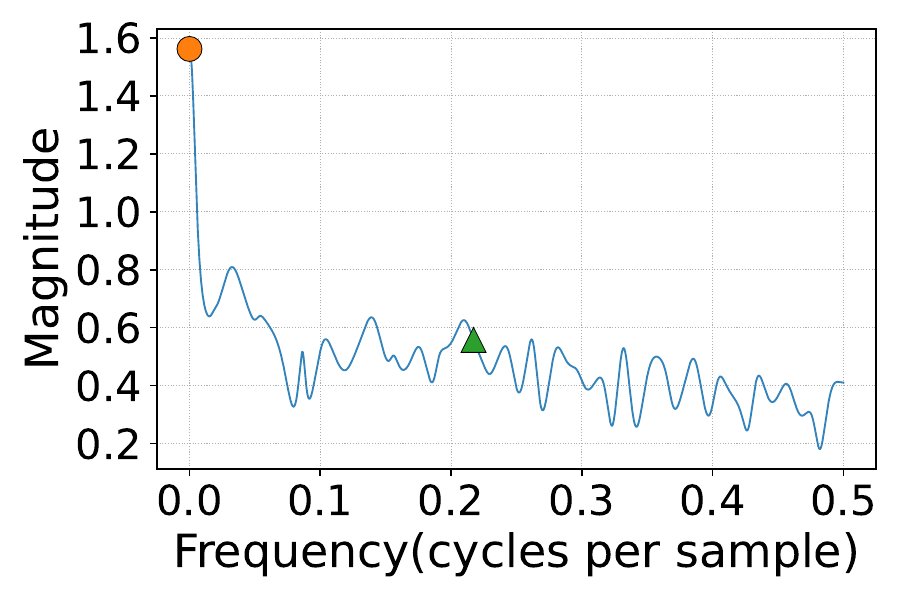}
    \includegraphics[width=0.245\linewidth]{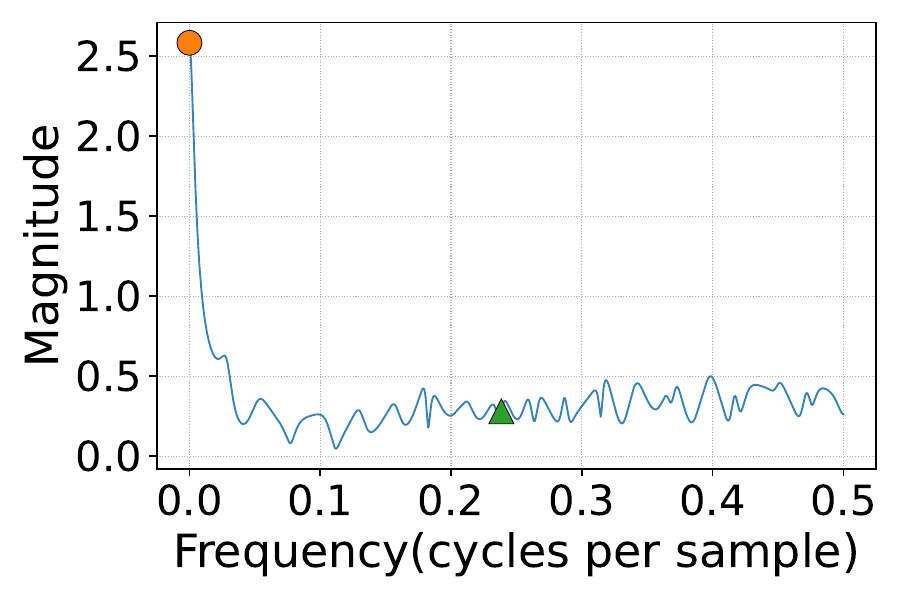}
    \includegraphics[width=0.245\linewidth]{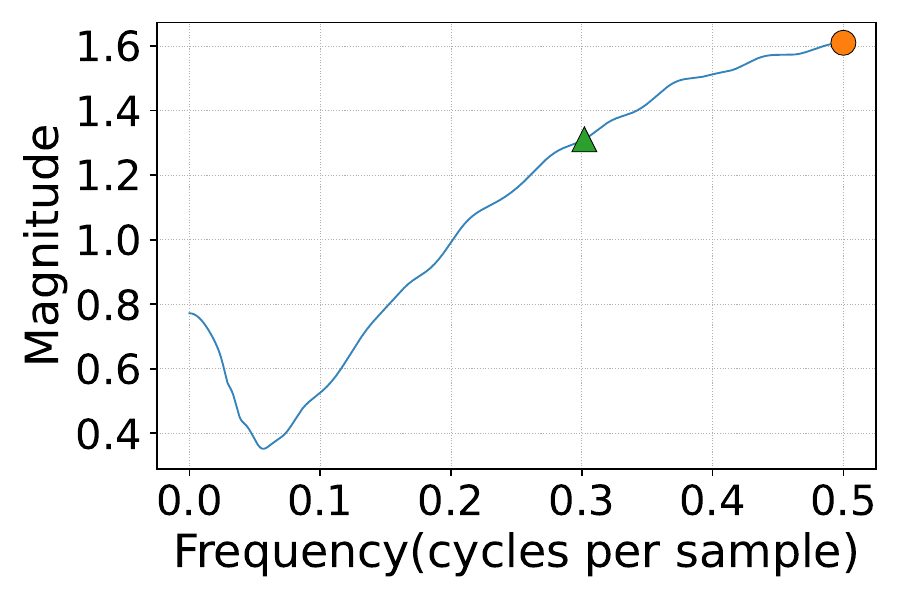}
    \caption{Forward kernels of layer 1 of CodeSSM-8k.}
    \end{subfigure}

    \begin{subfigure}[b]{\linewidth}
    \includegraphics[width=0.245\linewidth]{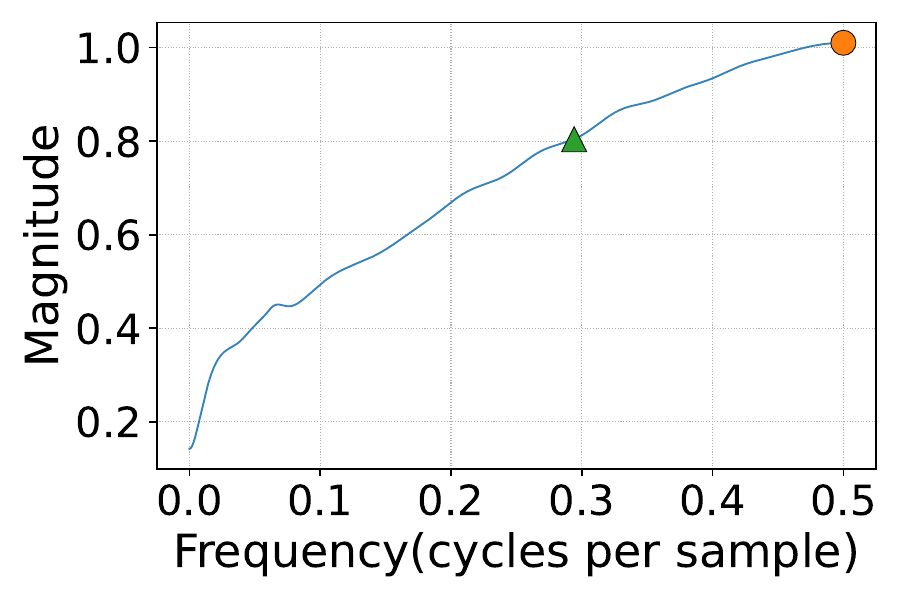}
    \includegraphics[width=0.245\linewidth]{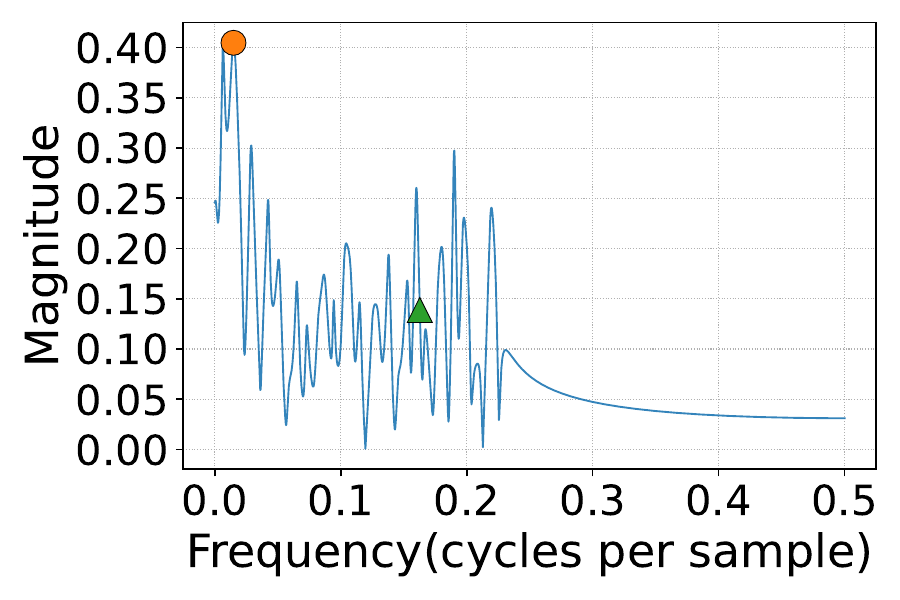}
    \includegraphics[width=0.245\linewidth]{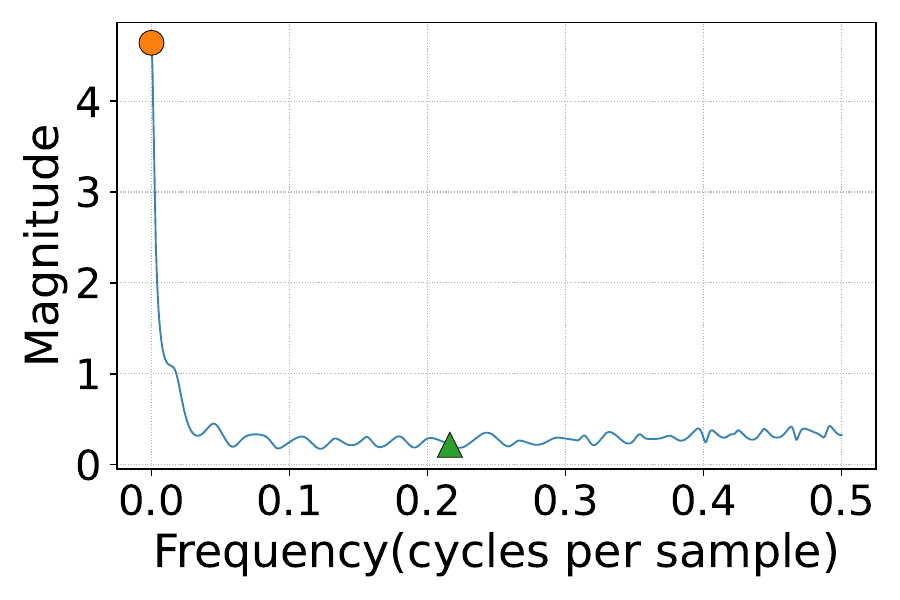}
    \includegraphics[width=0.245\linewidth]{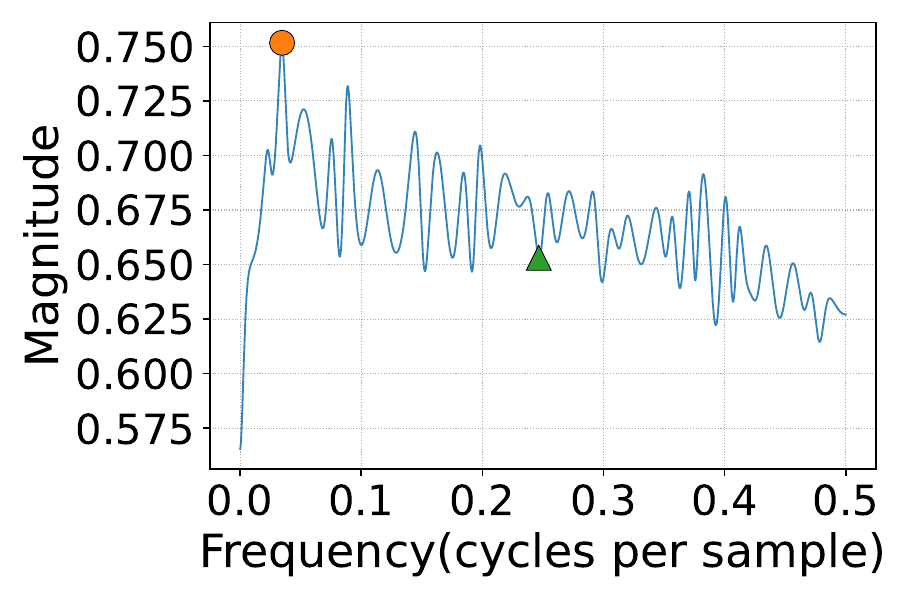}
    \includegraphics[width=0.245\linewidth]{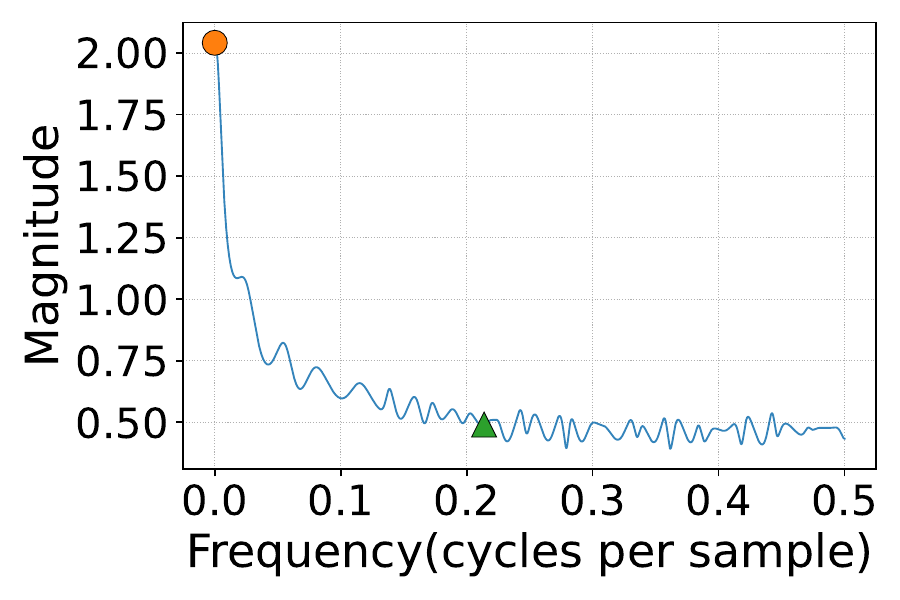}
    \includegraphics[width=0.245\linewidth]{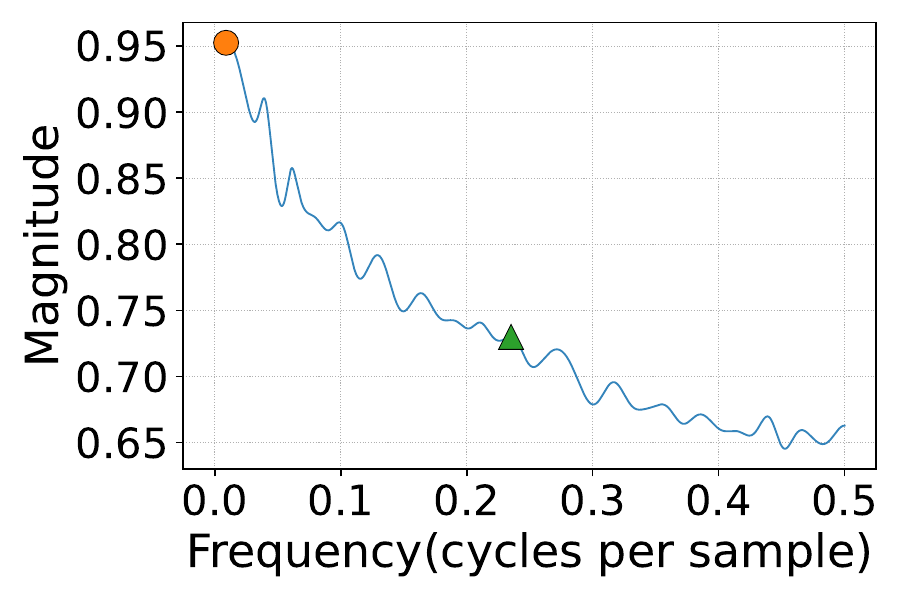}
    \includegraphics[width=0.245\linewidth]{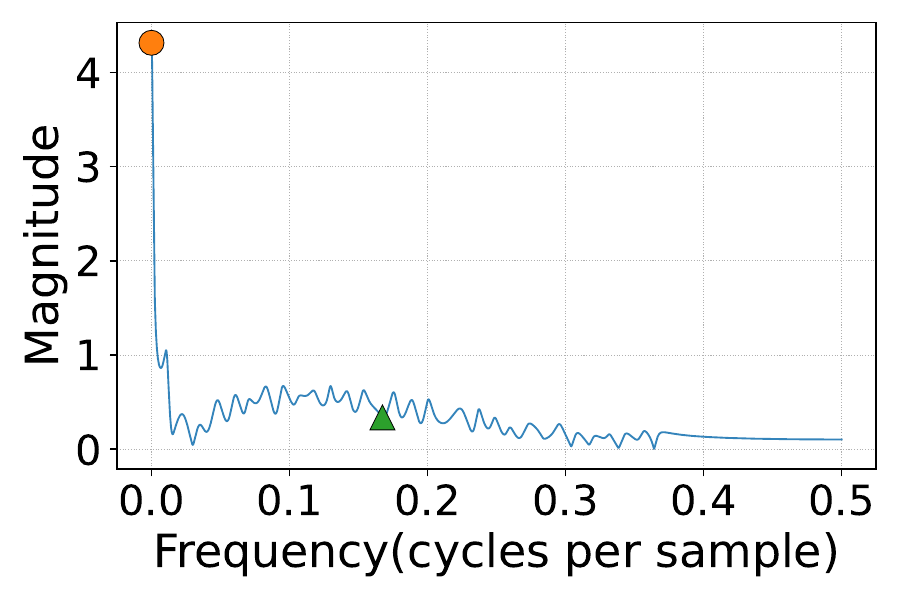}
    \includegraphics[width=0.245\linewidth]{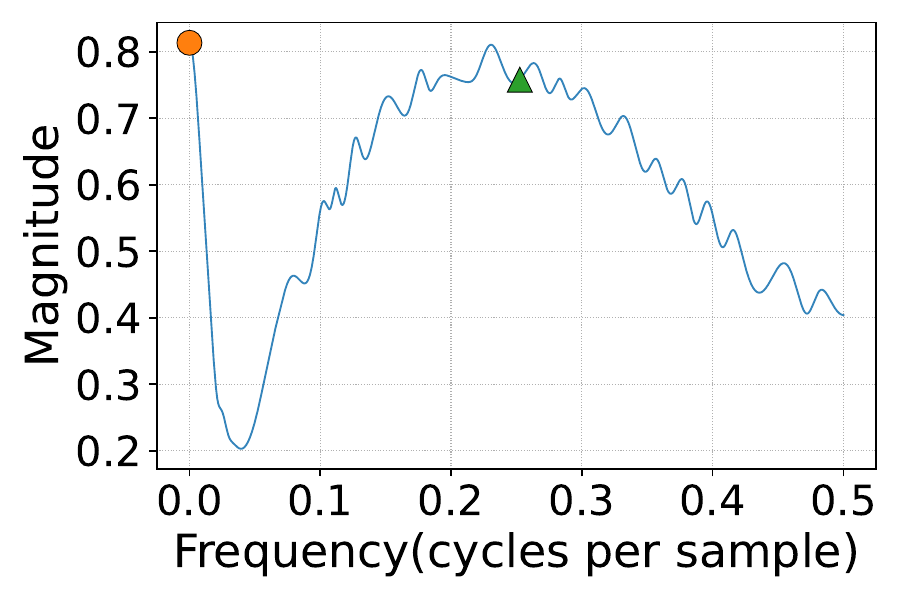}
    \caption{Backward kernels of layer 1 of CodeSSM-8k.}
    \end{subfigure}
    \caption{The forward and backward kernels of layer 1 of CodeSSM-8k model.}
    \label{fig: kernel_8}
\end{figure*}

\begin{figure*}[t]
\centering

     \includegraphics[width=\linewidth]{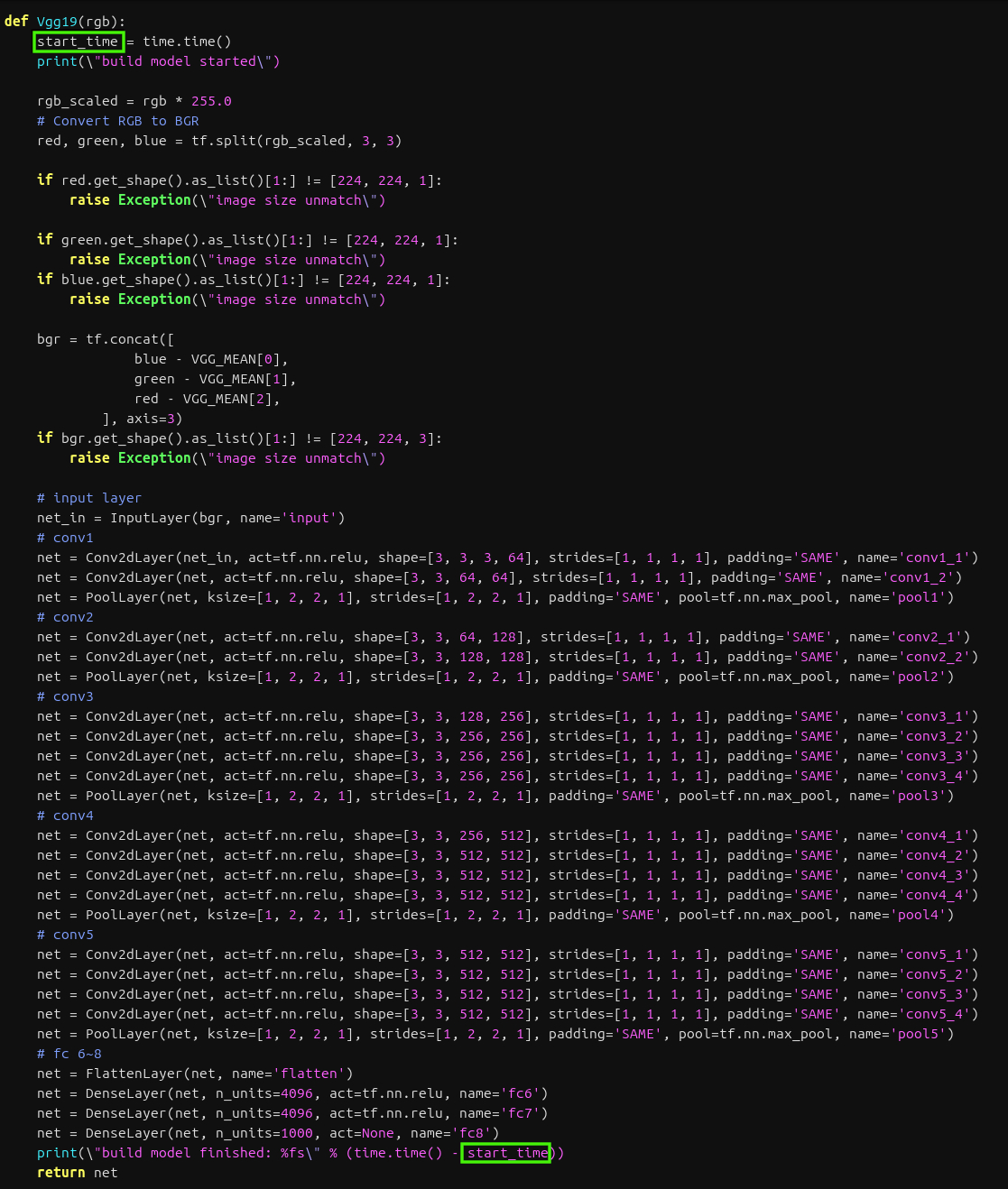} 

    \caption{Example of code where DFG represents long range dependencies. The highlighted variables have a DFG edge and have 1099 tokens between them.}
    \label{fig: dfgeg1}
    
\end{figure*}

\begin{figure*}[t]
\centering
     \includegraphics[width=\linewidth]{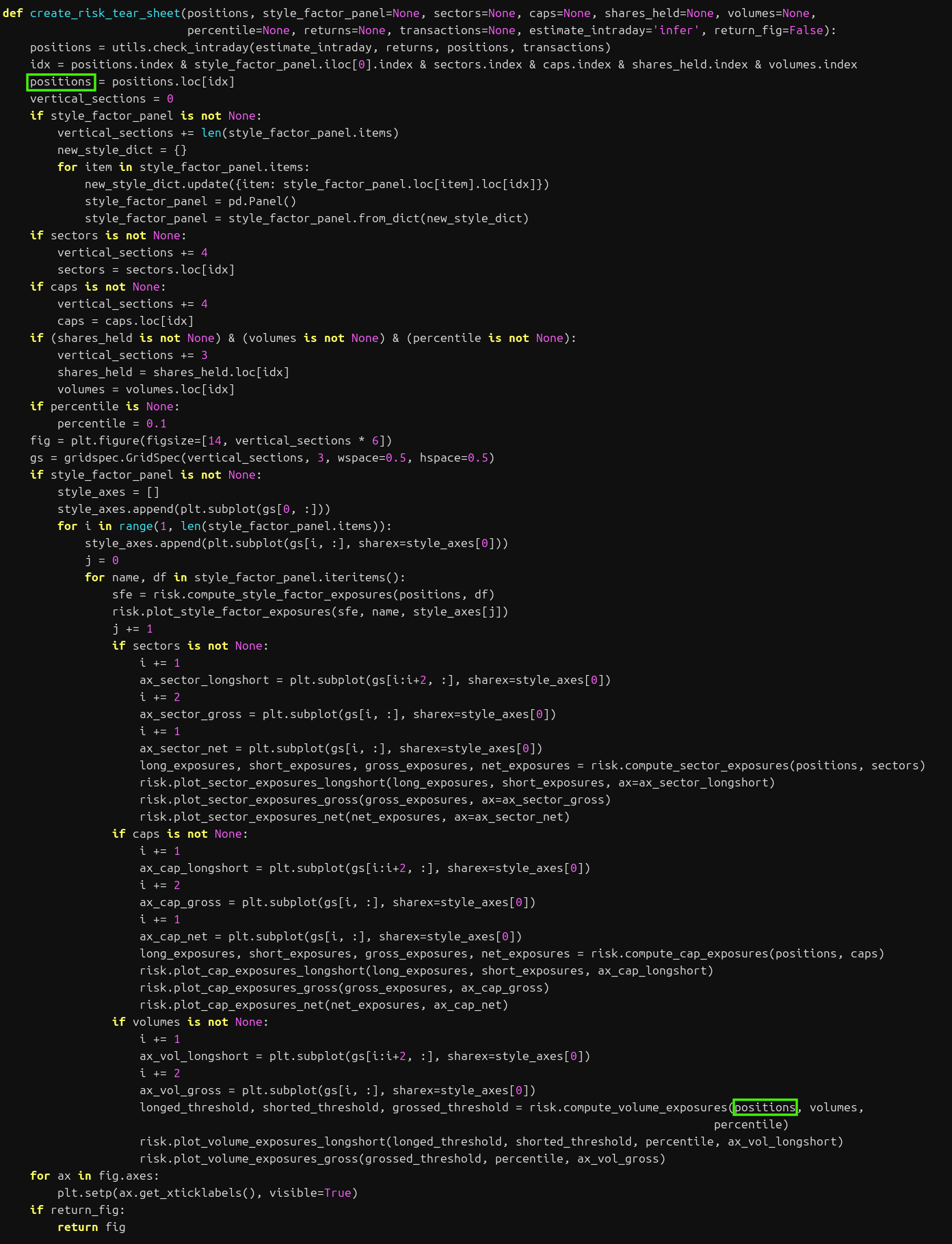}

    \caption{Example of code where DFG represents long range dependencies. The highlighted variables have a DFG edge and have 1486 tokens between them.}
    \label{fig: dfgeg2}
    
\end{figure*}

Additional forward and backward kernels have been shown in Figure \ref{more kernels}. The figure shows kernel for first layer, middle layer (layer 5) and last layer for both the models. The visualization of these kernels shows that CodeSSM-HF kernels learns more complex relations between tokens compared to CodeSSM across all layers.

\section{Training Details}
All variants of CodeSSM compared in Table \ref{div-table} were trained under same conditions. The pre-training was done on 4 A100 80GB GPUs. The models were first trained on Wikipedia data with a sequence length of 128 and and a batch size of 256 for 3 days. The models were then trained on 1.8 million git issues samples from StarCoder dataset \cite{starcoder} for 10 epochs with a per GPU batch size of 64 and 256 sequence length. Finally, the models were trained on 1.8 million code samples from StarCoder dataset with a per GPU batch size of 64 and 256 sequence length. For all pretraining, the learning rate is $5e-5$ with a cosine scheduler and 300 warm up steps. 

The training on git issues and code follows \citet{verma-etal-2025-codessm} but the training on Wikipedia data is significantly less than that of BiGS \citet{bigs}. This results in a slightly lower performance of CodeSSM trained by us compared to \citet{verma-etal-2025-codessm}. Similar to \citet{verma-etal-2025-codessm}, we used the CodeT5plus-220m tokenizer.

\section{CodeSSM-HF Group Size}
We experiment with group sizes of 4 and 8. We found that having a group size of 8, despite having fewer parameters, performed better than group size of 4. The results of both the configuration is shown in Table \ref{grp_abl}.

\begin{table}[t] \centering
\caption{Comparison of CodeSSM-HF with convolution group sizes of 4 and 8.}
\label{grp_abl}
\begin{center}
\begin{tiny}
\begin{sc}
\begin{tabular}{lccccccr}
\toprule
Model & Type Inference \\
 &   (F1) \\
\midrule
CodeSSM-hf-grp-4  &  58.76 \\
CodeSSM-hf-grp-8   & 60.04 \\

\bottomrule
\end{tabular}
\end{sc}
\end{tiny}
\end{center}
\vskip -0.1in
\end{table}

\end{document}